\pdfoutput=1

\documentclass[10pt]{article}
\usepackage[margin=1in]{geometry}

\usepackage[colorlinks,citecolor=red]{hyperref}

\usepackage{graphicx}
\DeclareGraphicsExtensions{.pdf}
\usepackage[justification=Centering,singlelinecheck=false,caption=false,font=footnotesize]{subfig}

\usepackage{natbib}
\bibpunct{(}{)}{;}{a}{}{,}

\usepackage[cmex10]{amsmath}
\usepackage{amsfonts}
\usepackage{bm}
\numberwithin{equation}{section}
\DeclareMathOperator{\argmax}{\textrm{argmax}}
\DeclareMathOperator{\argmin}{\textrm{argmin}}

\usepackage{multirow}
\usepackage{array}

\begin{document}

\title{\textbf{Recent Progress in Image Deblurring}}

\author{Ruxin Wang \hspace{1cm} Dacheng Tao \\
Centre for Quantum Computation \& Intelligent Systems\\Faculty of Engineering \& Information Technology\\University of Technology Sydney\\81-115 Broadway, Ultimo, NSW\\Australia\\
Email: Ruxin.Wang@student.uts.edu.au, dacheng.tao@uts.edu.au}

\date{11 August 2014}

\maketitle

\begin{abstract}
This paper comprehensively reviews the recent development of image deblurring, including non-blind/blind, spatially invariant/variant deblurring techniques. Indeed, these techniques share the same objective of inferring a latent sharp image from one or several corresponding blurry images, while the blind deblurring techniques are also required to derive an accurate blur kernel. Considering the critical role of image restoration in modern imaging systems to provide high-quality images under complex environments such as motion, undesirable lighting conditions, and imperfect system components, image deblurring has attracted growing attention in recent years. From the viewpoint of how to handle the ill-posedness which is a crucial issue in deblurring tasks, existing methods can be grouped into five categories: Bayesian inference framework, variational methods, sparse representation-based methods, homography-based modeling, and region-based methods. In spite of achieving a certain level of development, image deblurring, especially the blind case, is limited in its success by complex application conditions which make the blur kernel hard to obtain and be spatially variant. We provide a holistic understanding and deep insight into image deblurring in this review. An analysis of the empirical evidence for representative methods, practical issues, as well as a discussion of promising future directions are also presented.
\end{abstract}

\section{Introduction} \label{introduction}

Modern imaging sciences, such as consumer photography, astronomical imaging, medical imaging, and microscopy, have been well developed in recent years and a large number of progressive techniques have emerged. These developments have enabled the acquisition of images that are of both higher speed and higher resolution (also referred to as \emph{high-definition}). However, intrinsic or extrinsic factors behind such fast and high-resolution techniques may lead to degradation in the quality of the acquired image, of which blur is one example and is the focus of this paper. From an artistic viewpoint, image blur in photography is sometimes intentional, but in the most common imaging situations, the blur effect corrupts valuable image information and produces visually unattractive images. For example,

\noindent 1) A fast moving car captured by a surveillance system might exhibit significant blurriness in the video or image, leading to difficulty in the license-plate recognition (\emph{motion blur}).

\noindent 2) It is often difficult for photographers to stabilize hand-held cameras for a long period, especially when they are taking pictures in dim lighting conditions which require long exposure times, resulting in blurred images (\emph{camera shake blur}).

\noindent 3) Since most imaging systems have only one focus, the resultant images usually have at most one region that is focused while the others remain blurred (\emph{defocus blur}).

\noindent 4) When capturing a long distance scene, atmospheric particulate matter sometimes prevents photons from moving directly to the sensor, which produces a blurred image (\emph{atmospheric turbulence blur}).

\noindent 5) When a lens has a different refractive index for different wavelengths of light, the lens can fail to focus all colors to the same convergence point which also results in a blurred image (\emph{intrinsic physical blur}).

These blurry image data are a nuisance in a variety of high-quality image-based applications, e.g. image content recognition \citep{nishiyama2011facial}, medical diagnosis and surgery \citep{tzeng2010contourlet}, surveillance monitoring, remote sensing \citep{ma2009deblurring}, and astronomy. Therefore, reducing such blur, which is known as \emph{image deblurring} or \emph{image deconvolution}, is a crucial step in improving the resolution and contrast of high-quality images.

Image deblurring is a traditional inverse problem whose aim is to recover a sharp image from the corresponding degraded (blurry and/or noisy) image. Over the years, numerous methods have been proposed to tackle the \emph{non-blind deblurring problems} or the \emph{blind deblurring problems}, in which classic and well-known classification schemes are employed. The former case, non-blind deblurring, indicates that the blur kernel is assumed to be known and a sharp image can be induced from both the blurry image and the kernel. Typical methods include the Richardson-Lucy method \citep{richardson1972bayesian,lucy1974iterative} and Wiener filter \citep{wiener1949extrapolation}. By contrast, the blind deblurring problem, which is more practical, means that the blur kernel is unknown, and the task therefore becomes one of estimating the sharp image and/or the kernel from the degraded image. This kind of method dates back to the 1970s \citep{stockham1975blind,cannon1976blind}. In recent years, many novel approaches have been presented to handle both the non-blind and blind deblurring problem, driven by a variety of motivations. The above simple classification scheme is not sufficiently competent to discover the connections and details of modern image deblurring methods, thus we intend in this survey to manage the organization of this domain through an analysis of ill-posedness. Ill-posedness is the most severe problem in image deblurring. In the non-blind case, the observed blurry image does not uniquely and stably determine the sharp image due to the ill-conditioned nature of the blur operator \citep{bertero1998introduction}. This means that if the assumed/given blur kernel and the true kernel are slightly mismatched, or if the blurry image is also corrupted by noise, the recovered image may be much worse than the underlying sharp image. In the blind case, even if the blur operator is not ill-conditioned, the deblurring problem will still be inherently ill-posed since there is an infinite set of image-blur pairs that can synthesize the observed blurry image.

Contemporary researchers have been devoted to developing new models and new prototypes, or improving the efficiency of optimization methods, to deal with ill-posedness. From the model construction perspective, most methods can be grouped into the following five categories: \emph{Bayesian inference framework}, \emph{variational methods}, \emph{sparse representation-based methods}, \emph{homography-based modeling}, and \emph{region-based methods}. In the Bayesian inference framework, priors are introduced to impose uncertainty attributes on either the unknown sharp image or the unknown blur kernel, or both. This operation is intended to reduce the volume of the search space, where the problem's solution lies, to suppress the ill-posedness. Variational methods render the solution unique and stable through the incorporation of regularization techniques whose role is similar to the prior's in Bayesian inference, i.e. regularizing the solution into a constraint space. Sparse representation, a progressive topic in recent years, benefits from the fact that natural images are intrinsically sparse in some domains. The sparse property of the natural images in these domains is well suited to tackle the ill-posedness of the deblurring problem. The homography-based modeling and region-based methods are intentionally proposed for spatially variant deblurring (see Section \ref{imageblurform}). Since spatially variant blur cannot be modeled by a single blur kernel with limited support, researchers usually approximate the blurry effect by using a union of multiple kernels or homographies. Due to the growing number of unknowns, this type of method lead to a worse situation in the inverse problem. To overcome this issue, homography-based methods derive the spatially variant kernel as an integration of "temporal" homographies whose continuity is further constrained according to the properties of the blur effects in the image. Meanwhile, region-based methods focus on the restriction of spatial variations of the blur kernels. Other methods which do not fall into the above frameworks include Projection-based method, kernel regression, stochastic deconvolution, and spectral analysis.

Besides the above mathematical viewpoint, a progressive direction in deblurring concerns the development of hardware prototype systems. Traditional camera systems are generally equipped with a single lens, single focus, and single consecutive exposure. Using these systems, a blurred image is the only achievable resource, and a large amount of useful information is lost during imaging \citep{wehrwein2010computational}. However, by using hardware modifications, additional observations or principles can be easily accessed to help the derivation of either the sharp image or the blur kernel, resulting in reduced ill-posedness.

In reviewing the literatures on image deblurring, we have been attracted by a number of interesting discoveries. The one such discovery is that more and more approaches have been proposed to use multiple images to jointly deblur, or at least to assist the deblurring of a target image. These images may be either correlated or uncorrelated. Multiple image deblurring is possible because a set of common patterns exist behind natural scenes which could be used to generate all kinds of natural structures. One option to incorporate multiple images is to learn the deblurring functions or subspaces by using the acquirable sharp-blurry image pairs from additional datasets. Another option is hardware modification, as noted earlier. In terms of single image deblurring, another concern claims that the global solutions of certain models do not necessarily correspond to the true solution of the problem, and a more appropriate model should be discovered. Considering these findings, the possible future directions in the image deblurring field may be summarized as learning-based methods and hardware modifications. In fact, research in these areas has started and is flourishing. As well, new single image deblurring models are necessary yet challenging to be exploited to complement the drawbacks of existing models.

The rest of this paper is organized as follows: The formulation of image deblurring is first introduced in Section \ref{imageblurform}. Five classes of modeling methods are described and comprehensively analyzed in Sections \ref{bayesinferframe}-\ref{regionmethod} and other methods are listed in Section \ref{others}. Section \ref{practicalissue} discusses several issues usually encountered in designing deblurring methods. Insights to promising future directions in this field are given in Section \ref{futuredirect}. The experimental analyses and evaluations are given in Section \ref{evaluation}, and concluding remarks are made in Section \ref{conclusion}.

\section{Image Blur Formulation} \label{imageblurform}

The blur kernel, also known as (aka) \emph{point spread function} (PSF), causes an image pixel to record light photons from multiple scene points. Many factors can extrinsically or intrinsically cause image blur. As briefed above, blur is generally one of five types: object motion blur, camera shake blur, defocus blur, atmospheric turbulence blur, and intrinsic physical blur.\footnote{Strictly speaking, both the object motion blur and the camera shake blur belong to motion blur, while the defocus blur is an instance of the intrinsic physical blur. Here we separate them to highlight the focus of different approaches.} These types of blur degrade an image in different ways. An accurate estimation of the sharp image and the blur kernel requires an appropriate modeling of the digital image formation process. Hence before introducing the blur types, we first focus on analyzing the image formation model.

Recall that image formation encompasses the radiometric and geometric processes by which a 3D world scene is projected onto a 2D focal plane. In a typical camera system\footnote{Different imaging systems correspond to different image formulation models. Here, we start with the camera system and end with the most common formulation for other systems: see equation (\ref{simpleform}).}, light rays passing through a lens's finite aperture are concentrated onto the focal point. This process can be modeled as a concatenation of the perspective projection and the geometric distortion. Due to the non-linear sensor response, the photons are transformed into an analog image, from which the final digital image is formed by discretization \citep{delbracio2012non}.

Mathematically, the above process can be formulated as
\begin{equation}
y=\textrm{S}(f(\textrm{D}(\textrm{P}(s)*h_{ex})*h_{in}))+n,
\end{equation}
where $y$ is the observed blurry image plane, $s$ is the real planar scene, $\textrm{P}(\cdot)$ denotes the perspective projection, $\textrm{D}(\cdot)$ is the geometric distortion operator, $f(\cdot)$ denotes the nonlinear \emph{camera response function} (CRF) that maps the scene irradiance to intensity, $h_{ex}$ is the extrinsic blur kernels caused by external factors such as motion, $h_{in}$ denotes the blur kernels determined by intrinsic elements such as imperfect focusing, $*$ is the mathematical operation of convolution, $\textrm{S}(\cdot)$ denotes the sampling operator due to the sensor array, and n models the noise.

The above process explicitly describes the mechanism of blur generation. However, what we are interested here is the recovery of a sharp image having no blur effect, rather than the geometry of the real scene. Hence, focusing on the image plane and ignoring the sampling errors, we obtain
\begin{equation}\label{crfform}
y\approx f(x*h)+n,
\end{equation}
where $x$ is the latent sharp image induced from $\textrm{D}(\textrm{P}(s))$, and $h$ is an approximated blur kernel combining $h_{ex}$ and $h_{in}$, as assumed by most approaches. The effect of the CRF in this formulation, which will be discussed in Section \ref{crfeffects}, will have a significant influence on the deblurring process if it is not appropriately addressed. For simplification, most researchers neglect the effect of the CRF, or explore it as a preprocessing step. Let us remove the effect of the CRF to obtain a further simplified formulation:
\begin{equation}\label{simpleform}
y=x*h+n.
\end{equation}
This equation is the most commonly-used formulation in image deblurring.

Given the above formulation, the general objective is to recover an accurate $x$ (non-blind deblurring), or to recover $x$ and $h$ (blind deblurring), from the observation $y$, while simultaneously removing the effects of noises $n$. Taking into account a whole image, equation (\ref{simpleform}) is often represented as a matrix-vector form:
\begin{equation}\label{vectorform}
\mathbf{y}=\mathbf{H}\mathbf{x}+\mathbf{n},
\end{equation}
where $\mathbf{y}$, $\mathbf{x}$ and $\mathbf{n}$ are lexicographically ordered column vectors representing $y$, $x$ and $n$, respectively. $\mathbf{H}$ is a Block Toeplitz with Toeplitz Blocks (BTTB) matrix derived from $h$.

While the noise may originate during image acquisition, processing, or transmission, and is dependent on the imaging system, term $n$ is often modeled as Gaussian noise \citep{levin2011understanding,delbracio2012non}, Poisson noise \citep{kenig2010blind,carlavan2012sparse,lefkimmiatis2013poisson,ma2013dictionary} or impulse noise \citep{chan2010efficient,cai2010fast}. Equation (\ref{simpleform}) is not suitable for describing these noises since it only characterizes the plus case and the signal-uncorrelated case. On the other hand, Poisson and impulse noise are usually signal-correlated. A more detailed summary of the specific degradation models with respect to different noise assumptions can be found in Table \ref{noisetable}. In this paper, we prefer to consider Gaussian noise.

\begin{table*}[!ht]
\begin{center}
\caption{Summary on different noise assumption}\label{noisetable}
\footnotesize
\begin{tabular}{|>{\centering}p{1.7cm}|>{\centering}p{2.8cm}|c|}
\hline
Noise Assumption & Image Degradation Model & Noise Distribution\\
\hline
Gaussian noise & $y=x*h+n$ & $p(y|x,h)=\prod_i\frac{1}{\sigma\sqrt{2\pi}}\exp\left(-\frac{(y_i-(x*h)_i)^2}{2\sigma^2}\right)$\\
\hline
Poisson noise & $y=n(x*h)$ & $p(y|x,h)=\prod_i\frac{((x*h)_i)^{y_i}\exp\left(-(x*h)_i\right)}{y_i!}$\\
\hline
\multirow{2}{*}{Impulse noise} & \multirow{2}{*}{$y=n(x*h)$} &
$p(y_i|(x*h)_i)=
\begin{cases}
(x*h)_i, & \text{with probability } 1-r\\
n_{max}, & \text{with probability } r/2\\
n_{min}, & \text{with probability } r/2
\end{cases}$\\
\cline{3-3}
& &
$p(y_i|(x*h)_i)=
\begin{cases}
(x*h)_i, & \text{with probability } 1-r\\
n_i, & \text{with probability } r
\end{cases}$\\
\hline
\end{tabular}
\end{center}
\scriptsize Notes: All noises are assumed \emph{i.i.d.} with respect to the image location $\{i\}$.  In the Gaussian case, the noise is assumed to be of mean zero and variance $\sigma^2$. Impulse noise includes two types: salt-and-pepper noise (the upper model) and random-valued impulse noise (the lower model). In salt-and-pepper case, $n_{max}$ and $n_{min}$ denote the maximum and minimum of the intensity range, while in random-valued case $n_i$ is sampled from a uniform distribution in $[n_{min},n_{max}]$.
\end{table*}

Traditionally, the blur kernel in image deblurring methods is usually assumed to be \emph{spatially invariant} (aka \emph{uniform} blur), which means that the blurry image is the convolution of a sharp image and a single kernel \citep{lucy1974iterative,wiener1949extrapolation,fergus2006removing,kenig2010blind}. However in practice, it has been noted that the invariance is violated by complex motion or other factors. Thus \emph{spatially variant} blur (aka \emph{non-uniform} blur) is more practical \citep{levin2011understanding}, but is hard to address. In this case, the matrix $\mathbf{H}$ in equation (\ref{vectorform}) is no longer a BTTB matrix since different pixels in the image correspond to different kernels. The number of unknown variables  in the blind deblurring problem is therefore significantly increased, but fortunately, $\mathbf{H}$ can be characterized by the specific properties of the blur types. If we are given the specific motions or factors which cause the blur, the problem can be effectively constrained by prior knowledge. Next, we will introduce the attributes of the five blur types mentioned above.

\subsection{Blur Types} \label{blurtypes}

\subsubsection{Object Motion Blur}

\begin{figure}[t]
\centering
\subfloat[Simulated linear motion]{\includegraphics[height=1.5in]{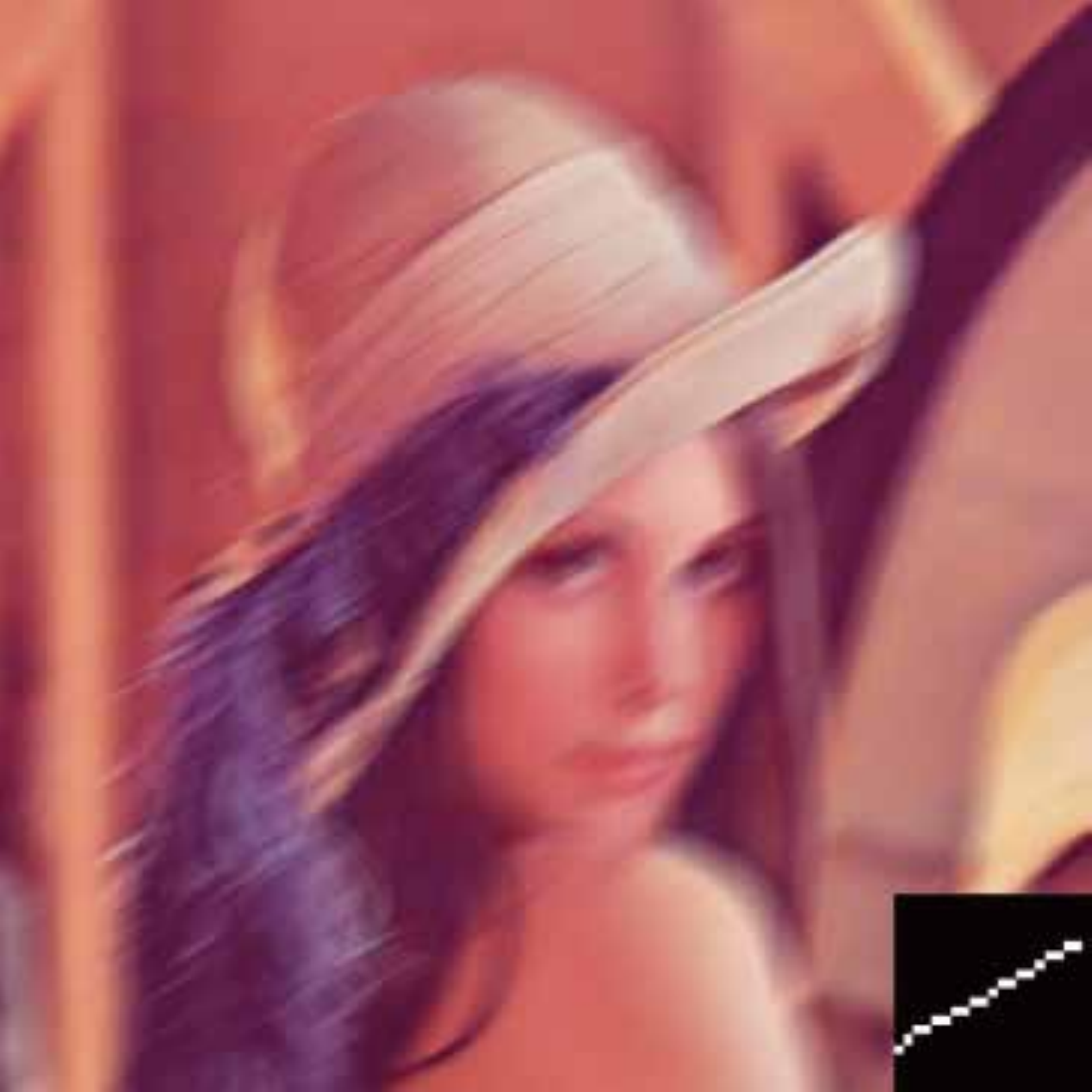}\label{figmotiona}}
\hspace{0.5cm}
\subfloat[Real motion]{\includegraphics[height=1.5in]{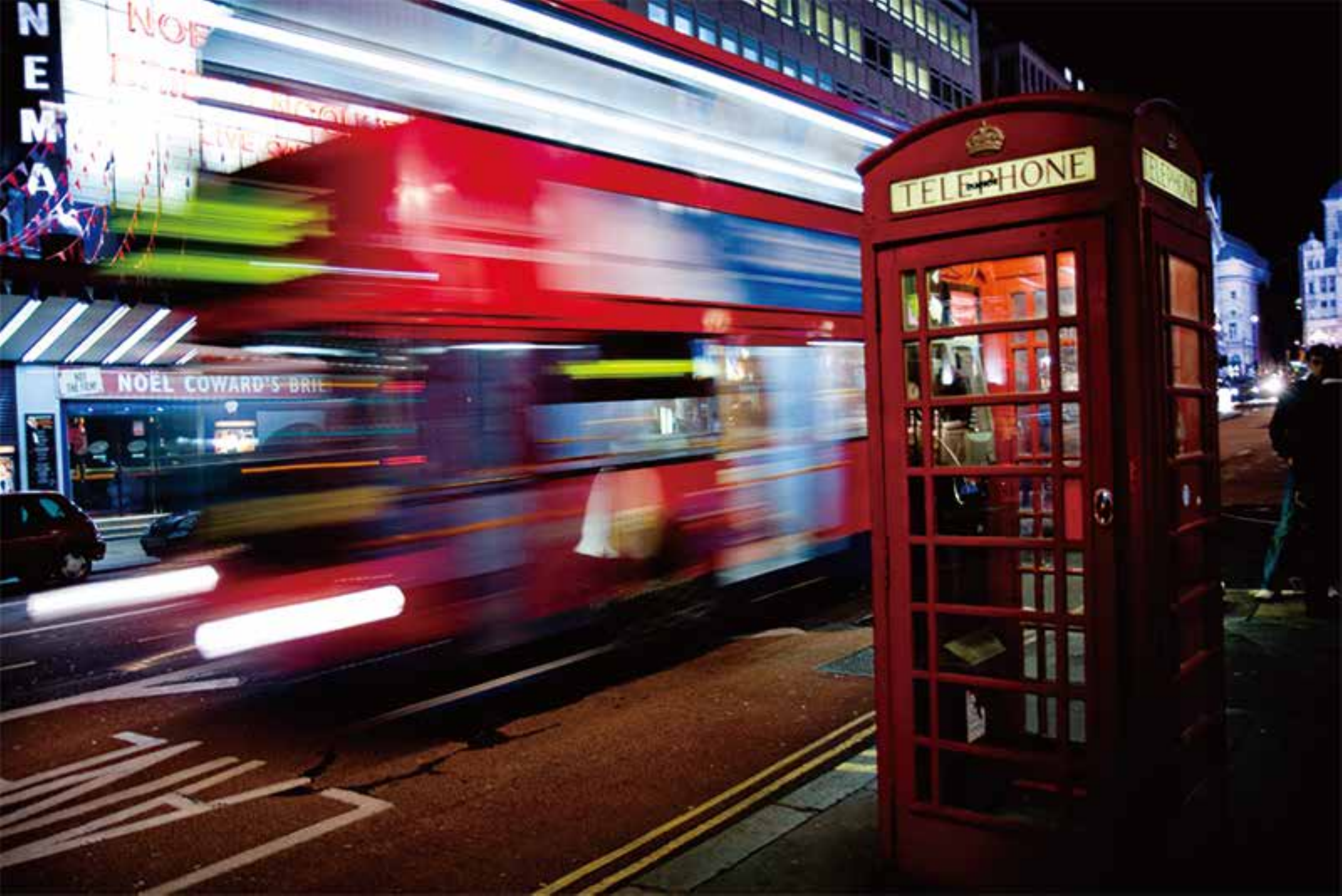}\label{figmotionb}}
\caption{Object Motion Blur}
\end{figure}

\emph{Object motion blur} is caused by the relative motion between an object in the scene and the camera system during the exposure time. This type of blur generally occurs in capturing a fast-moving object or when a long exposure time is needed. If the motion is very fast relative to the exposure period, we may approximate the resultant blur effect as a linear motion blur, which is represented as a 1D local averaging of neighboring pixels and given by
\begin{equation}\label{linearmotion}
h(i,j;L,\theta)=
\begin{cases}
\frac{1}{L}, &\text{if } \sqrt{i^2+j^2}\leq\frac{L}{2} \text{ and } \frac{i}{j}=-\tan\theta,\\
0, &\text{otherwise},
\end{cases}
\end{equation}
where $(i,j)$ is the coordinate originating from the center of $h$, $L$ the moving distance and $\theta$ the moving direction. Fig. \ref{figmotiona} gives a simulated example of a Lena standard test image corrupted by $30^\circ$-directional motion. In practice, however, real motions are extremely complex and cannot be approximated by such a simple parametric model. An appropriate way to handle this issue is to use the non-parametric model, i.e. no explicit shape constraint is imposed on the blur kernel, and the only assumption is that the kernel needs to follow the motion path. A more serious issue is in an image where only the region of moving objects is disturbed by the blur kernel, while other regions remain clear. As shown in Fig. \ref{figmotionb}, the bus was moving fast while the surroundings were static when the picture was taken. In this case, we cannot uniformly process the whole image by a single kernel even if the kernel accurately represents the true motion \citep{levin2006blind}. Since moving objects (e.g. the bus) occupy parts of the image, the most commonly-used approach to handle this problem is to segment the blurry regions from the clear background \citep{chakrabarti2010analyzing}, which will be detailed in Section \ref{regionmethod}, or to simulate the motion as a sequence of homographies \citep{tai2010coded}, which will be discussed in Section \ref{homographymethod}.

\subsubsection{Camera Shake Blur}

\begin{figure}[t]
\centering
\subfloat[Camera translation]{\includegraphics[height=1.5in]{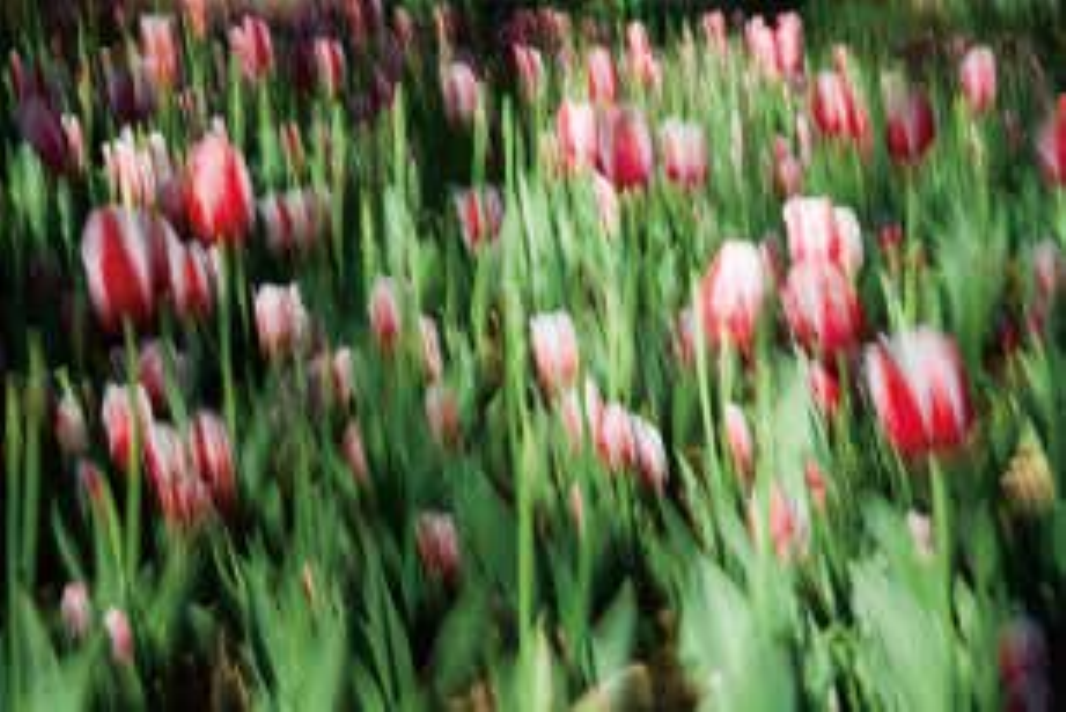}\label{figshakea}}
\hspace{0.5cm}
\subfloat[Camera rotation]{\includegraphics[height=1.5in]{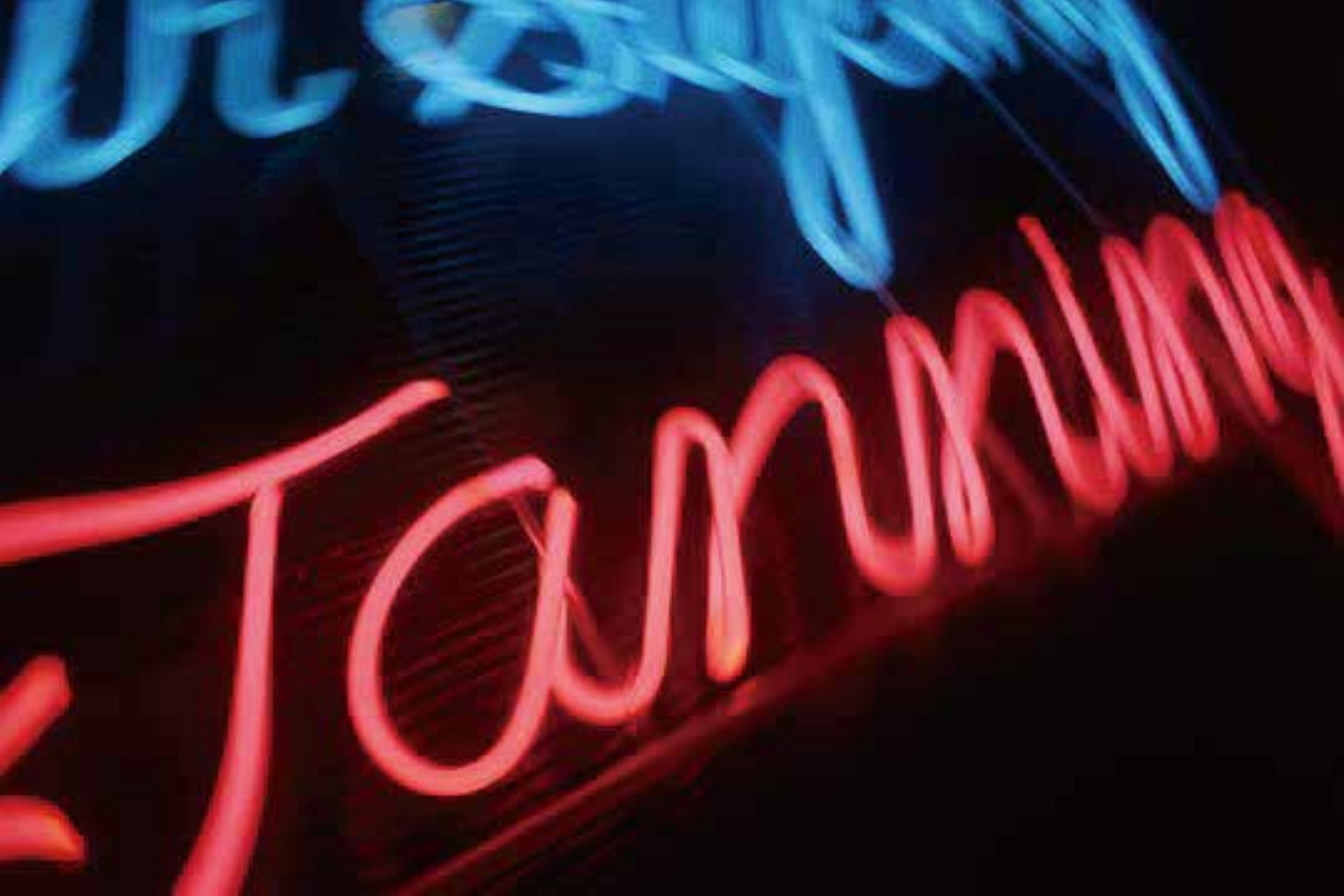}\label{figshakeb}}
\caption{Camera Shaken Blur}
\end{figure}

\emph{Camera shake blur} is induced by camera motion during the exposure period. This is particularly common in handheld photography and in low light situations, e.g. inside buildings or at night. This blur, like object motion blur, can be very complex, since the hands may move in an irregular direction when photographing, causing the camera translation or rotation in-plane or out-of-plane \citep{whyte2010non,whyte2012non}. An ideal situation exists if the camera is only slightly translated when capturing a long distance scene and the resultant blur is approximately spatially invariant, and can be modeled as linear motion blur in equation (\ref{linearmotion}). However this invariance will be violated when the camera undergoes significant translation or rotation during the exposure. In Fig. \ref{figshakea}, for example, the flowers are at different distances from the focal plane, implying that during the camera's translation, the nearer flowers undergo a large shift with respect to the focal plane, while the distant flowers experience a slight shift. Camera rotation is a more complicated case which includes in-plane rotation and out-of-plane rotation in terms of focal plane. In the case of in-plane rotation, the blur kernel varies significantly across the image, especially for the regions far from the axis along which the camera is rotated, as shown in Fig. \ref{figshakeb}. For out-of-plane rotation, the degree of spatial variance across the image is dependent on the focal length of the camera (for more detail see \citep{whyte2012non}). Clearly, the spatially variant blur is more suitable to explain all cases mentioned above than the spatially invariant blur. Fortunately, as \citet{harmeling2010space} point out, it is reasonable to expect that the blur kernel will vary smoothly across the whole image if the depth of the scene varies smoothly, in spite of the fact that this is not true when the objects have very different distances from the focal plane.

\subsubsection{Defocus Blur}

\begin{figure}[t]
\centering
\subfloat[Simulated defocus blur]{\includegraphics[height=1.5in]{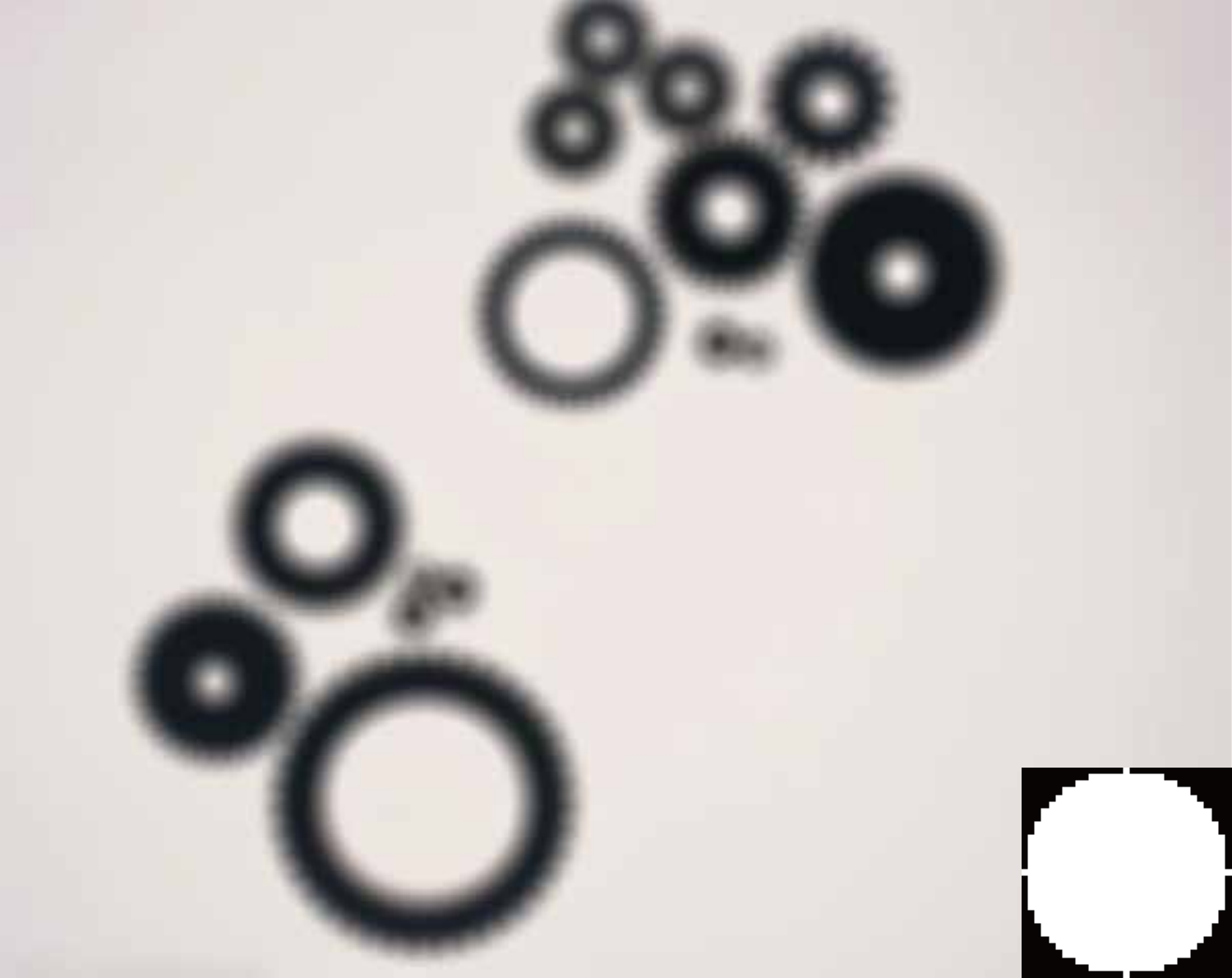}\label{figdefocusa}}
\hspace{0.5cm}
\subfloat[Real defocus blur]{\includegraphics[height=1.5in]{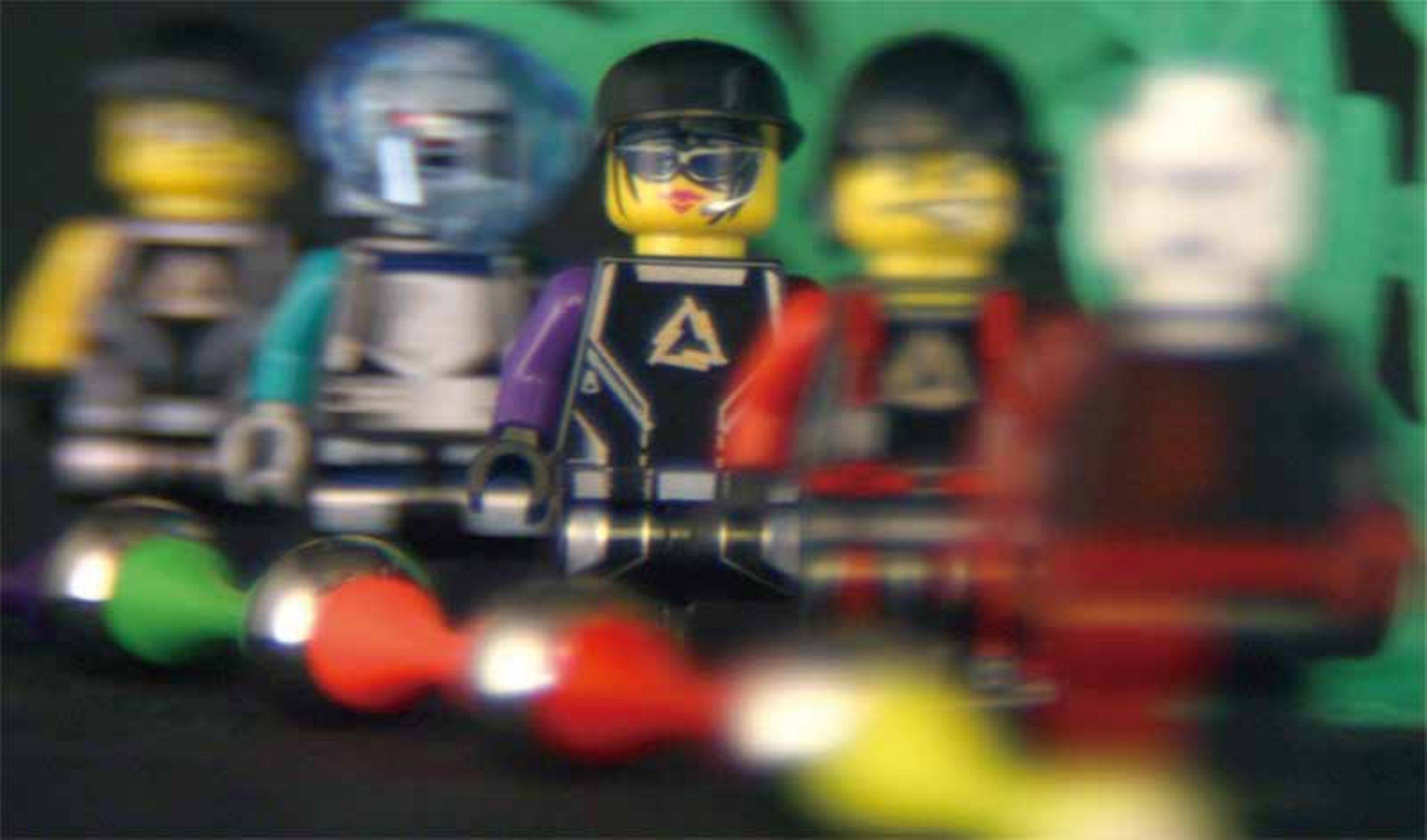}\label{figdefocusb}}
\caption{Defocus Blur}
\end{figure}

As a result of imperfect focusing by the imaging system or different depths of scene, the fields outside the focus field are defocused, giving rise to \emph{defocus blur}, or \emph{out of focus blur}. This blur is familiar in our everyday photos. For example, it is often hard for a photography beginner to focus on the target object by hand. Also, when a camera is equipped with only one lens, scenes outside the \emph{depth of field} (DOF, which is the range covered by all objects in a scene that appear acceptably sharp in an image) are all blurred in the resultant image, e.g. Fig. \ref{figdefocusb}. Traditionally, a crude approximation of a defocus blur is made as a uniform circular model:
\begin{equation}
h(i,j)=
\begin{cases}
\frac{1}{\pi R^2},& \text{if }\sqrt{i^2+j^2}\leq R,\\
0, & \text{otherwise},
\end{cases}
\end{equation}
where $R$ is the radius of the circle. This is valid if the depth of scene does not have significant variation and $R$ is properly selected. A simulated instance is shown in Fig. \ref{figdefocusa}. Practically, focusing on a target object is not difficult for modern consumer cameras since most of them are equipped with an auto-focus function; however, due to the limited DOF, it is not always possible to make the entire image sharp, as illustrated in Fig. \ref{figdefocusb}. To recover a full-focused image, the \emph{focus sweep} technique is usually utilized to sweep the plane of focus through a desired depth range during exposure, so that the depth of field is enlarged \citep{bando2013near}. An alternative way of handling this limit is to use coded aperture pairs \citep{zhou2011coded}, by which the depth of the scene can be also recovered.

\subsubsection{Atmospheric Turbulence Blur}

\begin{figure}[t]
\centering
\subfloat[Simulated blurry image]{\includegraphics[height=1.5in]{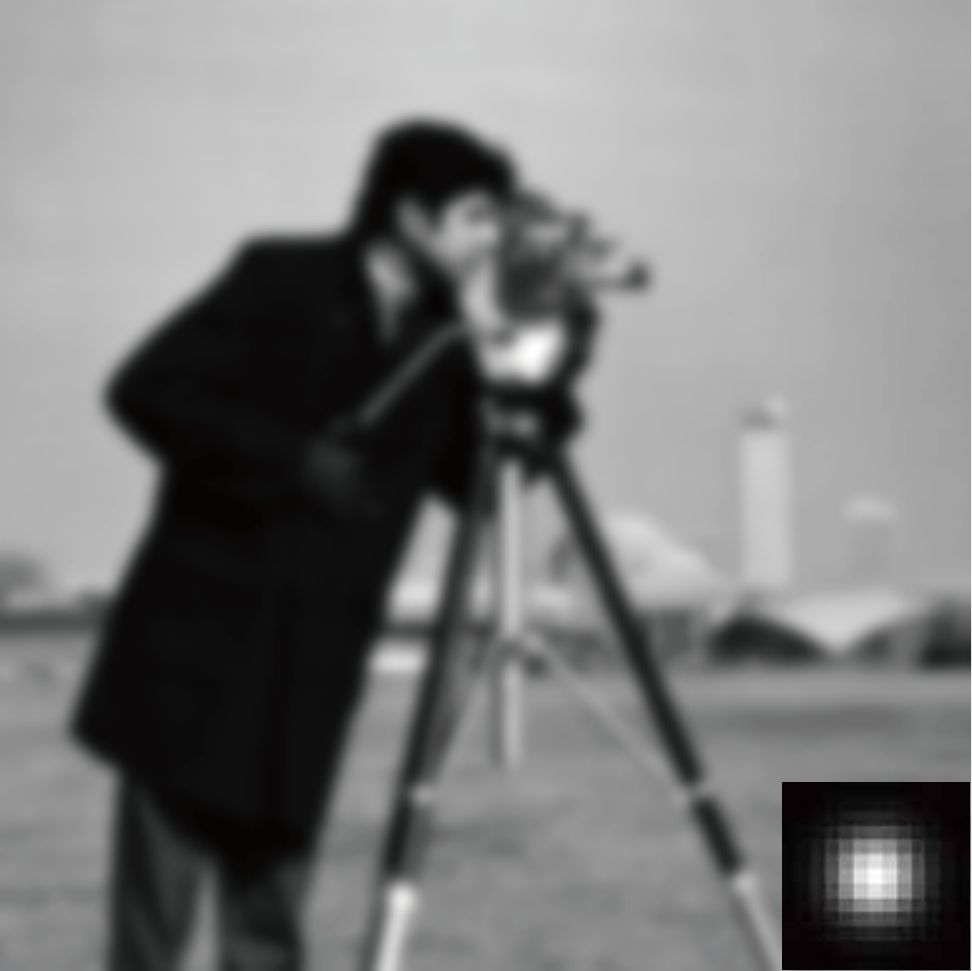}\label{figturba}}
\hspace{0.5cm}
\subfloat[Real blurry image]{\includegraphics[height=1.5in]{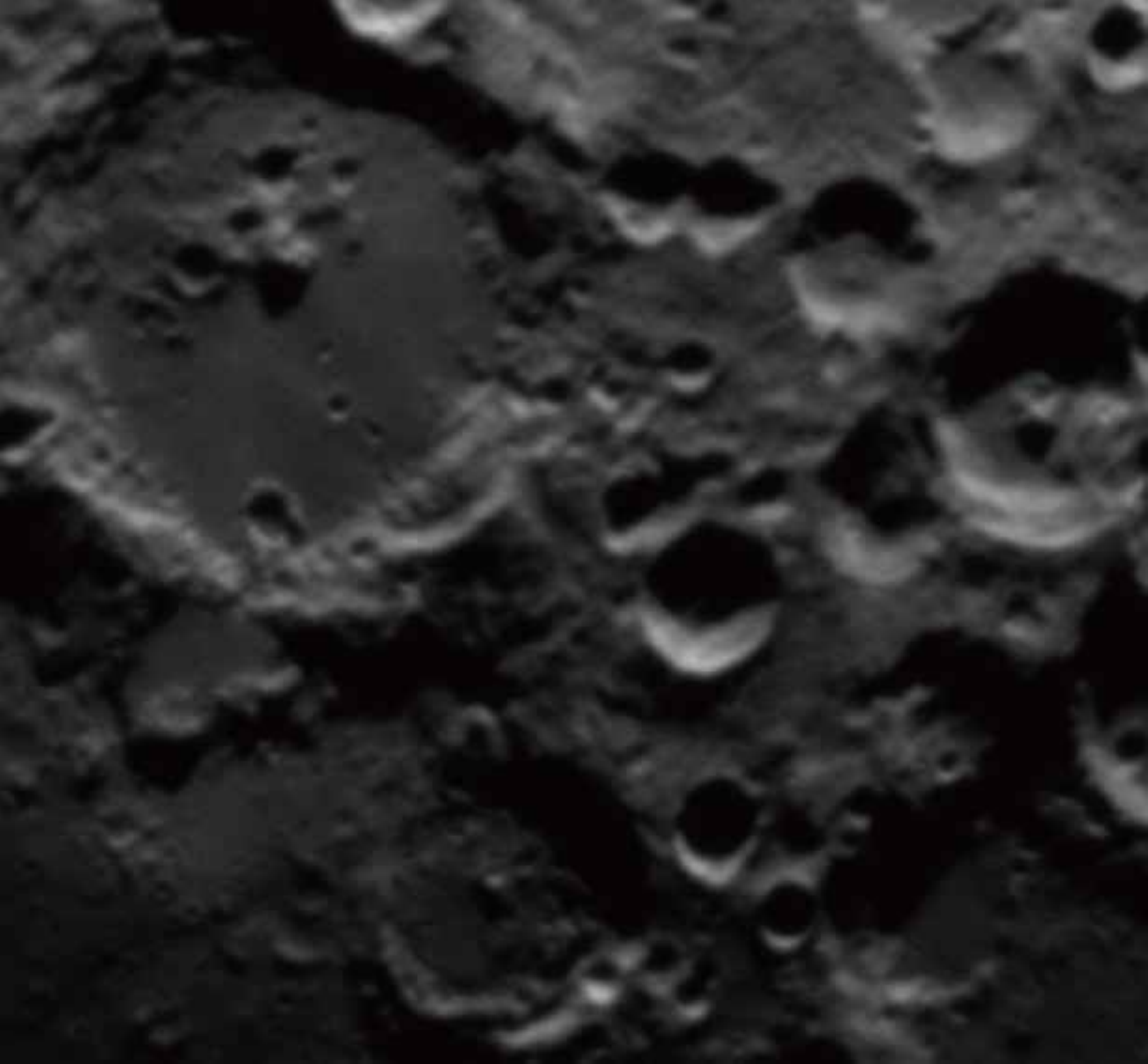}\label{figturbb}}
\caption{Atmospheric Turbulence Blur}
\label{figturb}
\end{figure}

\emph{Atmospheric turbulence blur} generally happens in long-distance imaging systems such as remote sensing and aerial imaging. This is mainly because of the randomly varying refractive index along the optical transmission path. For long-term exposure through the atmosphere, the blur kernel can be described by a fixed Gaussian model \citep{zhu2010image}, i.e.
\begin{equation}\label{gaussturb}
h(i,j)=Z\cdot\exp\left(-\frac{i^2+j^2}{2\sigma^2}\right),
\end{equation}
where $\sigma$ encodes the size of the kernel and $Z$ is the normalizing constant ensuring that the blur has a unit volume. It has been noted that the above equation is impracticable and in fact, this blur is a mixture of multiple degradations like geometric distortion, spatially and temporally variant defocus blur, and possibly motion blur \citep{hirsch2010efficient,zhu2011stabilizing,zhu2013removing}. Fig. \ref{figturb} illustrates a blurry cameraman image degraded by the kernel in (\ref{gaussturb}), as well as a real image of Moon Surface from \citep{zhu2013removing}.

\subsubsection{Intrinsic Physical Blur}

\begin{figure}[t]
\centering
\includegraphics[height=1.5in]{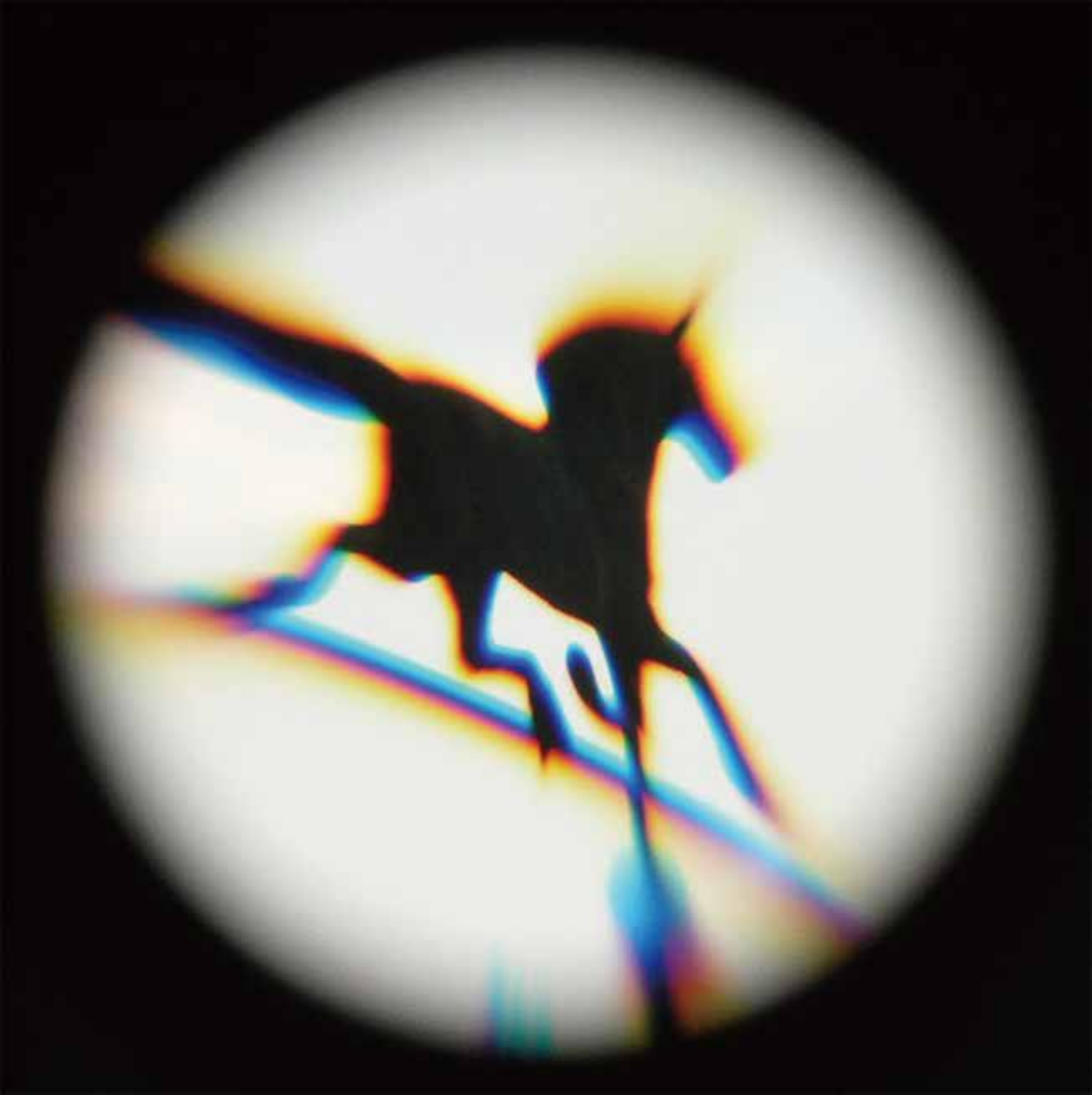}
\caption{Optical Aberration}
\label{chroab}
\end{figure}

\emph{Intrinsic physical blur} is inherent in a number of imaging systems, due largely to intrinsic factors such as light diffraction, lens aberration, sensor resolution, and anti-aliasing filters. An important instance of this type is the optical aberration. In an ideal optical system, all rays of light from a point in the real world will converge to the same point on the focal plane, generating a sharp image. In reality, however, any departure of an optical system from this principle is called an optical aberration. For a real system with spherical optics, it is unrealistic that the light rays from a point source are all parallel with the optical axis, leading to monochromatic aberration, which is a branch of optical aberration. Another branch is chromatic aberration. Since lenses have different refractive indexes for different wavelengths of light, it is difficult to focus all colors on the same convergence point. Generally, the acquired image is corrupted by various optical aberrations, and the induced blur is spatially variant \citep{schuler2012blind}. See Fig. \ref{chroab} for an example. To address this type of blur, additional calibration techniques are introduced to assist the estimation of the blur, such as the utilization of Poisson noise pattern \citep{delbracio2012non}, or the checkerboard test chart \citep{kee2011modeling}. \citet{schuler2011non} proposed that all optical aberrations could be corrected by digital image processing, and \citet{tzeng2010contourlet} exploited the specific property of the fluidic lens camera system in which one of three color planes remains sharp in the imaging process.

\subsection{Effects of the CRF $f(\cdot)$} \label{crfeffects}

Recall that the \emph{camera response function} $f(\cdot)$ in equation (\ref{crfform}) is a nonlinear model mapping the scene irradiance to image intensity. The purpose of this design is to mimic the response of the human visual system or to compress the dynamic range of scenes, or for aesthetic consideration \citep{grossberg2004modeling,lin2005determining}. However, inappropriate processing of CRF leads to severe ringing artifacts in deblurring problems. \citet{chen2012theoretical} have theoretically pointed out three kinds of CRF effects on blur inconsistency $\mathrm{\Delta}=f(\tilde{x}*h)-x*h$. Here $\tilde{x}$ is the sharp irradiance image corresponding to $x$. These effects are:

\noindent 1) The nonlinear CRF has no influence on the intensity of the uniform regions, i.e. $\mathrm{\Delta}=0$;

\noindent 2) In low-frequency regions, the irradiance is approximately equal to its corresponding intensity, i.e. $\mathrm{\Delta}\approx0$, under the condition of small kernel $h$ and smooth function $f(\cdot)$;

\noindent 3) In high-frequency high contrast regions, the nonlinearity of $f(\cdot)$ can significantly damage the blur consistency, particularly when the local minimum and maximum pixel intensities are very different.

\begin{figure}[t]
\centering
\subfloat[Input image]{\includegraphics[width=1.5in]{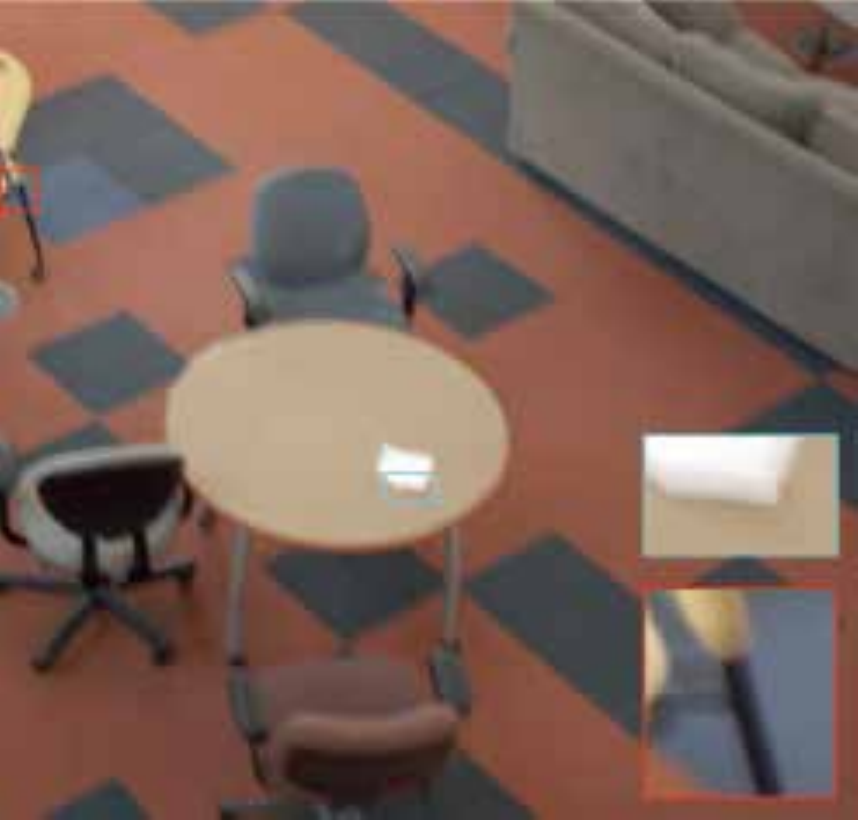}\label{figcrfa}}
\hspace{0.1cm}
\subfloat[Estimated CRF]{\includegraphics[width=1.5in]{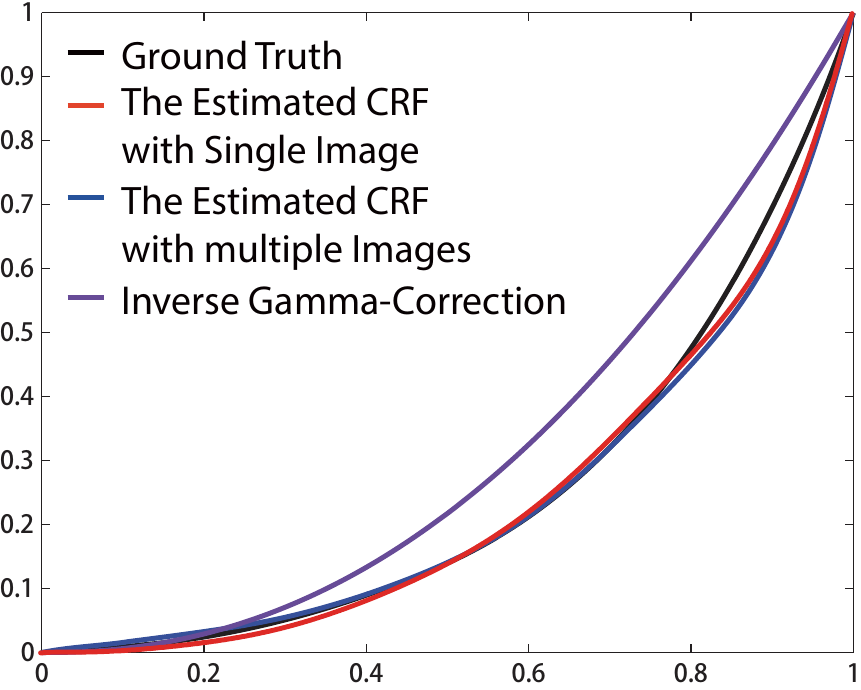}\label{figcrfb}}
\hspace{0.1cm}
\subfloat[Deblurred image without CRF correction]{\includegraphics[width=1.5in]{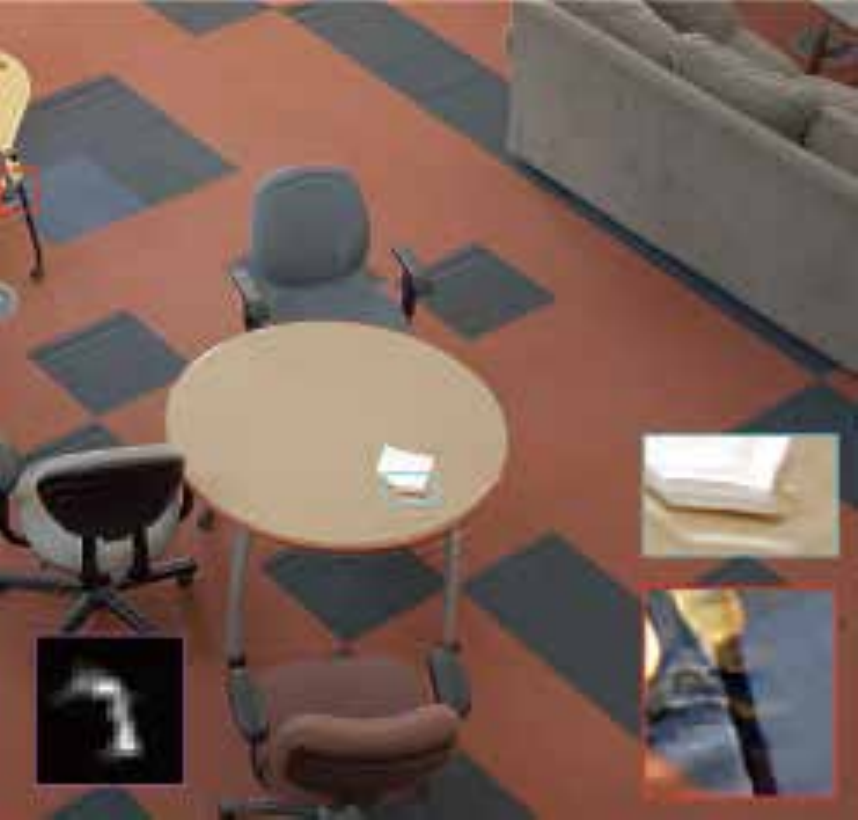}\label{figcrfc}}
\hspace{0.1cm}
\subfloat[Deblurred image with CRF correction]{\includegraphics[width=1.5in]{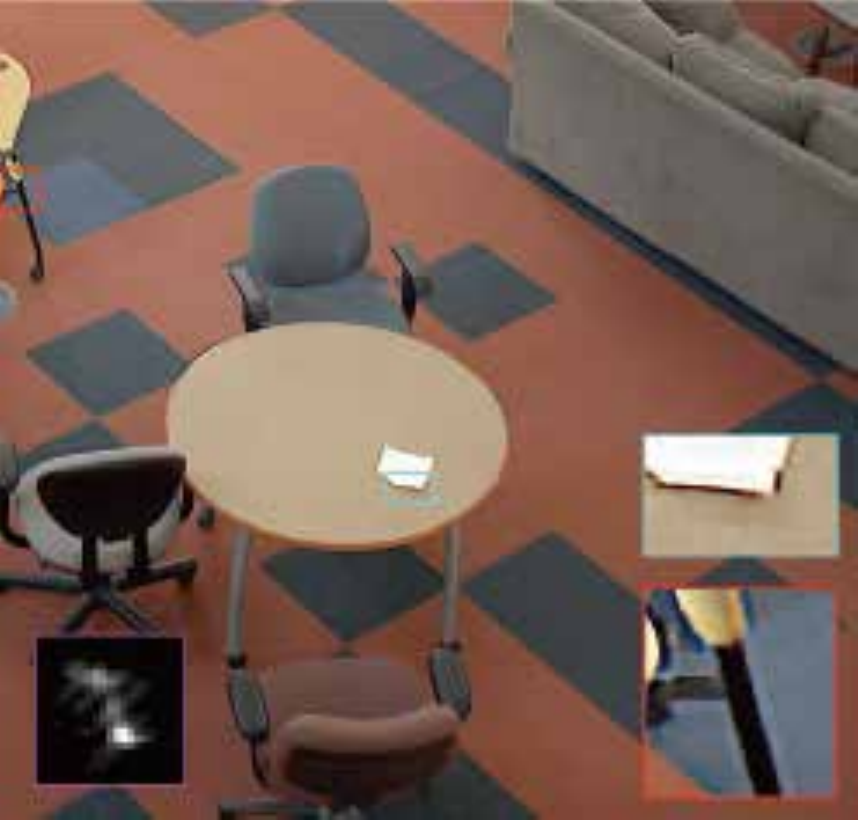}\label{figcrfd}}
\caption{Effects of the nonlinear CRF on deblurring results \citep{kim2012nonlinearsup}}
\label{figcrf}
\end{figure}

From these claims, we readily find that even if a spatially invariant blur $h$ is assumed to be in irradiance, $f(\cdot)$ could turn $h$ into a spatially variant case in the intensity domain \citep{kim2012nonlinear}. To estimate the function, given a blurry/sharp image pair, \citet{chen2012theoretical} exploited Generalized Gamma Curve Model (GGCM) to fit $f(\cdot)$. \citet{kim2012nonlinear} and \citet{tai2013nonlinear} addressed the estimation in two ways, one of which is based on a least-squares formulation when the blur kernel is known, while the other is solved via rank minimization without the need for a known kernel. Fig. \ref{figcrf} illustrates a result of \citep{kim2012nonlinear,kim2012nonlinearsup}, indicating that large number of artifacts remain in the selected regions of the deblurred image without the CRF correction.

In summary, each of the five blur types discussed in this section has specific properties, inspiring researchers to develop effective and efficient algorithms for deblurring. Needless to say, a good deblurring model should be suitable for all blur types which, however, is difficult to be explored. Fortunately, in the deblurring community, there have been many progressive methods proposed in recent years.  In the following sections, we will discuss the five categories of existing methods from the modeling perspective in detail. First let us focus on Bayesian inference framework.

\section{Bayesian Inference Framework} \label{bayesinferframe}

In statistics, Bayesian inference updates the states of a probability hypothesis by exploiting additional evidence. Bayes' rule is the critical foundation of Bayesian inference and can be expressed as
\begin{equation}\label{bayes}
p(A|B)=\frac{p(B|A)p(A)}{p(B)},
\end{equation}
where $A$ stands for the hypothesis set and $B$ corresponds to the evidence set. This rule states that the true posterior probability $p(A|B)$ is based on our prior knowledge of the problem, i.e. $p(A)$, and is updated according to the compatibility of the evidence and the given hypothesis, i.e. the likelihood $p(B|A)$. In our scenario for the non-blind deblurring problem, $A$ is then the underlying sharp image $x$ to be estimated, while $B$ denotes the blurry observation $y$. For the blind case, a slight difference is that $A$ means the pair of $(x,h)$ since $h$ is also a hypothesis in which we are interested. Explicitly (\ref{bayes}) can be written for both cases as
\begin{eqnarray}
\text{Non-blind:}&&p(x|y,h)=\frac{p(y|x,h)p(x)}{p(y)},\label{nonblindmap}\\
\text{Blind:}&&p(x,h|y)=\frac{p(y|x,h)p(x)p(h)}{p(y)}.\label{blindmap}
\end{eqnarray}
Note that either $x$ or $y$ and $h$  are usually assumed to be uncorrelated. Irrespective of case, the likelihood $p(y|x,h)$ is dependent on the noise assumption, as listed in Table \ref{noisetable}. How to further infer the equation (\ref{bayes}) in the literatures inspires us to explore three directions: maximum a posteriori, minimum mean square error, and variational Bayesian methods.

\subsection{Maximum a Posteriori} \label{MAP}

\begin{figure*}[t]
\centering
\includegraphics[width=5.6in]{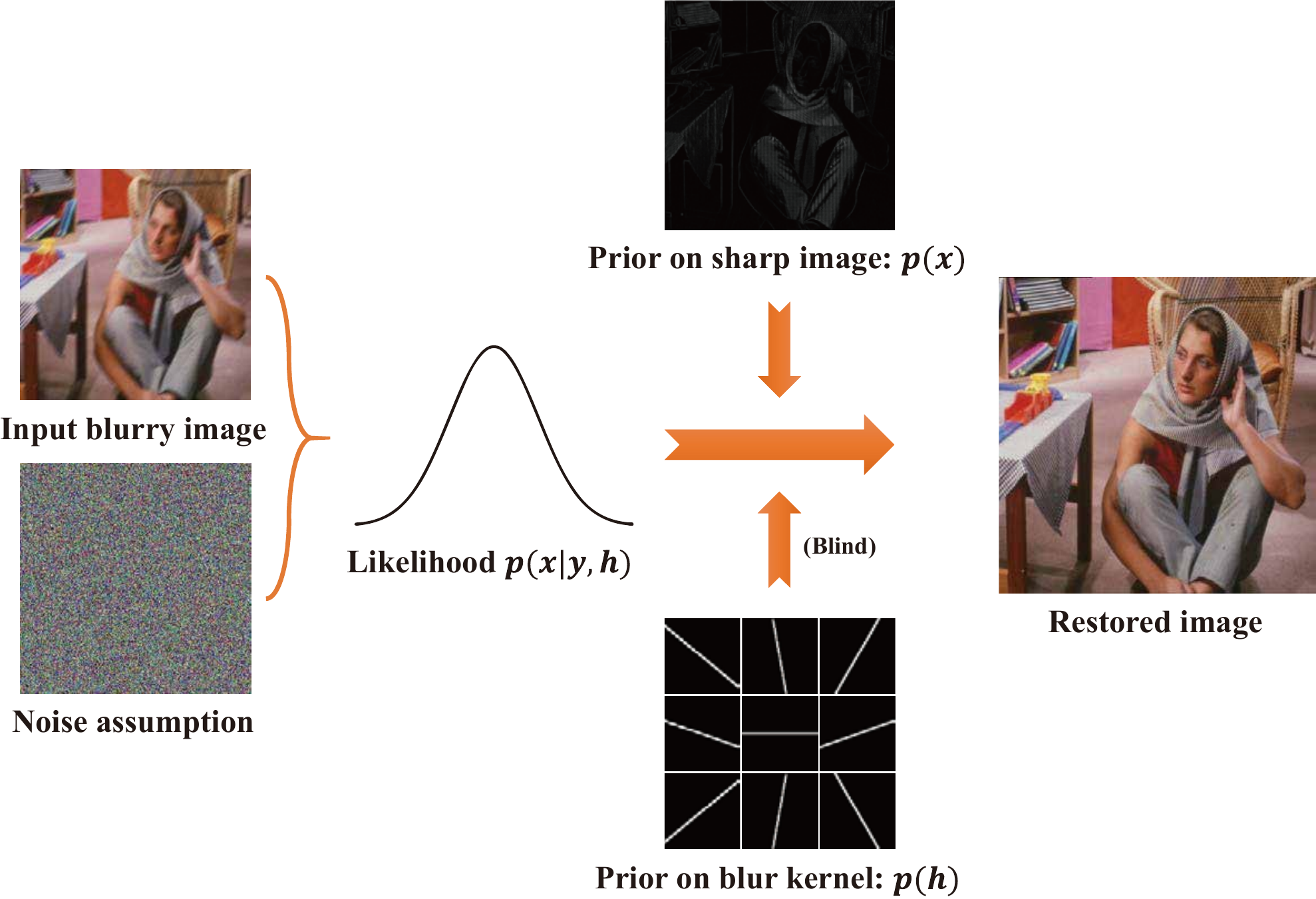}
\caption{MAP framework for image deblurring}
\label{MAPframefig}
\end{figure*}

The most commonly-used estimator in a Bayesian inference framework is the \emph{maximum a posteriori} (MAP). This strategy attempts to find the optimal solution $A^*$ which maximizes the distribution of the hypothesis set $A$ given the evidence set $B$ in (\ref{bayes}). In the blind case,
\begin{eqnarray}\label{MAPXH}
(x^*,h^*)&=&\mathop{\argmax}\limits_{x,h}p(x,h|y)\nonumber\\
&=&\mathop{\argmax}\limits_{x,h}p(y|x,h)p(x)p(h),
\end{eqnarray}
while in the non-blind scenario, the term $p(h)$ is discarded according to equation (\ref{nonblindmap}). Fig. \ref{MAPframefig} shows a diagram of this type of method.

\subsubsection{Maximum Likelihood Estimation} \label{maximumlikelihood}

We start with the introduction of a classic non-blind algorithm, Richardson-Lucy (RL) deconvolution \citep{richardson1972bayesian,lucy1974iterative}, which is widely used in astronomical imaging and medical imaging.\footnote{In the standard RL, no specific noise assumption is assumed. Here, we introduce RL from the MLE perspective under Poisson noise assumption to facilitate the ongoing analysis.} Assuming that the prior $p(x)$ takes the form of a uniform distribution, the MAP estimator then becomes a maximum likelihood estimator (MLE), i.e.
\begin{eqnarray} \label{eqmle}
x^*&=&\mathop{\argmax}\limits_x p(y|x,h)\nonumber\\
&=&\mathop{\argmin}\limits_x -\log p(y|x,h),
\end{eqnarray}
where in the second equation the minimized equation is known as negative log-likelihood. Under the noise assumption of Poisson distribution \citep{shepp1982maximum}, (\ref{eqmle}) is derived as
\begin{equation}
x^*=\mathop{\argmin}\limits_x \sum_i \left((x*h)_i-y_i\log(x*h)_i\right).
\end{equation}
Taking the derivative with respect to $x$ and setting it to zero, we can get
\begin{equation}
\left(\mathbf{1}-\frac{y}{x*h}\right)*h_{-}=\mathbf{0}, \label{MLderi}
\end{equation}
where $\frac{y}{x*h}$ denotes the point-wise division, $\mathbf{1}$ is the spatially invariant function which is one everywhere, and $h_-$ is the symmetrical reflection of $h$, meaning $h_-(i,j)=h(-i,-j)$. From the unit property of $h_-$, i.e., $\mathbf{1}*h_-=\mathbf{1}$, (\ref{MLderi}) can be written as
\begin{equation}
\frac{y}{x*h}*h_-=\mathbf{1}.
\end{equation}
Multiplying both sides by $x$ and utilizing the \emph{Banach fixed-point} theorem yields
\begin{equation}\label{RLiteration}
x^{t+1}=x^t\odot\left[\left(\frac{y}{x^t*h}\right)*h_-\right],
\end{equation}
where $\odot$ is point-wise multiplication. The above RL deconvolution procedure produces a sequence of estimations $x^t$ and eventually converges to the optimal solution $x^*$. If we take the blind case into consideration, (\ref{RLiteration}) should be incorporated with an additional step for the estimation of $h$ in each iteration:
\begin{eqnarray}
x^{t+1}&=&x^t\odot\left[\left(\frac{y}{x^t*h^t}\right)*h_-^t\right],\\
h^{t+1}&=&\frac{h^t}{\mathbf{1}*x_-^t}\odot\left[\left(\frac{y}{x^t*h^t}\right)*x_-^t\right].
\end{eqnarray}
Note that this iterative process is not guaranteed to converge to the global solution since $\min_{(x,h)} -\log p(y|x,$ $h)$ is not convex, and a small error in $h$ will lead to significant artifacts in the resultant images. To handle this ill-posedness, the penalized MLE has been introduced:
\begin{eqnarray}\label{penalizeMLE}
(x^*,h^*)=\mathop{\argmin}\limits_{x,h}&&\sum_i \left((x*h)_i-y_i\log(x*h)_i\right)\nonumber\\
&&+\lambda_x\mathrm{\Psi}_1(x)+\lambda_h\mathrm{\Psi}_2(h),
\end{eqnarray}
where $\mathrm{\Psi}_1(\cdot)$ and $\mathrm{\Psi}_1(\cdot)$ are the penalty functions on $x$ and $h$, respectively. Direct derivation leads to the iterations for the penalized version, i.e.,
\begin{eqnarray}
x^{t+1}&=&x^t\odot\frac{\left(\frac{y}{x^t*h^t}\right)*h_-^t}{\mathbf{1}+\lambda_x\frac{\partial\mathrm{\Psi}_1(x^t)}{\partial x^t}},\\
h^{t+1}&=&h^t\odot\frac{\left(\frac{y}{x^t*h^t}\right)*x_-^t}{\mathbf{1}*x_-^t+\lambda_h\frac{\partial\mathrm{\Psi}_2(h^t)}{\partial h^t}}.
\end{eqnarray}

Under this formula, \citet{temerinac2012multiview} utilized the total variation (TV) regularization for $\mathrm{\Psi}_1(x)$ and derived a multi-view RL algorithm. In their method, the underlying sharp 3D image is composed of images observed from multiple viewpoints, each of which is corrupted by a different blur kernel. The multiple blurry images are integrated in a unified formulation from which the sharp image is jointly deduced. \citet{lefkimmiatis2013poisson} employed the Schatten norms of Hessian matrix on each image pixel as the regularization on $x$. This kind of regularizer is based on the second-order derivatives instead of the first-order derivatives, thus favoring piecewise-smooth solutions, as opposed to TV which produces piecewise-constant solutions.

Regarding $\mathrm{\Psi}_2(h)$, \citet{keuper2013blind} discovered a specific property of the wide-field fluorescence microscopy (WFFM) system, which suggests that the \emph{optical transfer function} (OTF), i.e. the Fourier transform of the kernel $h$, is well localized and smooth. They proposed imposing the TV constraint on both $x$ to preserve edges, and OTF to ensure the smoothness of the kernel. \citet{kenig2010blind} developed a novel approach to restrict $h$ to a kernel subspace which was produced by either linear or kernel \emph{principal component analysis} (PCA) based on the general forms of the WFFM PSF. This is easily integrated into the iterative RL deconvolultion procedure. Additionally, the residual denoising operation is employed to avoid over-smoothing the useful high-frequency details, which is also used in \citep{keuper2013blind}.

Most of the above methods are based on the assumption of a spatially invariant kernel. \citet{tai2011richardson} developed the projective motion RL algorithm to tackle the spatially variant case. Under the proposed projective motion blur model, the operations of convolution and correlation in the conventional RL algorithm are replaced by a sequence of forward projective motions and their inverses via \emph{homographies}, which will be discussed in Section \ref{homographymethod}. Various regularizers are applicable in their framework.

\subsubsection{Priors for MAP}

The penalized MLE in equation (\ref{penalizeMLE}) is exactly the MAP since the penalty functions act as the priors in that $p(x)=\exp(-\lambda_x\mathrm{\Psi}_1(x))$ and $p(h)=\exp(-\lambda_h\mathrm{\Psi}_2(h))$. Various MAP based methods focus on the development of priors to obtain attractive deblurring results.

It is commonly agreed that for a natural image, the gradients of the sharp image tend to obey a heavy-tailed distribution, meaning that the distribution of gradients has most of its mass on small values but assigns significantly more probability to large values than a Gaussian distribution \citep{field1994goal,fergus2006removing}. Thus natural images often contain large regions of constant intensity or gentle intensity whose gradients are interrupted by occasional large changes at the edges or occlusion boundaries. To express such a prior, various approximations of the heavy-tailed distribution are used to describe the gradients. One classic method is to inform a Laplace distribution on the magnitude of the gradients:
\begin{equation}
p^{Lap}(\nabla x)=\prod_i\frac{1}{2b}\exp\left(-\frac{\|\nabla x_i\|_1}{b}\right),
\end{equation}
where $\nabla$ is the gradient operator, $\|\cdot\|_1$ is the $\ell_1$ norm, and $b$ denotes the scale parameter. Additionally, for computation efficiency, a generalized Gaussian model is used by \citet{levin2007image} in their design of an optimal aperture filter, and a Gaussian prior is imposed on the gradient patches instead of gradient pixels by \citet{hu2012psf}. The autocorrelation function of the blur kernel is shown in relation to the covariance matrix of the Gaussian distribution. Through the Fourier transform of the autocorrelation matrix and an additional phase retrieval stage, the blur kernel can be easily recovered. Unlike the MAP methods involving repeated reconstructions of the sharp image, this approach directly relies on basic statistics of the blurry image and is therefore efficient.

Due to the imperfect fitness of the above priors to the real gradient distribution of sharp natural images, these methods tend to remove mid-frequency textures, even though structures such as edges can be preserved. To improve the fitness to the heavy-tailed distribution, \citet{fergus2006removing} proposed using a \emph{Gaussian mixture model} (GMM) having finite mixture numbers. \citet{chakrabarti2010analyzing} extended this to Gaussian scale mixture (GSM) which is a mixture of infinite Gaussian models with a continuous range of variances. The form of the corresponding zero-mean GMM and GSM are respectively
\begin{eqnarray}
p^{GMM}(\nabla x)&=&\prod_i\sum_{c=1}^C \mathcal{N}(\nabla x_i|0,\xi_c),\\
p^{GSM}(\nabla x)&=&\prod_i\int_{\xi}\mathcal{N}(\nabla x_i|0,\xi)p(\xi)d\xi, \label{GSM}
\end{eqnarray}
where $\xi$ is the standard derivation of the Gaussian distribution, $c$ and $C$ in GMM is the index and the total number of mixtures, and $p(\xi)$ in GSM is a probability distribution on $\xi$. In terms of GSM, a critical issue is the infinite selection of $\xi$, which makes it computationally expensive. According to the theoretical analysis by \citet{palmer2005variational}, equation (\ref{GSM}) can be derived from the variational perspective as
\begin{equation}\label{adapGSM}
p(\nabla x)=\prod_i\sup_{\xi_i>0}\mathcal{N}(\nabla x_i|0,\xi_i)p(\xi_i),
\end{equation}
where $p(\xi)$ takes the form of $\exp(f(\frac{\xi}{2}))$. This theoretical convenience is directly utilized by \citep{chakrabarti2010analyzing} and further investigated by \citet{zhang2013multi,zhang2013non}. In \citep{zhang2013multi}, authors proposed an adaptive sparse prior based on (\ref{adapGSM}). They coupled multiple blurry observations in a joint deblurring procedure by assuming that the parameters $\{\xi_i\}$ in (\ref{adapGSM}) are shared across all images. In their subsequent work \citep{zhang2013non}, the idea of the shared parameter is extended to the single image deblurring problem by additionally assuming the homography expression of blur kernels. In both works, the sparse regularizer is adaptively adjusted according to the noise level estimated in each iteration.

Another question raises: even though the assumed prior coincides perfectly with the gradient distribution of the sharp image, is the recovered gradient distribution able to fit our prior as we expected? A recent research by \citet{cho2012image} responded in the negative. According to their analysis, the above heavy-tailed priors are generally independently forced on each pixel or each local patch, failing to capture the global statistics of gradients. Thus the reconstructed image favors flat regions, resulting in the gradient distribution of the MAP estimates failing to match that of the original sharp image. To address this issue, the authors proposed penalizing the \emph{Kullback-Leibler} (KL) divergence between the gradient distribution of the estimated image $p_e(\nabla x)$ and a reference distribution $p_r(\nabla x)$,
\begin{equation}
KL(p_e||p_r)=\int_{\nabla x}p_e(\nabla x)\ln\left(\frac{p_e(\nabla x)}{p_r(\nabla x)}\right)d\nabla x,
\end{equation}
which induces a constraint and can be combined with (\ref{MAPXH}). The reference distribution is characterized as the generalized Gaussian model and immediately estimated from the blurry image by using the approach of \citep{cho2010content}. Taking the same idea, \citet{zhuo2010robust} presented a more direct method which forces the gradients of the deblurred image close to those of the reference image. It is direct because: 1) a \emph{Lorentzian} error norm is imposed on the difference of the two gradients rather than the difference of their distributions, i.e.
\begin{equation}
\rho(\nabla x_e, \nabla x_r)=\log\left(1+\frac{1}{2}\left(\frac{\nabla x_e-\nabla x_r}{\varrho}\right)^2\right),
\end{equation}
where $\varrho$ is a predefined constant; and 2) the reference image is a flash image of the same scene which is sharp but corrupted by a degraded illumination.

Besides the priors on image gradients, the knowledge on image intensities or transformed domains is extremely helpful in specific applications. For example, \citet{chen2011effective} developed a content-aware prior for document image deblurring in which the histogram of the whole image is the weighted sum of the histogram of the foreground (dark pixels) and of the background (white pixels). Additionally, a local upper-bound constraint together with TV is imposed to restrain the artifacts in the resultant images. Similarly, \citet{cho2012text} excavated the specific properties of text images which were then used in the deblurring process. In both methods, the domain-specific knowledge is incorporated into the optimization by employing the variable splitting techniques. \citet{shaked2011iterative} saddled the generalized Gaussian prior on the representing coefficients in a linear transformed domain and then derived an efficient algorithm to solve the corresponding MAP problem.

As with the PCA technique used in RL deconvolution \citep{kenig2010blind}, subspace techniques have recently attracted researchers' attention. The above-mentioned priors express general aspects of human knowledge about natural images. As noted, however, natural images are composed of repetitive local patterns, and some classes of images, such as faces, even lie on a subspace. Discovering this information can help to develop novel priors under the MAP framework. \citet{joshi2010personal} proposed a specific restoration problem for personal photo images. Under the MAP framework, they constrained the target face image close to the space expressed by eigen-faces and mean-face generated from the photos of the same person. Benefitting from this space constraint and a sparse gradient constraint, more facial details can be recovered in the resultant photo, while the artifacts are well suppressed. A more general example is for natural images, where the extracted patches can be constrained to be close to a low dimensional manifold \citep{peyre2009manifold}. Based on this observation, \citet{ni2011example} derived a non-blind deblurring method based on a manifold prior learnt from databases. The benefit of these subspace-based priors is the incorporation of more collaboration between the sharp and blurry images, rather than only imposing human heuristic assumptions. This idea is also the key in hardware modification-based methods, as well as learning-based methods, which will be discussed in Section \ref{futuredirect}.

\subsubsection{Edge Emphasizing Operation}

In the MAP framework, an auxiliary operation is usually employed to produce promising deblurring results; that is, the \emph{edge emphasizing operation}. Typically, the aim of the edge emphasizing operation is to detect and restore the large-scale step edges which generally occur when the blurred edges drift far away from the latent sharp edges, meaning that the blur kernel is large. These operations include the shock filter \citep{osher1990feature}, the fuzzy operator \citep{russo1992fuzzy}, the morphological filtering \citep{schavemaker2000image}, the forward-and-backward diffusion process \citep{gilboa2002forward}, and the recently proposed edge prediction techniques \citep{joshi2008psf,cho2011blur}. However, these methods often fail when narrow edges or highly textured regions appear in the image, since these patterns exhibit a wide spread of edges in the blurring process. Edge emphasizing operations manipulates over local structures in an image, and thus occasional noise can significantly influence the performance of these operations \citep{wang2012analyzing,wang2013nonedge,faramarzi2013unified}.

To handle this issue, \citet{wang2012analyzing} conducted both theoretical and empirical analyses on the relationship between edge emphasizing operations and MAP estimators. They showed that the advantages of MAP could compensate for the drawbacks of edge emphasizing operations because MAP depends on image statistics which cannot be affected by local structure variations and noises. Nevertheless, the MAP estimator is based on specific blur models and lacks generalization with respect to different models. Fortunately, edge emphasizing operations can address various blur models without any adaptation. Therefore, incorporating the edge emphasizing operation into the iterative MAP estimation can remedy the limitations of both types of method, resulting in improved performance \citep{cho2009fast}. In Wang et al.'s work, the large-scale step edges are recovered by pre-smoothing and shock filter. The narrow edges are then restored by proposing a strongness-aware prior for the MAP scheme, measuring the strongness of local structures. The deblurring procedure is iterated between the large-scale step edge sharpening and MAP estimation. \citet{faramarzi2013unified} also proved that the edge-emphasizing smoothing operation is beneficial for an accurate estimation of blur kernels. They employed the method of \citet{xu2011image} to simultaneously sharpen the edges and remove large numbers of low-amplitude structures via the $\ell_0$ gradient minimization. Alternatively, \citet{cho2011blur} showed that the Radon projections of a blur kernel can be derived from the edges detected in the blurry image, and the kernel can be recovered by enough Radon projections. As a result, a constraint on the blur kernel expressed by Radon projections is added to the likelihood $p(y|x,h)$, forming the so-called RadonMAP. \citet{almeida2010blind} instead described the edge emphasizing operation as a prior modeled as a response of a set of edge detectors in different directions. They imposed a sparse assumption on this prior, favoring piece-wise constant image estimates. To avoid over-smoothing, the operation of gradually decreasing the regularization parameter was used to produce a promising result. \citet{xu2010two} pointed out that not all strong edges are profitable for kernel estimation. They proposed a new metric to measure the usefulness of the image edges. The selected edges according to this metric are then used to induce the kernel formation. What follows is an adaptive kernel refinement procedure which is to ensure the sparseness of the estimated kernel without damaging its large-valued elements.

\subsubsection{Marginalization Techniques}

Even though appropriate priors or suitable edge emphasizing operations have been utilized in the MAP framework and have resulted in improved performance, intrinsic problems remain. A recent outstanding research by \citet{levin2009understanding,levin2011understanding} comprehensively analyzes the failure of the MAP scheme and points out how to make the MAP estimation successfully recover the true blur kernel. The $\text{MAP}_{x,h}$ scheme for blind deconvolution in (\ref{MAPXH}) can be rewritten as
\begin{equation}\label{MAPfail}
(x^*,h^*)=\mathop{\argmin}\limits_{x,h}\|y-x*h\|^2+\lambda(\sum_i|\nabla_h x_i|^\alpha+|\nabla_v x_i|^\alpha)
\end{equation}
under the Gaussian noise assumption and a sparse derivative prior or heavy-tailed prior on image gradients. $\nabla_h$ and $\nabla_v$ are horizontal and vertical derivatives of the image. According to Levin et al.'s conclusions, the solution of (\ref{MAPfail}) under the sparse prior usually favors a blurry result rather than a sharp result, even though y is generated up to infinitely large image samples from the perfect prior. This observation is also known as the \emph{no-blur explanation} or \emph{delta effect} of MAP, which means that the solution of (\ref{MAPfail}) has a higher probability of being
\begin{equation}
\begin{cases}
x^*&=y\\
h^*&=\delta
\end{cases}
\end{equation}
than the true solution. From the estimation theory perspective, it is evident that we cannot gather sufficient measurements for the $\text{MAP}_{x,h}$ problem since the number of unknown variables grows with the image size, even to infinity.

As noted by \citet{levin2009understanding,levin2011understanding}, the strong asymmetry between the dimensionality of $x$ and $h$ provides a favorable property for handling the blind deconvolution. This means that while the dimensionality of $x$ increases with the image size, the support of $h$ remains fixed and is small relative to the image size. From this viewpoint, $h$ can achieve an increased number of measurements when the image size becomes large. Thus, estimation theory tells us through sufficient measurements on $h$ that the recovered blur kernel under $\text{MAP}_h$ can be arbitrarily close to the true kernel. Mathematically, the $\text{MAP}_h$ is
\begin{eqnarray}\label{MAPmargin}
h^*&=&\mathop{\argmax}\limits_h p(h|y)\nonumber\\
&=&\mathop{\argmax}\limits_h \int p(x,h|y)dx,
\end{eqnarray}
where $h^*$ is the true kernel, as stated in Claim 3 of \citep{levin2011understanding}. Once the kernel is estimated, $x$ can then be solved in a non-blind deblurring scheme. In their subsequent work \citep{levin2011efficient}, Levin et al. noted that $\text{MAP}_h$ is generally complex and hard to directly compute because the marginalization in (\ref{MAPmargin}) involves all possible $x$ explanations, which is computationally intractable. An approximation method was proposed to derive the $\text{MAP}_h$. To estimate the blur kernel, they assumed the \emph{i.i.d.} Gaussian imaging noise and the GMM prior on image derivatives, as well as a uniform distribution on $h$. Equation (\ref{MAPmargin}) can then be written as
\begin{eqnarray}
h^*&=&\mathop{\argmax}\limits_h p(y|h)\nonumber\\
&=&\mathop{\argmax}\limits_h \int p(x,y|h)dx.
\end{eqnarray}
The above problem is solved by the \emph{expectation}-\emph{maximization} (EM) framework that alternates between the estimation of $p(x|y,h)$ which is still a Gaussian (E-step), and the computation of $h$ under the minimum mean square error (M-step). In the E-step, however, calculating the mean and covariance of $p(x|y,h)$ under a sparse prior is generally hard, so the authors proposed approximating the conditional distribution by using variational inference, which will be discussed in Section \ref{VBM}.

\citet{wang2013nonedge} have discovered several intrinsic issues between edge emphasizing operations and image statistics through a large number of experiments on ImageNet \citep{deng2009imagenet} composed of 1.2 million images in total. Their research points out that the limited number of large scale step edges within a natural image cannot ensure a robust estimation of the blur kernel. Additionally, due to the diversity of natural images, the sparse derivative priors are not consistent across them and it is almost impossible to find a robust measurement that favors sharp explanations for all of them. Different from their previous work \citep{wang2012analyzing} which uses $\text{MAP}_{x,h}$, they adopted the marginalization scheme in (\ref{MAPmargin}) and developed an adaptive sparse prior composed of two components to ensure robustness. The first component is the commonly-used sparse derivative prior, while the second encodes the edge emphasizing operation.

\subsection{Minimum Mean Square Error} \label{MMSE}

The Bayesian framework aims to estimate $x$ or $h$ from the posterior $p(x,h|y)$ and a loss function $L((x^*,h^*),(x,$ $h))$. The expected loss is computed over all unknown variables,
\begin{equation}\label{Bayesexp}
\tilde{L}((x^*,h^*),(x,h)|y)=\int L((x^*,h^*),(x,h))p(x,h|y)dxdh,
\end{equation}
which is called Bayesian expected loss \citep{brainard1997bayesian}. The optimal solution $(x^*,h^*)$ is then chosen to minimize $\tilde{L}((x^*,h^*),(x,h))$. If we take $L((x^*,$ $h^*),(x,h))$ as a Dirac delta loss function, i.e. $L((x^*,h^*),$ $(x,h))=1-\delta((x^*,h^*)-(x,h))$, (\ref{Bayesexp}) becomes the $\text{MAP}_{x,h}$ problem. Alternatively, if $L((x^*,h^*),(x,h))$ is the square error loss, we can obtain the \emph{minimum mean square error} (MMSE) formulation:
\begin{eqnarray}\label{MMSEXH}
(x^*,h^*)&=&\mathop{\argmin}\limits_{\hat{x},\hat{h}}\int \|\hat{x}-x\|^2\|\hat{h}-h\|^2p(x,h|y)dxdh\nonumber\\
&=&\mathop{\argmin}\limits_{\hat{x},\hat{h}}\mathbb{E}\{x,h|y\}.
\end{eqnarray}
For the non-blind deblurring case, the above equation turns into
\begin{eqnarray}\label{MMSEX}
x^*&=&\mathop{\argmin}\limits_{\hat{x}}\int \|\hat{x}-x\|^2p(x|y,h)dx\nonumber\\
&=&\mathop{\argmin}\limits_{\hat{x}}\mathbb{E}\{x|y,h\}.
\end{eqnarray}

The equivalence between MMSE and MAP has been proved by \citet{levin2011understanding}, that is \emph{if $p(h|y)$ has a unique maxima, then for large images, the $\text{MAP}_{h}$ estimator followed by an $\text{MMSE}_{x}$ image estimation is equivalent to a simultaneous $\text{MMSE}_{x,h}$ estimation of both $x$ and $h$.}

In spite of this, the empirical demonstrations on image denoising have shown the advantages of MMSE over MAP approaches \citep{schmidt2010generative}. MAP solutions usually exhibit piecewise constant regions and result in incorrect statistics of the output image, as previously noted, whereas MMSE can achieve the desired statistics by exploiting the uncertainty of the model. Furthermore, the image restoration performance of the MMSE estimator is highly correlated with the generative quality of the model. This observation is particularly useful since MMSE benefits from a powerful learnt generative model even without any regularization weight. As is well known, the regularization parameter is related to the noise level of the degraded image. By taking the above superiority of MMSE, \citet{schmidt2011bayesian} integrated the noise estimation process into the MMSE framework by treating the noise standard deviation as a variable of the posterior, i.e., given $h$ and $y$,
\begin{equation}
p(x|y,h)=\int p(x,\sigma|y,h)d\sigma.
\end{equation}

Compared with MAP, one problem in MMSE remains. Due to the lack of complete knowledge on the joint distribution ($p(x,h|y)$ and $p(x|y,h)$) in real applications, it is difficult to take expectation in (\ref{MMSEXH}) and (\ref{MMSEX}) over all possible explanations. To handle this problem, \citet{schmidt2010generative,schmidt2011bayesian} proposed to use the Gibbs sampling method to alternatively generate the sequence of the variable samples, e.g. in deblurring $\{(x^1,z^1,\sigma^1),...,(x^T,z^T,\sigma^T)\}$ where $z$ is a latent variable. Another way to address the above issue is to abandon the full optimality requirements and use a particular class of estimators to approximate MMSE, such as linear MMSE:
\begin{equation}\label{linearMMSE}
\mathbb{E}\{A|B\}=WB+v,
\end{equation}
where $A$ and $B$ are as those in (\ref{bayes}), and $W$ and $v$ are the parameters encoding the deblurring process. Minimizing $\mathbb{E}\{\|A-WB-v\|^2\}$ with respect to $W$ and $v$ can result in the optimal $W$ and $v$ expressed as
\begin{equation}
\begin{cases}
W&=\mathbb{C}_{AB}\mathbb{C}_{B}^{-1},\\
v&=\mathbb{E}\{A\}-W\mathbb{E}\{B\},
\end{cases}
\end{equation}
where $\mathbb{C}_{AB}$ denotes the cross-covariance matrix between $A$ and $B$, $\mathbb{C}_{B}$ is the auto-covariance matrix of $B$. Thus the linear MMSE estimator is given by
\begin{equation}
A^*=\mathbb{C}_{AB}\mathbb{C}_B^{-1}(B-\mathbb{E}\{B\})+\mathbb{E}\{A\}.
\end{equation}
Clearly, the linear MMSE estimator is dependent on the first- and second-order moments of $A$ and $B$.

In a multi-image deblurring setting, the underlying sharp image is generally interrupted by different degradation operations, yielding multiple observations. A typical case is a pair of blurry/noisy images. These two images are correlated with each other and can be restored simultaneously. To be clear, the noisy image can first be denoised to produce a nearly sharp image which can then be used as a guideline or a constraint in the deblurring process. A recent approach by \citet{michaeli2012partially} has exploited this strategy by proposing the partially linear MMSE (PLMMSE) estimator. Denoting the denoised image $C$ as a constraint on the estimation of $W$ and $v$ in (\ref{linearMMSE}), PLMMSE is
\begin{equation}
\mathbb{E}\{A|B,C\}=W(C)B+v(C),
\end{equation}
where $W(C)$ and $v(C)$ are functions of $C$. However, the above estimator is not applicable since it needs the knowledge of the conditional covariance $\mathbb{C}_{AB|C}$ which is difficult to acquire \citep{michaeli2012partially}. Therefore a relaxation of the restriction is conducted, resulting in a separable partially linear MMSE:
\begin{equation}
\mathbb{E}\{A|B,C\}=WB+v(C).
\end{equation}
In this case, the PLMMSE estimator is given by
\begin{equation}
A^*=\mathbb{C}_{A\tilde{B}}\mathbb{C}_{\tilde{B}}^{-1}\tilde{B}+\mathbb{E}\{A|C\},
\end{equation}
where
\begin{equation}
\tilde{B}=B-\mathbb{E}\{B|C\}.
\end{equation}
For more details of the derivation, please see the Appendix in \citep{michaeli2012partially}. The superiority of PLMMSE over LMMSE lies in the fact that PLMMSE only requires the knowledge of the second-order statistics of $A$ and $B$, and this estimator can reach the lowest worst-case MSE among all estimators which depend solely on the second-order statistics of $A$ and $B$.

\subsection{Variational Bayesian methods} \label{VBM}

\emph{Variational Bayesian methods} approximate the intractable integrals arising in a Bayesian inference framework. This situation occurs when there are unknown variables and latent variables over which we want to marginalize without explicitly computing them, such as the marginalization over $x$ in the above $\text{MAP}_h$ problem. This type of method, on one hand, provides an analytical approximation of the posterior probability of the unobserved variables to deduce the statistical properties of these variables \citep{fergus2006removing}. On the other hand, it derives a lower bound for the marginal likelihood of the observed data, which can be used to model selection \citep{levin2011efficient} and high-order statistics analysis on the likelihood of the unobserved variables \citep{zhang2010denoising}.

Variational Bayesian methods have recently been applied to image deblurring. \citet{levin2011understanding} have made the important comment that the variational Bayes approximation experimentally outperforms all existing methods with a different estimation strategy in motion deblurring tasks.

The most frequently-used type of variational Bayes is known as \emph{mean-field variational Bayes}. Recall that the posterior distribution $p(A|B)$ is over the set of unobserved variables $A$ given the observed data $B$. Here we try to approximate $p(A|B)$ by finding a variational distribution $q(A)$ that is restricted to a family of distributions with a simpler form than $p(A|B)$, i.e. $q(A)\approx p(A|B)$. The mean-field variational Bayes utilizes the KL divergence to measure the difference between the two distributions,
\begin{equation}\label{VariaionalBayesDef}
\mathcal{L}(q):=D_{KL}(q||p)=\int q(A)\log\frac{q(A)}{p(A|B)}dA.
\end{equation}
To apply this strategy to the deblurring task, \citet{miskin2000ensemble} developed an ensemble learning strategy. In their method, the unobserved set $A$ is an ensemble of the sharp image $x$, the blur kernel $h$ and the noise variance $\sigma^2$ if the Gaussian noise is assumed. The addition of $\sigma^2$ frees the user from tuning this parameter, as well as adaptively balancing the data-fidelity term and the prior term. For simplification, a separable factorization of the $q$ distribution is employed, which corresponds to the \emph{mean field theory} \citep{bishop2006pattern},
\begin{equation}
q(x,h,\sigma^2)=q(x)q(h)q(\sigma^2).
\end{equation}
This leads to
\begin{eqnarray}\label{meanfieldLQ}
\mathcal{L}(q)&=&\mathbb{E}_{q(x)}\left\{\log\frac{q(x)}{p(x|y)}\right\}
+\mathbb{E}_{q(h)}\left\{\log\frac{q(h)}{p(h|y)}\right\}\nonumber\\
&&+\mathbb{E}_{q(\sigma^2)}\left\{\log\frac{q(\sigma^2)}{p(\sigma^2|y)}\right\}.
\end{eqnarray}
The variable $B$ in (\ref{VariaionalBayesDef}) is the observed blurry image $y$. Our goal is to minimize $\mathcal{L}(q)$ with respect to $q(x,h,\sigma^2)$. By taking the convenience of the above factorization, we can take an alternating update procedure of coordinate descent to solve (\ref{meanfieldLQ}), that is: minimizing one factor (for example $q(x)$) while marginalizing out the other factors ($q(h)$ and $q(\sigma^2)$).  The updates are performed by computing the closed-form optimal parameter updates, and performing line-search along the direction of these updated values. According to \citep{bishop2006pattern}, the optimal factors obtained from the corresponding sequential updates are given by the expectation of the joint distribution with respect to all unobserved variables except the one of interest, and thus
\begin{eqnarray}
\log q^*(x)&=&\mathbb{E}_{q(h),q(\sigma^2)}\{\log p(x,h,\sigma^2,y)\}+const\nonumber\\
&=&\mathbb{E}_{q(h)}\{\log p(y|x,h)\}+\log p(x)+const\label{vbx},\nonumber\\
&&\\
\log q^*(h)&=&\mathbb{E}_{q(x),q(\sigma^2)}\{\log p(x,h,\sigma^2,y)\}+const\nonumber\\
&=&\mathbb{E}_{q(x)}\{\log p(y|x,h)\}+\log p(h)+const\label{vbh},\nonumber\\
&&\\
\log q^*(\sigma^2)&=&\mathbb{E}_{q(x),q(h)}\{\log p(x,h,\sigma^2,y)\}+const\nonumber\\
&=&\log p(\sigma^2)+const,\label{vbsig}\nonumber\\
\end{eqnarray}
Following the calculation of the above equations, the final estimates of the sharp image $x$, the blur kernel $h$ and the noise variance $\sigma^2$ are taken as the mean values of the distributions $q^*(x)$, $q^*(h)$ and $q^*(\sigma^2)$, respectively.

Using this framework, \citet{fergus2006removing} and \citet{whyte2010non,whyte2012non} operated the variational formulation on the gradient domain, i.e., $x$ in the above equations is replaced by the image gradients $\nabla x$, to facilitate the statistical assumption on model priors, such as sparse gradients. \citet{babacan2012bayesian}] proposed a super-Gaussian prior over the image intensities and treated the parameter of this prior as a latent variable in the variational inference. The optimal solutions are similar to equations (\ref{vbx}-\ref{vbsig}).

\citet{levin2011efficient} employed the variational inference to handle the marginalization problem in $\text{MAP}_h$. By re-arranging the equation (\ref{VariaionalBayesDef}), a free energy is defined as
\begin{eqnarray}
\mathcal{F}(q)&=&\int q(x,z)\log q(x,z)dzdx\nonumber\\
&&-\int q(x,z)\log p(y,x,z|h)dzdx\nonumber\\
&=&D_{KL}(q(x,z)||p(x,z|y,h))-\log p(y|h),
\end{eqnarray}
where $z$ is an auxiliary latent variable specifying the prior distribution. Due to the non-negativeness of the KL-divergence, minimizing $\mathcal{F}(q)$ is equivalent to maximizing the lower bound of the likelihood $p(y|h)$ which is the goal of $\text{MAP}_h$. Following an iterative optimization procedure, the optimal distribution can be solved. Levin et al.'s method differs from \citet{fergus2006removing} and \citet{whyte2010non,whyte2012non} in that the target distribution to be approximated is $p(x|y,h)$ rather than $p(x,h|y)$. However, Fergus et al.'s approach selects $h^*$ from the estimated $q(x,h)$ distribution by marginalizing out all possible $x$'s, and thus belongs to the $\text{MAP}_h$ approach.

Rather than applying variational inference for deblurring tasks, \citet{zhang2010denoising} explored a different application that compares the reliability of two restoration tasks in the camera shake situation: one is to estimate from a blurry image (deblurring) and the other addresses a sequence of noisy images (denoising). The posterior probabilities of these two tasks are $p_b=p(x|y_b,\sigma_b^2)$ and $p_n=p(x|\{y_n^k\}_{k=1}^N,\sigma_n^2)$, respectively, where $y_b$ is the blurry image, $y_n^k$ is the $k$-th noisy image, $\sigma_b^2$ and $\sigma_n^2$ are the noise variance in two types of image. Since the Hessian matrices of $\log p_b$ and $\log p_n$ with respect to $x$ provide an uncertainty measurement associated with the estimation of $x$, the authors applied the variational distributions over the hidden variables (here, the motion paths) to approximate the posterior, i.e. $\mathcal{L}_b(q)=D_{KL}(q||p_b)$ and $\mathcal{L}_n(q)=D_{KL}(q||p_n)$. The comparison result between the Hessian matrix of $\mathcal{L}_b(q)$ and that of $\mathcal{L}_n(q)$ reveals that restoration by denoising multiple images is generally more reliable than deblurring a single image.

\section{Variational Methods} \label{variationalmethod}

\emph{Variational methods} stem from the calculus of variations and are typically used as approximation methods in a wide variety of settings, such as quantum mechanics, classical mechanics, finite element analysis, and statistics. Such approximations are engaged to convert an ill-posed problem into a well-posed problem which is characterized by exploring additional constraints to reduce the size of the solution space of the unknown variables \citep{jordan1999introduction}. To approximate the problem, a typical setting involves the extremum (maximum or minimum) of a functional composing a function and the associated constraints:
\begin{equation}\label{unifyvariational}
\min_A \mathrm{\Phi}(A;B)+\lambda\mathrm{\Psi}(A),
\end{equation}
where $A$ is the undetermined variables and $B$ is the observations. In variational principle, $\mathrm{\Phi}(A;B)$ is called the data-fidelity function, $\mathrm{\Psi}(A)$ is the regularization function, and $\lambda$ denotes the regularization parameter. Under this formulation, the non-blind image deblurring problem can be written as
\begin{equation}
\min_{x}\mathrm{\Phi}(x;y,h)+\lambda_x\mathrm{\Psi}_x(x),
\end{equation}
while the blind case is
\begin{equation}\label{blindvariational}
\min_{x,h}\mathrm{\Phi}(x,h;y)+\lambda_x\mathrm{\Psi}_x(x)+\lambda_h\mathrm{\Psi}_h(h).
\end{equation}
The term $\mathrm{\Phi}$ is determined according to the noise assumptions listed in Table \ref{noisetable}. In this section, we generally assume the Gaussian noise model, and the corresponding $\mathrm{\Phi}$ is given by
\begin{equation}
\mathrm{\Phi}=\|y-x*h\|_2^2.
\end{equation}
The discussion of the variational methods is organized from three aspects: regularization, optimization, and analysis/synthesis reformulation.

\subsection{Regularization Techniques} \label{regularizationtech}

Similar to the character of the prior in the Bayesian inference framework, the regularizers in a variational framework express human knowledge on the interested blurry images. Such knowledge can constrain the solution space such that the deblurred images are favored by human sense.

\subsubsection{Regularization in Single-image Deblurring}

Classically, to stabilize the deblurring result, it is expected the solution to have a small norm, and thus the Tikhonov-Miller regularizer \citep{tikhonov1963solution} is imposed on the sharp image:
\begin{equation}
\mathrm{\Psi}_x(x)=\|x\|^2.
\end{equation}
However, this choice is rarely used in modern deblurring tasks because of the property whereby the resultant images will have over-smoothed edges. With this regard, the development of the first-order regularizers which maximally preserve the significant details is more frequently adopted. A typical example is the \emph{total variation} (TV) proposed by \citet{rudin1992nonlinear}
\begin{equation}
TV_i(x)=\|\sqrt{|\nabla_h x|^2+|\nabla_v x|^2}\|_1,
\end{equation}
where the subscript $i$ means it is the \emph{isotropic} version. Complementarily, the \emph{anisotropic} TV is
\begin{equation}
TV_a(x)=\||\nabla_h x|+|\nabla_v x|\|_1.
\end{equation}
Both $\text{TV}_i$ and $\text{TV}_a$ enhance the visualization of edges in the resultant images. They mainly differ from each other in their sensitivity to edge directions. From the formulations, we can see that $\text{TV}_i$ enforces the same strength on the edges with different directions, whereas $\text{TV}_a$ favors certain directions. Both methods have proven to be useful in numerous applications, such as image denoising, decomposition, super-resolution, inpainting, and non-blind deblurring. Nevertheless, when applied to blind deblurring problems, some failures occur.

Note that TV is intrinsically an $\ell_1$ norm of the image gradients, and thus induces sparsity over image gradients. According to the delta-effect of MAP mentioned in Section \ref{bayesinferframe}, simultaneously estimating $x$ and $h$ will result in a blurry image. Another perspective from the $\ell_1$ properties can assist the understanding of TV failure. For a sharp image of natural scenes, the gradient magnitude is typically sparse, meaning that most values are either zero or very small, but may occasionally be large. If a blur kernel is operated on this image, the high-frequency bands will be attenuated, leading to the magnitudes being un-sparse. To recover the original sparsity, a natural choice is the $\ell_0$ measure, an important property of which is the scale-invariance, i.e., $\min \ell_0(\nabla x)=\min \ell_0(a\cdot\nabla x)$ for any positive values of $a$. Minimizing $\ell_0$ will only lead to a sparse effect, without destroying the magnitudes of large values, thus preserving the energy of original gradients. However, $\ell_0$ is difficult to optimize because of the lack of derivative information everywhere, and then $\ell_1$ is utilized as an alternative to approximate $\ell_0$. Unfortunately, the blurring process in itself reduces the $\ell_1$ norm of the gradients. Minimizing $\ell_1$ fails to preserve or recover the energy of the original gradients. Additionally, the scale variant property makes $\ell_1$ sensitive to the setting of the regularization parameter $\lambda$. Therefore, various methods of approximating the $\ell_0$ norm while maintaining the scale-invariance property are proposed.

\begin{figure}[t]
\centering
\subfloat[$\ell_0$]{\includegraphics[height=1.5in]{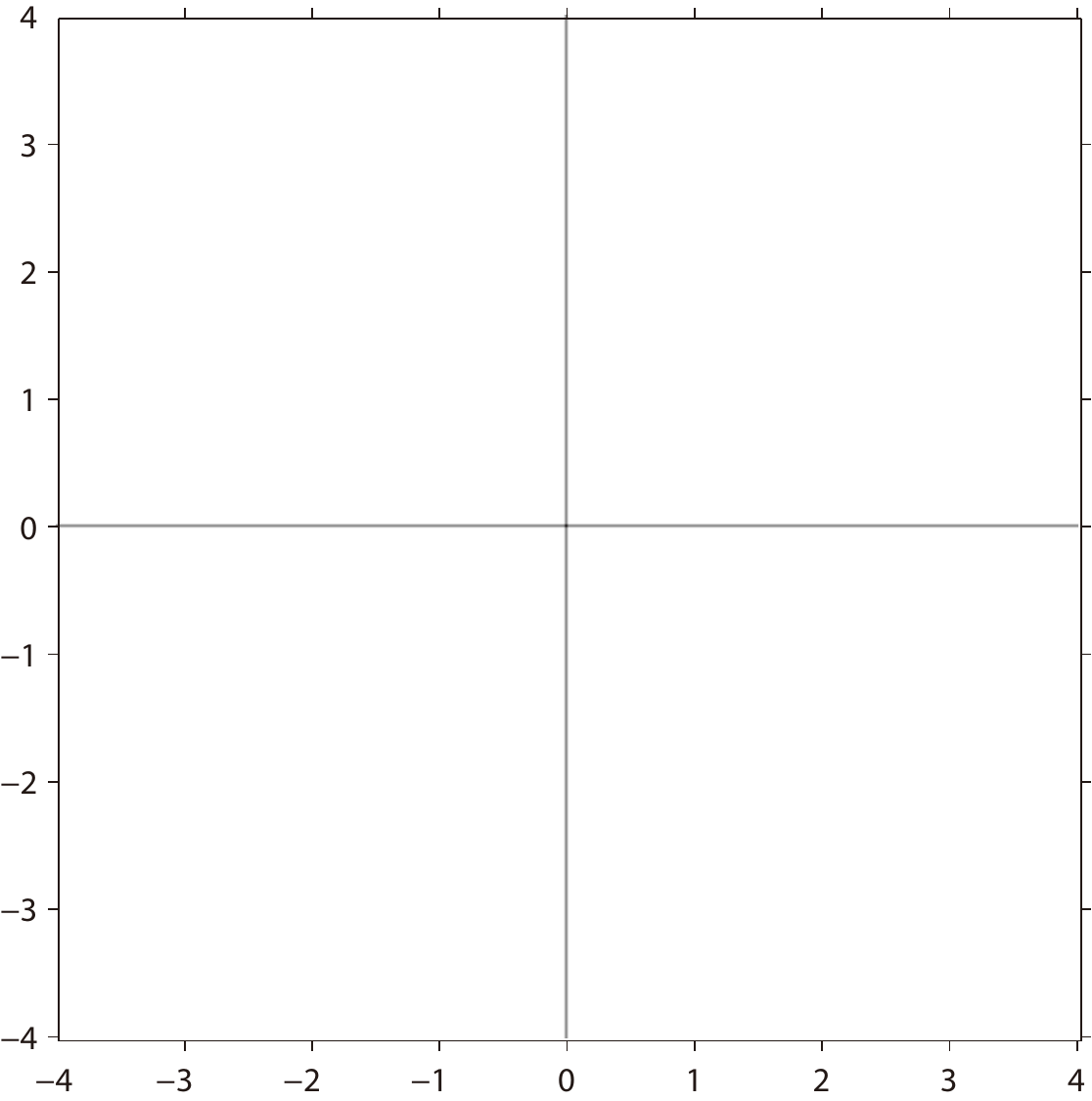}\label{norml0}}
\hspace{0.1cm}
\subfloat[$\ell_1$]{\includegraphics[height=1.5in]{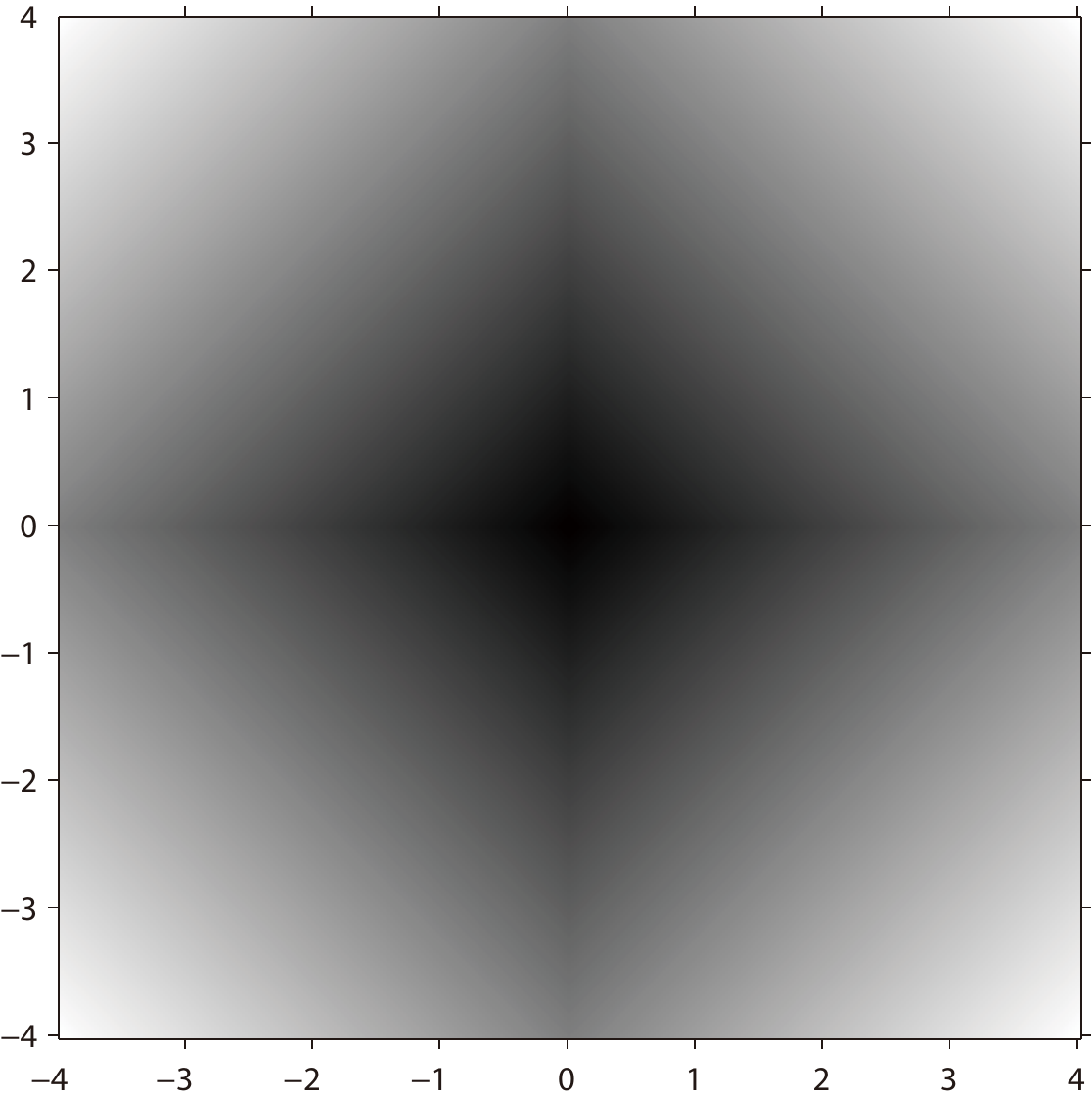}\label{norml1}}
\hspace{0.1cm}
\subfloat[$\ell_1/\ell_2$]{\includegraphics[height=1.5in]{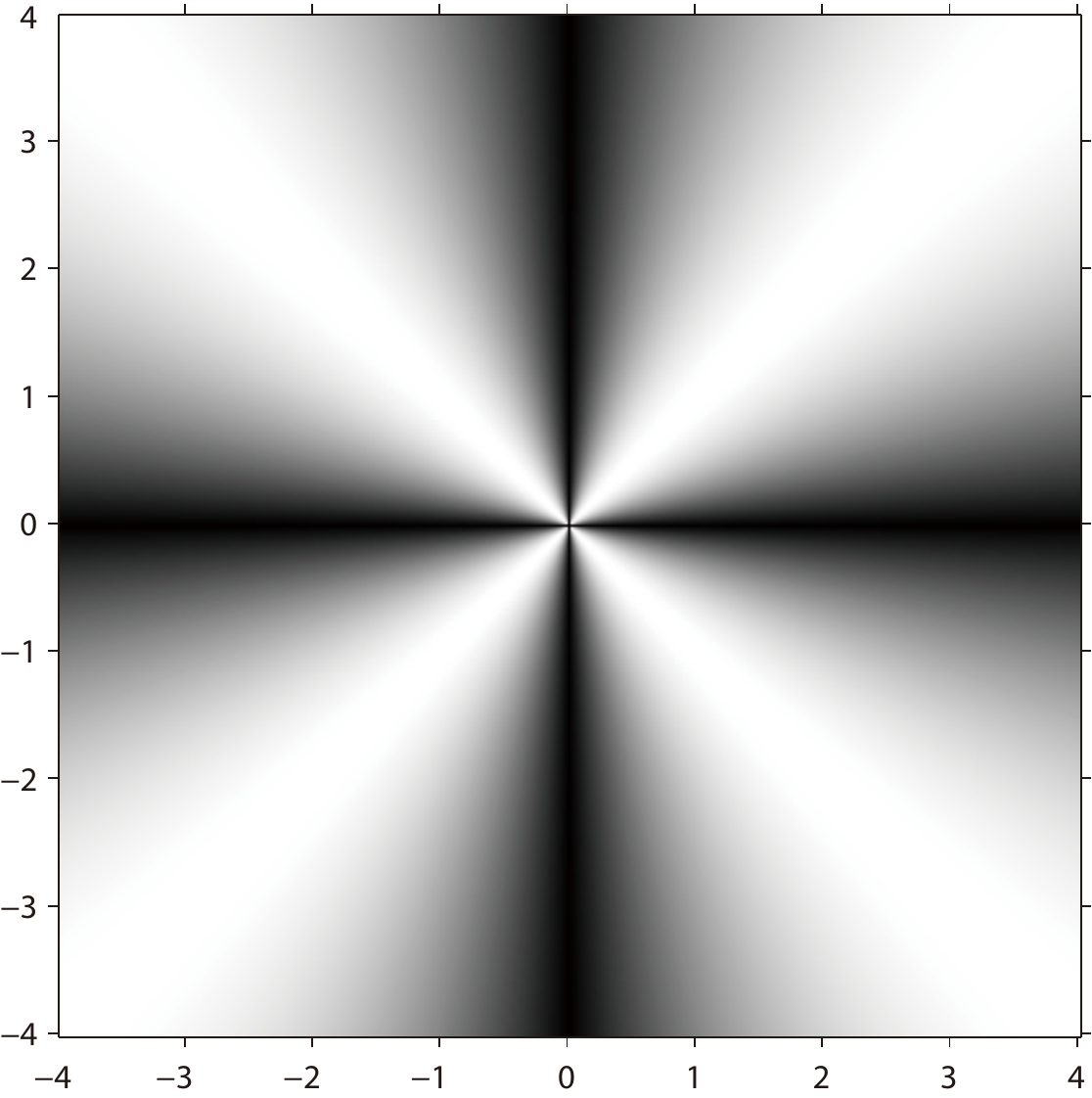}\label{norml12}}\\
\subfloat[Unnatural $\ell_0$ ($\epsilon=0.4$)]{\includegraphics[height=1.5in]{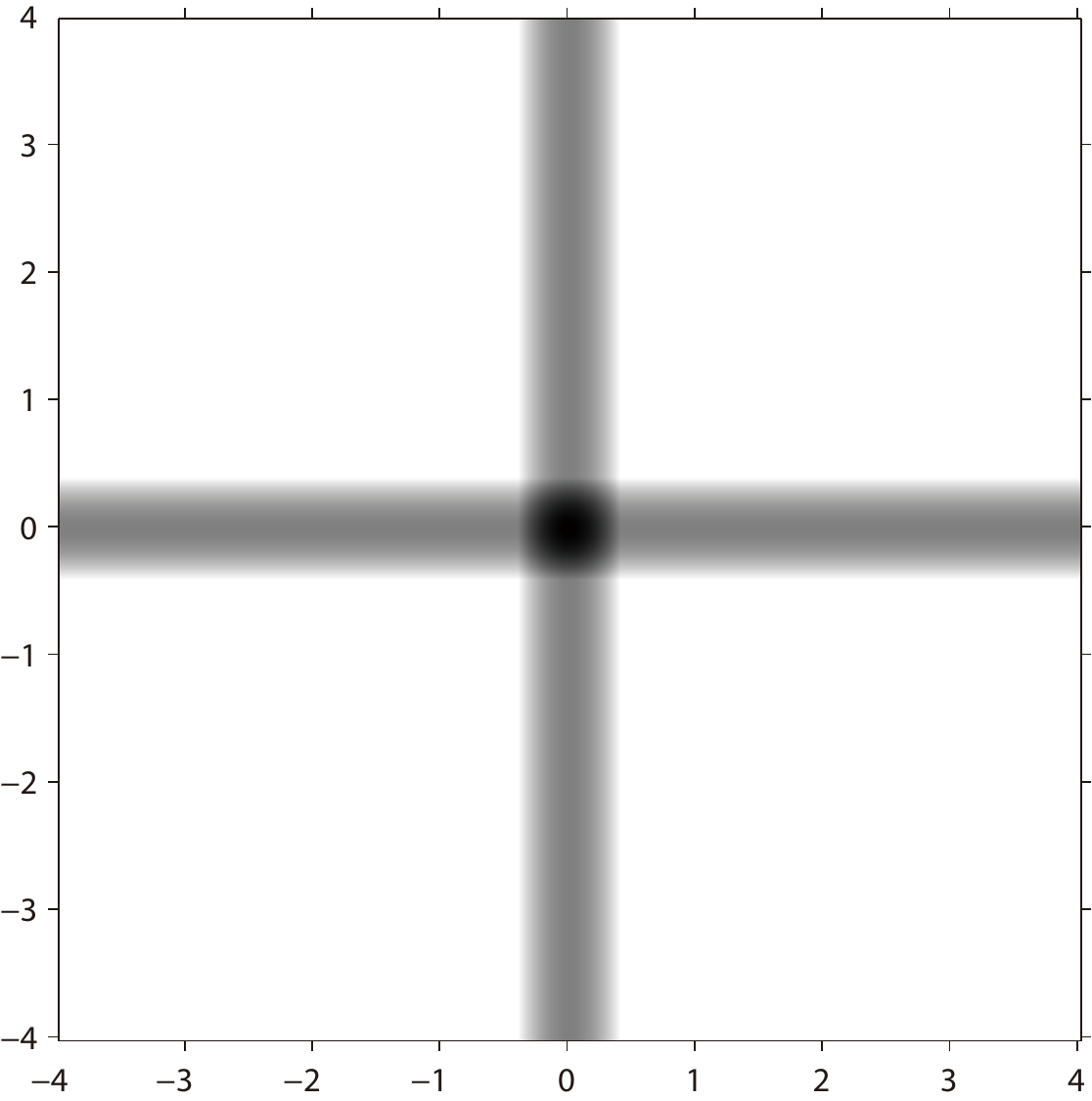}\label{norml0u1}}
\hspace{0.1cm}
\subfloat[Unnatural $\ell_0$ ($\epsilon=0.1$)]{\includegraphics[height=1.5in]{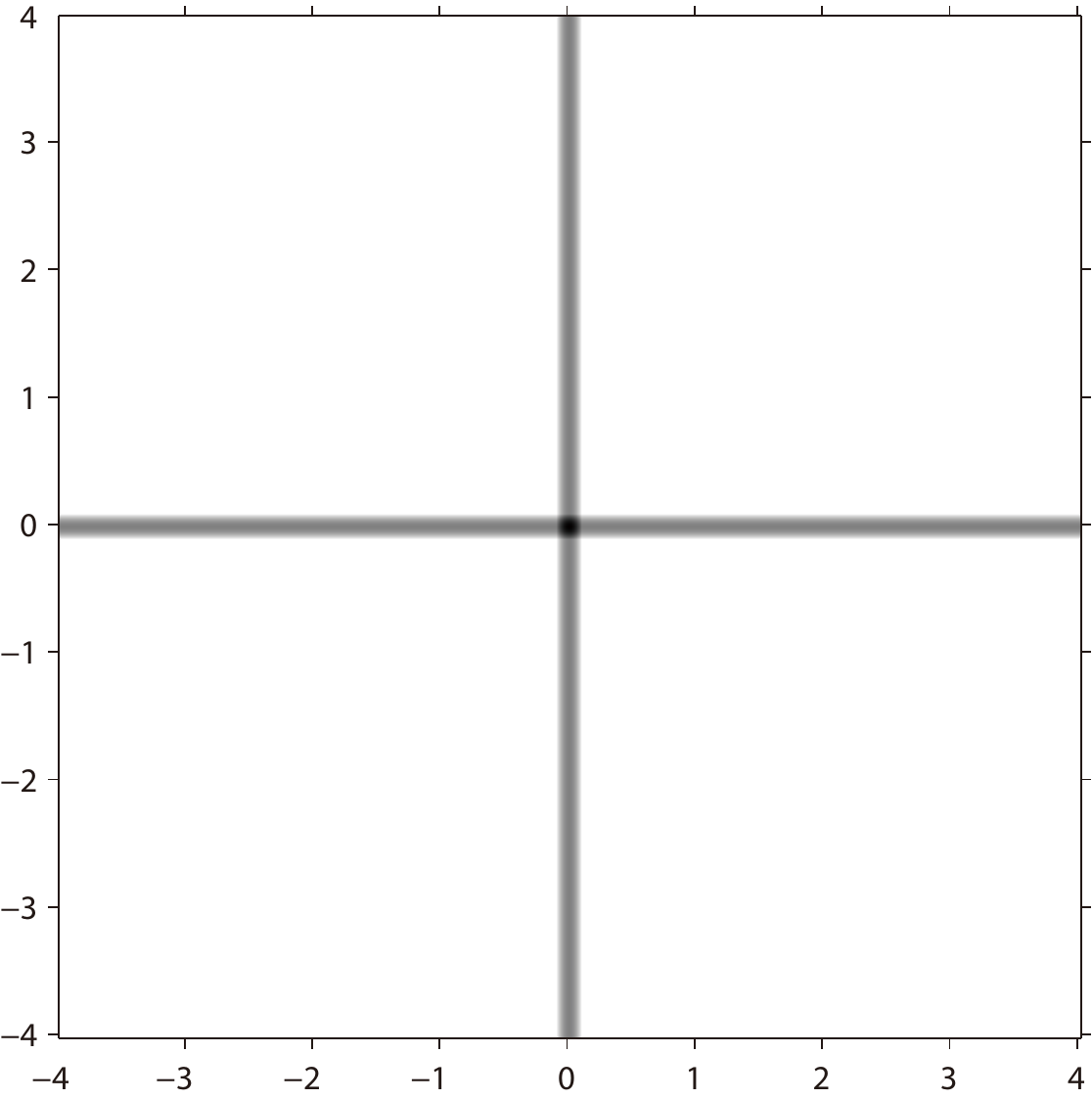}\label{norml0u2}}
\hspace{0.1cm}
\subfloat[Unnatural $\ell_0$ ($\epsilon=0.1$)]{\includegraphics[height=1.5in]{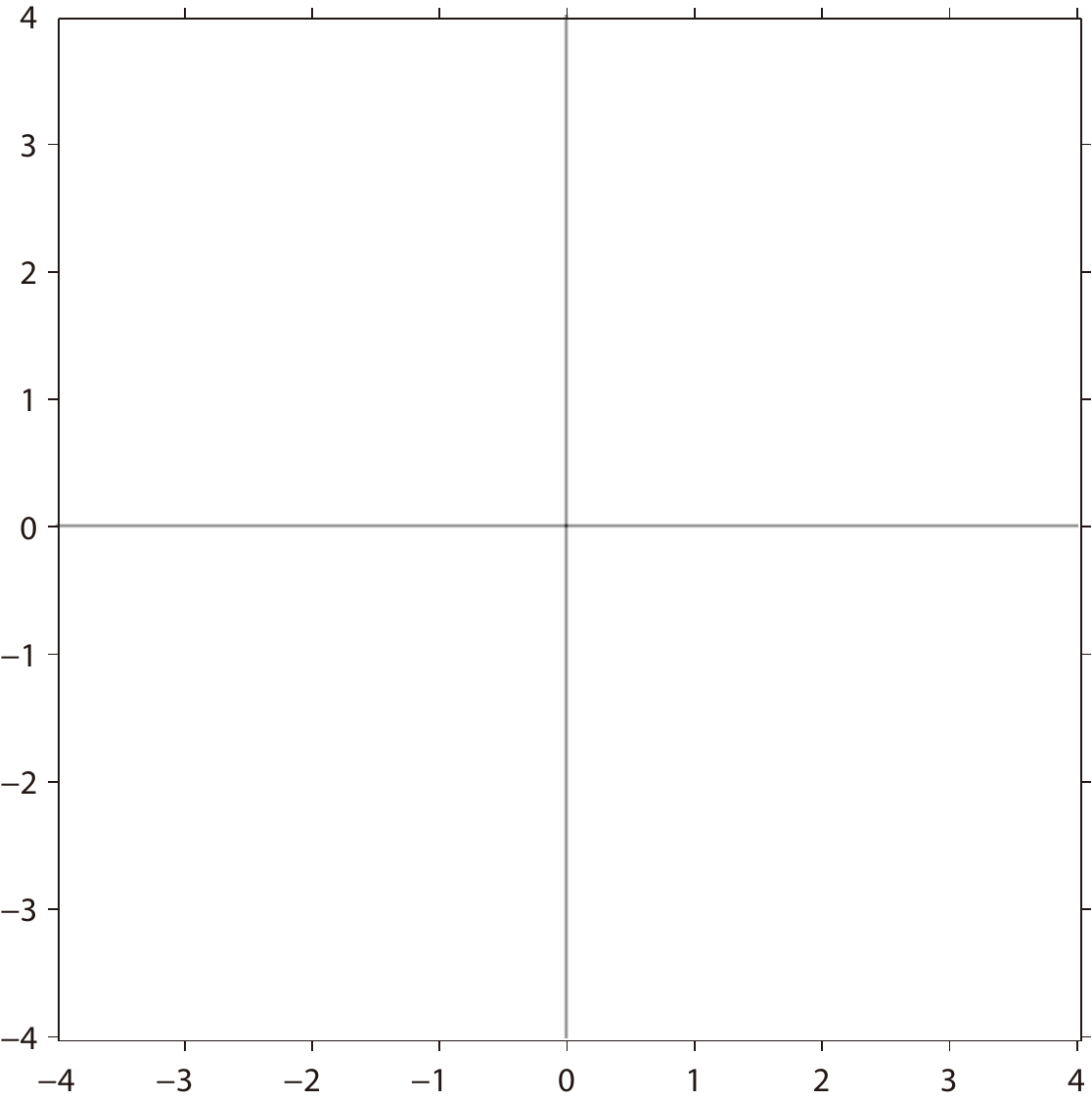}\label{norml0u3}}
\caption{Visualization of different measures}
\label{norml}
\end{figure}

\citet{krishnan2011blind} recently extended the $\ell_1$ norm to a normalized version:
\begin{equation}
\mathrm{\Psi}(\nabla x)=\frac{\|\nabla x\|_1}{\|\nabla x\|_2}.
\end{equation}
To understand this regularizer, let us focus on the denominator, $\ell_2$ norm. The blurring process reduces the $\ell_2$ norm of the gradients as well. Fortunately, $\ell_2$ is reduced more than the numerator $\ell_1$ norm, leading to an increased ratio of the two terms. Fig. \ref{norml12} illustrates that the minimum of this ratio is along the axes. The blurry effect will drive the ratio away from the axes. Therefore, minimizing this regularizer will deduce the blurry effect in the image without destroying the magnitude of the true gradient because $\ell_1/\ell_2$ is evidently scale invariant, just as we expected.

Another example of the approximation is the unnatural $\ell_0$ regularizer which is proposed by \citet{xu2013unnatural}. The unnaturalness stems from the observation that in most iterative deblurring methods, the intermediate image results only contain high-contrast and step-like structures while suppressing others. These images are different from natural scenes, and hence the term 'unnatural' is exploited. To incorporate the step-edge properties in an unnatural representation, the authors utilized the unnatural $\ell_0$ scheme to preserve the salient changes (i.e., the gradients) in the image. The resultant regularizer is formulated as
\begin{equation}
\mathrm{\Psi}(\nabla_h x)=\sum_i\mathrm{\psi}(\nabla_h x_i),
\end{equation}
where
\begin{equation}
\mathrm{\psi}(z)=
\begin{cases}
\frac{1}{\epsilon^2}|z|^2,&\text{if }|z|\leq\epsilon,\\
1,&\text{otherwise}.
\end{cases}
\end{equation}
The definition on the vertical derivative $\nabla_v x_i$ is similar. Depending on the formulation, the gradient magnitudes smaller than $\epsilon$ are penalized by $\mathrm{\psi}(\cdot)$ while the larger values result in a constant $1$ in the objective function. Minimizing this regularizer will remove fine structures and keep useful salient details in the result. Fig. \ref{norml0u1}-\ref{norml0u3} illustrates three plots under different values of $\epsilon$. When $\epsilon$ approaches to zero, this regularizer can be fitted perfectly to the $\ell_0$ norm. Another property ensuring the unnatural $\ell_0$ superior to $\ell_1$ is its scale invariance property, as previously stated. By using this regularization technique in the estimation of blur kernels, the deblurring performance has been notably improved.

While the above regularizers are all based on first-order derivatives, second-order regularization techniques have also proven to be useful in image denoising tasks, and have recently been introduced to deblurring images. \citet{lefkimmiatis2012hessian} extended the first-order TV functional to two second-order cases by defining the mixed norms including $\ell_1-\ell_{\infty}$ and $\ell_1-\ell_2$. These regularizers maintain favorable properties of TV (such as convexity, homogeneity, rotation and translation invariance) well, and can effectively suppress the staircase effect. To solve the resultant variational problem, an efficient algorithm is proposed based on the majorization-minimization approach. Rather than only enforcing the second-order regularization in deblurring tasks, \citet{papafitsoros2014combined} handled the combined problem involving both first- and second-order functionals. The benefit is that the first-order term recovers the step-edges as well as possible, while the second-order term eliminates the artifacts of the staircase produced by the first-order regularizer, without introducing any serious blur in the reconstructed image. Further, the existence and uniqueness of the solution to the combined problem is proved, and numerical solutions are provided based on the split Bregman iteration \citep{goldstein2009split}.

A limitation of most regularizers is noted to be based on the local principle, i.e. regularizing the local structures, which can be overcome by the nonlocal ideas. Inspired by the development of nonlocal TV \citep{gilboa2007nonlocal,gilboa2008nonlocal} in the image deblurring task \citep{lou2010image}, \citet{jung2011nonlocal} derived a nonlocal \emph{Mumford-Shah} (MS) regularizer by applying the nonlocal operators to the multichannel approximations of the MS regularizer. Due to the repetitiveness of the textures in nature, this regularizer performs better than the local counterpart in various image applications.

\subsubsection{Regularization in Multi-image Deblurring}

In the case of multi-image deblurring, the regularizers should not only express human knowledge but also exhibit mutual constraints among different images. In this setting, we need to involve a multiple blurring model:
\begin{equation}\label{mutliblurmodel}
y_k=x*h_k+n_k, \text{  for }k=1,2,...
\end{equation}
Different observation $y_k$ is obtained by convolving the same latent image $x$ with a different kernel $h_k$, interrupted by specific noise $n_k$. Given equation (\ref{mutliblurmodel}), the blind deblurring process can be written as a joint variational formulation:
\begin{eqnarray}\label{multideblurmodel}
(x^*,\{h_k^*\})&=&\mathop{\argmin}\limits_{x,\{h_k\}} \sum_k\|y_k-x*h_k\|_2^2+\lambda_x\mathrm{\Psi}_x(x)\nonumber\\
&&+\lambda_h\sum_k\mathrm{\Psi}_h(h_k).
\end{eqnarray}
Detailed discovery of the relationship among the kernels $\{h_k\}$ can help us handle the above inverse problem. Specifically in terms of two-image deblurring, \citet{li2011theory} presented a theory of \emph{Coprime Blurred Paris} (CBP). CBP means that in the blurry image pair, the $z$-transforms of the two kernels are coprime. Mathematically, the equations in (\ref{mutliblurmodel}) are transformed into
\begin{equation}
\begin{cases}
\tilde{y}_1(z_1,z_2)&=\tilde{x}(z_1,z_2)\cdot\tilde{h}_1(z_1,z_2)+\tilde{n}_1(z_1,z_2)\\
\tilde{y}_2(z_1,z_2)&=\tilde{x}(z_1,z_2)\cdot\tilde{h}_2(z_1,z_2)+\tilde{n}_2(z_1,z_2)
\end{cases},
\end{equation}
where $\tilde{y}$, $\tilde{x}$, $\tilde{h}$ and $\tilde{n}$ are the $z$-transform of $y$, $x$, $h$ and $n$, respectively. The coprimeness between $\tilde{h}_1(z_1,z_2)$ and $\tilde{h}_2(z_1,z_2)$ will theoretically lead to an approximation of the optimal solution $x^*$, that is
\begin{equation}\label{GCD}
\tilde{x}^*=\text{GCD}\{\tilde{y}_1(z_1,z_2),\tilde{y}_2(z_1,z_2)\},
\end{equation}
where GCD is the abbreviation of the greatest common divisor \citep{pillai1999blind}, a classic method in \emph{Number Theory}. However, computing (\ref{GCD}) requires great computation power and system memory, making GCD impractical in deblurring tasks. To address this issue, the authors suggested recovering the sharp image by factoring the B\'{e}zout Matrix of $\tilde{y}_1$ and $\tilde{y}_2$ to determine the coprimality and can be efficiently solved.

With the assistance of additional hardware, \citet{li2011exploring} captured two images of the same scene disturbed by camera shake blur. These two images are correlated according to a mapping $g$, and can be formulated as
\begin{equation}\label{mappingimage}
\begin{cases}
y_1&=x*h+n_1\\
y_2&=g(x)*h+n_2
\end{cases}.
\end{equation}
The mapping $g$, as expected, needs to be bijective isometry, and satisfies that $g^{-1}(a*b)=g^{-1}(a)*g^{-1}(b)$ for all $a,b\in \mathbb{R}^n$. Imposing the inverse mapping of $g$ to the second equation of (\ref{mappingimage}), we obtain
\begin{equation}
g^{-1}(y_2)=x*g^{-1}(h)+n_2=x*h'+n_2,
\end{equation}
where $h'=g^{-1}(h)$. Substituting these two observation models into equation (\ref{multideblurmodel}), the two-step optimization is
\begin{equation}
\begin{cases}
h^*=&\mathop{\argmin}\limits_h\|y_1-x*h\|_2^2+\|y_2-g(x)*h\|_2^2\\
&+\lambda_h\mathrm{\Psi}_h(h),\\
x^*=&\mathop{\argmin}\limits_x\|y_1-x*h\|_2^2+\|g^{-1}(y_2)-x*g^{-1}(h)\|_2^2\\
&+\lambda_x\mathrm{\Psi}_x(x).
\end{cases}
\end{equation}
In practice, the mapping $g$ is instantiated as $rot90$, i.e. an image is the $90^\circ$-rotated version of the other. This is achieved by employing a beam splitter in the consumer camera. An advantage of this method is the avoidance of the blurry image alignment which, if not properly processed, brings serious artifacts into the restored images.

Even though the alignment problem is crucial in multi-image deblurring, it is possible to handle without an accurate alignment. \citet{hacohen2013deblurring} addressed the deblurring task by discovering the dense correspondence \citep{hacohen2011non} between a blurry image and a sharp reference image. The reference image acts as a regularizer such that the blurry information can be restored from the corresponding sharp information. An assumption in this method is that the two images are required to share the same content but undergo non-rigid geometric transformations; and there is no need to construct an alignment between them.

\subsection{Optimization Methods} \label{optimizationmethod}

In a variational framework, as well as in other related schemes, a good optimization algorithm can achieve a fast convergence rate and produce an accurate solution. For the problem of deblurring, the general formulation is
\begin{equation}\label{optimizationfunc}
\min_{\mathbf{x}} \frac{1}{2}\|\mathbf{y}-\mathbf{Hx}\|^2+\lambda\mathrm{\Psi}(\mathbf{x}),
\end{equation}
where we utilize the matrix-vector expression in (\ref{vectorform}). Even though this is for the non-blind deblurring problem, we can see from the discussion in previous sections that the blind case is generally decomposed into a two-step procedure, in which the blur kernel is first estimated by a similar function to (\ref{optimizationfunc}) and the sharp image is then calculated according to (\ref{optimizationfunc}).

A standard algorithm for solving the problem of (\ref{optimizationfunc}) is the so-called \emph{iterative shrinkage/thresholding} (IST) algorithm, which depends on the shrinkage/thresholding function. For example, if we set the $\mathrm{\Psi}$ as the $\ell_0$ norm on $x$, the corresponding shrinkage/thresholding function is the hard-threshold function \citep{donoho1994ideal}:
\begin{equation}
\mathcal{T}_{\lambda\ell_0}(\mathbf{w})=\mathbf{w}\odot\mathbf{1}_{\mathbf{w}\geq\sqrt{2\lambda}},
\end{equation}
where $\mathbf{w}$ is the observation to be approximated, $\mathbf{1}_{\mathbf{w}\geq\sqrt{2\lambda}}$ is the indicator function determined by the condition of the subscript. If $\mathrm{\Psi}$ is $\ell_1$ norm, the soft-threshold function \citep{donoho1994ideal} is then utilized:
\begin{equation}
\mathcal{T}_{\lambda\ell_1}(\mathbf{w})=\text{sign}(\mathbf{w})\odot\max(|\mathbf{w}|-\lambda,0).
\end{equation}
For solving (\ref{optimizationfunc}), the IST iteration is given by
\begin{equation}
\mathbf{x}^{t+1}=\mathcal{T}_{\lambda\mathrm{\Psi}}\left(\mathbf{x}^t-\frac{1}{\gamma}\mathbf{H}^{\ast}(\mathbf{Hx}^t-\mathbf{y})\right),
\end{equation}
where $\mathbf{H}^{\ast}$ is the adjoint of the matrix $\mathbf{H}$, and $\frac{1}{\gamma}$ is the step size. The convergence rate of IST is determined by the parameter $\lambda$ and the matrix $\mathbf{H}$. Small values of $\lambda$ and/or the ill-condition of $\mathbf{H}$ results in slow convergence. To accelerate IST, variants have been proposed including the two-step IST algorithm \citep{bioucas2007new}, fast IST algorithm \citep{beck2009fast}, and sparse reconstruction by separable approximation \citep{wright2009sparse}. \citet{michailovich2011iterative} worked on a more general version of TV which is based on multidirectional gradients rather than only the horizontal and vertical derivatives in $\text{TV}_i$ and $\text{TV}_a$, and then derived an IST-type algorithm to solve the model.

Another popular scheme for solving the problem (\ref{optimizationfunc}) is the \emph{alternating direction method of multipliers} (ADMM), a variant of the \emph{augmented Lagrangian Method} (ALM). Formally, (\ref{optimizationfunc}) can be transformed into the following constrained problem by introducing an auxiliary variable $\mathbf{z}$:
\begin{eqnarray}\label{VSform}
\min_{\mathbf{x},\mathbf{z}}&&f_1(\mathbf{x})+f_2(\mathbf{z}),\nonumber\\
s.t.&&\mathbf{x}=\mathbf{z},
\end{eqnarray}
where $f_1(\mathbf{x})=\frac{1}{2}\|\mathbf{y}-\mathbf{Hx}\|^2$ and $f_2(\mathbf{z})=\lambda\mathrm{\Psi}(\mathbf{z})$. This procedure is called \emph{variable splitting}. By incorporating the ALM techniques, (\ref{VSform}) can be further written as
\begin{equation}
\min_{\mathbf{x},\mathbf{z},\bm{\beta}} \mathcal{L}_{\mu}(\mathbf{x},\mathbf{z},\bm{\beta})=f_1(\mathbf{x})+f_2(\mathbf{z})+\frac{\mu}{2}\|\mathbf{x}-\mathbf{z}\|_2^2-\bm{\beta}^T(\mathbf{x}-\mathbf{z}),
\end{equation}
where $\mu\geq 0$ is the penalty parameter and $\bm{\beta}$ is a vector of Lagrange multipliers. Then ADMM alternatively optimizes $\mathbf{x}$ and $\mathbf{z}$ \citep{boyd2011distributed}, associated with an additional step to estimate the Lagrange parameters
\begin{eqnarray}
\mathbf{x}^{t+1}&=&\mathop{\argmin}\limits_{\mathbf{x}}\mathcal{L}_{\mu}(\mathbf{x},\mathbf{z}^t,\bm{\beta}^t),\\
\mathbf{z}^{t+1}&=&\mathop{\argmin}\limits_{\mathbf{z}}\mathcal{L}_{\mu}(\mathbf{x}^{t+1},\mathbf{z},\bm{\beta}^t),\\
\bm{\beta}^{t+1}&=&\bm{\beta}^t+\mu(\mathbf{x}^{t+1}-\mathbf{z}^{t+1}).
\end{eqnarray}
If a transformed signal (say the gradients) of $\mathbf{x}$ is regularized, the constraint in (\ref{VSform}) is replaced by the corresponding equations ($\mathbf{Gx}=\mathbf{z}$ where $\mathbf{G}$ is the matrix expression of the derivatives). Under this framework, \citet{afonso2010fast} recently proposed a so-called SALSA (\emph{split augmented Lagrangian shrinkage algorithm}) to tackle the problem (\ref{optimizationfunc}) with a non-smooth regularizer (e.g. TV or $\ell_1$ norm on wavelet coefficients). \citet{carlavan2012sparse} used the same scheme to deblur Poisson noisy images. \citet{yang2013efficient} presented the multidimensional anisotropic TV problem which is then solved by ADMM. To accelerate the algorithm, it is recommended to decompose the original multidimensional problem into a set of independent 1D TV problems, which can be processed in parallel, thus reducing the time cost. \citet{ng2013coupled} unified the image deblurring, decompositioin and inpainting problem into one formulation, and recommended an algorithm based on ADMM to solve it.

Newton's method and its variants are also used to solve (\ref{optimizationfunc}). \citet{barbero2011fast} developed a Newton-type method under the dual formulation of the TV-proximity problem to solve the $\ell_2$-TV problem. Similarly, in \citet{chan2010efficient}'s work, Fenchel-duality and semi-smooth Newton techniques are utilized to handle a $\ell_1$-TV problem.

A crucial issue in solving the variational problem is the determination of the regularization parameter. A good selection of the parameter will result in a promising deblurring result, whereas a bad choice may lead to slow convergence as well as the existence of severe artifacts in the results. Generally, when the degradation in the blurry image is significant, the value of $\lambda$ needs to be set large, to reduce the blur as much as possible. However, in the continuing iterations, the blurry effect is decreased gradually. In this case, small value of $\lambda$ is required since a large value will damage the fine detail in the image. By considering these effects, a direct implementation is to set $\lambda$ from large to small according to an empirical reduction rule \citep{tai2011richardson,almeida2010blind,faramarzi2013unified}:
\begin{equation}
\lambda^{t+1}=\max(\lambda^t\cdot r,\lambda_{min}),
\end{equation}
which depends on the initial value $\lambda_0$, the minimal value $\lambda_{min}$ and the reduction factor $r\in(0,1)$. Usually, $r=0.5$. This setting ensures the improvement of the convergence speed of the algorithm if, at each step in the outer iteration, the optimal solution of its immediate predecessor is used as a starting point for the inner iterative steps (\emph{warm start}). The adaptive adjustment of the parameter by considering the intermediate images in each iteration is more favorable. For example, \citet{montefusco2012iterative} proposed an update rule based on the values of the loss function in two previous iterations. If we define the loss function in (\ref{optimizationfunc}) as $\mathcal{L}(\mathbf{x},\lambda)$, we then have
\begin{equation}
\lambda^{t+1}=\lambda^t\cdot\frac{\mathcal{L}(\mathbf{x}^t,\lambda^t)}{\mathcal{L}(\mathbf{x}^{t-1},\lambda^{t-1})}.
\end{equation}
If the penalty in (\ref{optimizationfunc}) is  $\ell_1$-based functional, the utilization of this adaptive rule can guarantee the convergence of the forward-backward splitting algorithm. \citet{almeida2013parameter} suggested using the residual whiteness measure as a guideline for the parameter selection and the stopping criterion. The rationale is that if the restored image is well-estimated, the residual image is spectrally white. Their strategy is experimentally demonstrated in both blind and non-blind deblurring tasks.

\subsection{Analysis \& Synthesis} \label{analysissynthesis}

There are two different kinds of formulation for modelling the variational image restoration problem, which are called \emph{analysis} and \emph{synthesis} respectively \citep{elad2010sparse,elad2007analysis}. In the analysis formulation, the image $\mathbf{x}$ is associated with an analysis operator $\mathbf{U}$ by which $\mathbf{x}$ is effectively decomposed in a transformed domain, forming the analysis representation $\mathbf{Ux}$. Constraining this representation and merging it with the Gaussian noise assumption, the problem is formalized as
\begin{equation}\label{analysisform}
\mathbf{x}^*=\mathop{\argmin}\limits_{\mathbf{x}}\|\mathbf{y}-\mathbf{Hx}\|_2^2+\lambda\|\mathbf{Ux}\|_p,
\end{equation}
where $\|\cdot\|_p$ is the $\ell_p$ norm. The TV problem is an instance of the above formulation by setting $\mathbf{U}$ as the gradient operator and $p=1$. For different forms of $\mathbf{U}$ and $p$, we can find various examples of the analysis representation in Section \ref{regularizationtech}, and iterative optimization methods in Section \ref{optimizationmethod}.

The synthesis formulation indicates that the vectorized image $\mathbf{x}\in\mathbb{R}^N$ is to be represented as a linear combination of atoms taken from the columns of a full-rank matrix $\mathbf{D}\in\mathbb{R}^{N\times M}$, where $N\leq M$. The synthesis representation is then taken as $\mathbf{x}=\mathbf{D}\bm{\alpha}$, where $\bm{\alpha}\in\mathbb{R}^M$ is expected to be sparse. Under the same noise assumption as in the analysis case, the synthesis formulation is given by
\begin{equation}\label{synthesisform}
\bm{\alpha}^*=\mathop{\argmin}\limits_{\bm{\alpha}}\|\mathbf{y}-\mathbf{HD}\bm{\alpha}\|_2^2+\lambda\|\bm{\alpha}\|_p,
\end{equation}
and the optimal solution $\mathbf{x}^*=\mathbf{D}\bm{\alpha}^*$. Synthesis-based methods stem from the basis pursuit method by \citet{chen1998atomic} and have been well developed in the past years. A particularly popular synthesis representation is sparse representation, which will be detailed in the next section. Curvelet \citep{candes2000curvelets}, contourlet \citep{do2005contourlet}, ridgelet \citep{candes1999ridgelets} and their variants are all excellent works under this framework and most of them are focused on the denoising problem. In terms of deblurring, \citet{tzeng2010contourlet} recently exploited the collaborative property of multiband deblurring in the fluid lens camera system. Since the green plane of the image remains sharp in the image formation process, this sharp information can be used to recover the contourlet coefficients of other blurry planes. \citet{zhang2013blur} proposed the double discrete wavelet transform (DDWT) whose coefficients possess near-blur-invariant properties, thus enabling the identity of the blur kernel in the DDWT domain. Under the tight wavelet frame system, \citet{ji2012robust} derived a deblurring scheme by considering the kernel error in a unified formulation, whereas \citet{cai2012framelet} used framelet decomposition to derive an analysis-based sparse prior rather than the synthesis-based prior.

Analysis and synthesis modeling share a similar structure in their formulation, but the resultant $\mathbf{x}$ from them are not necessarily equal, and may even be significantly different from each other in most cases. According to \citep{elad2007analysis}, the equivalence between the two solutions occurs in anyone of the following three situations: 1) if $\mathbf{U}$ takes as a square and non-singular analyzing operator and $\mathbf{D}=\mathbf{U}^{-1}$; 2) if $\mathbf{H}=\mathbf{I}$, $\mathbf{U}\in\mathbb{R}^{L\times N} (L\leq N)$ is a full-rank operator, and $\mathbf{D}=\mathbf{U}^+$; or 3) if $p=2$, $\mathbf{U}\in\mathbb{R}^{L\times N} (L>N)$ is a full-rank operator, and $\mathbf{D}=\mathbf{U}^+$. Even if one of these equivalences exist, the difference between the analysis scheme and the synthesis scheme suggests each has merits and demerits in specific tasks. For example, the redundancy of the operators $\mathbf{U}$ and $\mathbf{D}$ has a different effect on the analysis approach and the synthesis approach, respectively. While a redundant $\mathbf{D}$ enables the synthesis scheme to describe more complex signals, redundant $\mathbf{U}$ in the analysis case makes the signal inconsistent. On the other hand, in the synthesis scheme, considering the dependence among the atoms in $\mathbf{D}$, the incorrect selection of one atom can propagate to the others; in contrast, for the analysis scheme, considering each pixel is related to its neighbor pixels, this dependence is significantly reduced, which ensures the imprecise estimation of one pixel will not seriously affect the estimation of the other pixels \citep{elad2007analysis}.

The above analysis is based on a single analysis formulation or a single synthesis formulation. Nevertheless, a recent state-of-the-art work \citep{danielyan2012bm3d,danielyan2010image} suggested a combined scheme to address the deblurring problem, i.e. \emph{block-matching 3-D} (BM3D) image modeling (originally proposed for the image denoising task \citep{dabov2007image}). \citet{danielyan2012bm3d,danielyan2010image} deduced the BM3D modeling from both the analysis formulation and the synthesis formulation, proposing the so-called BM3D frames. Within the formulations of (\ref{analysisform}) and (\ref{synthesisform}), the authors extended BM3D modeling to modern variational image restoration techniques. Furthermore, a combination of analysis/synthesis is formulated as a \emph{generalized Nash equilibrium} (GNE) problem including two objective functions, which are
\begin{equation}
\begin{cases}
\mathbf{x}^*=\mathop{\argmin}\limits_{\mathbf{x}}\|\mathbf{y}-\mathbf{Hx}\|_2^2,&\textit{s.t. }\|\mathbf{x}-\mathbf{D}\bm{\alpha}^*\|_2^2\leq\varepsilon_1,\\
\bm{\alpha}^*=\mathop{\argmin}\limits_{\bm{\alpha}}\lambda\|\bm{\alpha}\|_p,&\textit{s.t. }\|\bm{\alpha}-\mathbf{Ux}^*\|_2^2\leq\varepsilon_2,
\end{cases}
\end{equation}
where $\varepsilon_1,\varepsilon_2>0$ and $p=0,1$. The optimization algorithm iteratively solves the two functions. As noted by the authors, simultaneously minimizing the two functions cannot be achieved since minimization of either one of them will lead to an increase of the other. This effect is called the \emph{Nash equilibrium} and provides a balance between the fit of the restoration $\mathbf{x}$ to observation $\mathbf{y}$ and the complexity of the model $\|\bm{\alpha}\|_p$. This balance plays a vital role in the image restoration.

\citet{xie2012alternating} and \citet{shen2011accelerated} worked on the balanced formulation:
\begin{equation}\label{balanceform}
\min_{\bm{\alpha}} \|\mathbf{y}-\mathbf{H}\mathbf{D}\bm{\alpha}\|_2^2+\gamma\|(\mathbf{I}-\mathbf{D}^T\mathbf{D})\bm{\alpha}\|_2^2+\bm{\lambda}^T|\bm{\alpha}|,
\end{equation}
where $\mathbf{D}$ and $\bm{\alpha}$ are same as those in (\ref{synthesisform}), $\mathbf{I}$ is the identity matrix, $\gamma>0$, $\bm{\lambda}$ is a given nonnegative weight vector, and $|\cdot|$ denotes the element-wise absolute operator. This formulation can be viewed as a combination of the analysis formulation and the synthesis formulation. Specifically, when $\gamma=0$, the problem (\ref{balanceform}) is turned into the synthesis scheme, and while $\gamma=\infty$, (\ref{balanceform}) is then reduced as the analysis problem. Therefore, by taking $0<\gamma<\infty$, this formulation balances the sparsity of the coefficients $\bm{\alpha}$ and the smoothness of the reconstructed image. To solve (\ref{balanceform}),\citet{xie2012alternating} proposed a fast ADMM-type algorithm by exploiting the special structures of $\mathbf{H}$ and $\mathbf{D}$. In \citet{shen2011accelerated}'s work, it is noted that the problem in (\ref{balanceform}) is not strictly convex, and thus the optimal solution is not unique. The authors addressed this issue by adding an additional regularizer $\|\bm{\alpha}\|_2^2$ to constrain the solution space, and an accelerated proximal gradient algorithm was developed to solve the problem.

\section{Sparse Representation-based Methods} \label{sparserepmethod}

Sparse representation accounts for a decomposition that represents a signal $\mathbf{x}\in\mathbb{R}^N$ as a sparse linear combination of basis atoms $\mathbf{D}_m\in\mathbb{R}^N(m=1,...,M)$. Given an over-complete dictionary $\mathbf{D}=[\mathbf{D}_1,...,\mathbf{D}_M]$ where $N\ll M$, the underlying sharp image $\mathbf{x}$ in (\ref{vectorform}) is sparsely represented as follows
\begin{eqnarray}
&\mathbf{x}=&\mathbf{D}\bm{\alpha},\nonumber\\
&s.t.&\|\bm{\alpha}\|_0\ll N,\nonumber\\
&&\|\mathbf{y}-\mathbf{Hx}\|_2<\epsilon,
\end{eqnarray}
where $\bm{\alpha}\in\mathbb{R}^M$, $\epsilon>0$ and the second constraint stems from the Gaussian noise assumption. Taking into consideration the noise and the representation error, the above problem can be formulated as
\begin{equation}\label{sparserepresentcom}
\min_{\bm{\alpha}}\|\mathbf{y}-\mathbf{HD}\bm{\alpha}\|_2^2+\lambda\|\bm{\alpha}\|_0.
\end{equation}
Since solving this $\ell_0$ regularized problem is NP-hard and computationally prohibitive \citep{bruckstein2009sparse}, approximation algorithms are usually considered. A general scheme is to replace the $\ell_0$ with an $\ell_1$ norm, leading to a reformulation of (\ref{sparserepresentcom}) as
\begin{equation}\label{sparsereprensentl1}
\min_{\bm{\alpha}}\|\mathbf{y}-\mathbf{HD}\bm{\alpha}\|_2^2+\lambda\|\bm{\alpha}\|_1,
\end{equation}
which can be solved by \emph{basis-pursuit} (BP) \citep{chen1998atomic} or LASSO \citep{tibshirani1996regression}. More general optimization methods for (\ref{sparsereprensentl1}) are reviewed in \citep{zibulevsky2010l1}.

\subsection{Smooth Enhancement}

The sparse representation framework has been successfully applied to various image processing tasks, such as denoising \citep{elad2006image}, deblurring \citep{cai2009blind}, inpainting \citep{elad2005simultaneous}, and super-resolution \citep{yang2010image}. In these approaches, the sign $\mathbf{x}$ in the above equations typically represents a patch in the image. Mathematically, the minimization over the whole image can be written as
\begin{equation}\label{patchsparse}
\min_{\bm{\alpha}}\sum_i\|\mathbf{R}_i\mathbf{y}-\mathbf{HD}\bm{\alpha}_i\|_2^2+\lambda\sum_i\|\bm{\alpha}_i\|_1,
\end{equation}
where $\bm{\alpha}=[\bm{\alpha}_1,...,\bm{\alpha}_K]$, $K$ is the total number of patches, $\mathbf{R}_i$ denotes the extraction of the patch at location $i$, and the patch $\mathbf{x}_i=\mathbf{D}\bm{\alpha}_i$. To reconstruct the sharp image, the estimated patches need to be fused. A general operation is to chop the image into overlapping patches, which are then averaged to calculate each pixel's value. However, as noted by \citet{cai2012framelet}, a crude fusing scheme in the synthesis approach often leads to visible artifacts along the image edges, exhibiting less smoothness in the reconstructed image.

To suppress the artifacts and enhance the smoothness, additional constraints have been proposed that involve kinds of formations. For example, \citet{ma2013dictionary} added a TV regularizer about the estimated image into the sparse formulation, since the TV regularization in the analysis approaches has the effect of enhancing edges and restraining the changes in non-edge regions. Based on the Poisson noise assumption, the objective (\ref{patchsparse}) changes to
\begin{eqnarray}
&\min\limits_{\bm{\alpha},\mathbf{x},\mathbf{D}}&\gamma<\mathbf{Hx}-\mathbf{y}\log\mathbf{Hx},\mathbf{1}> +\sum_i\|\mathbf{R}_i\mathbf{x}-\mathbf{D}\bm{\alpha}_i\|_2^2\nonumber\\
&&+\sum_i\mu _i\|\bm{\alpha}_i\|_0 + \eta\|\nabla\mathbf{x}\|_1,
\end{eqnarray}
where $\gamma$ and $\eta$ balance the corresponding terms, and $<\cdot,\cdot>$ denotes the inner product. An algorithm based on the variable splitting method is also proposed to iteratively solve different unknown variables.

Inspired by the superiority of nonlocal strategy over the local estimation strategy, \citet{dong2011image} introduced a combination of two adaptive regularizers, one of which involves a set of \emph{auto-regression} (AR) models $\{\mathbf{a}_1,...,\mathbf{a}_L\}$ characterizing the local structures, while the other encodes the nonlocal similarity. The resultant formulation is
\begin{eqnarray}\label{ARnonlocal}
\min_{\bm{\alpha}}&&\sum_i\|\mathbf{R}_i\mathbf{y}-\mathbf{H}\mathbf{D}\bm{\alpha}_i\|_2^2+\sum_i\lambda_i\|\bm{\alpha}_i\|_1\nonumber\\
&&+\underbrace{\gamma\sum_i\|x_i-\mathbf{a}_l^T\mathbf{x}_{\backslash i}\|_2^2}_{\text{AR regularization}}+\underbrace{\eta\sum_i\|x_i-\mathbf{b}_i^T\bm{\beta}_i\|_2^2}_{\text{Nonlocal regularization}},
\end{eqnarray}
where $\{\lambda_i\}$, $\gamma$, $\eta$ are regularization parameters. For the $i$-th patch, $x_i$ and $\mathbf{x}_{\backslash i}$ are the central pixel and the non-central pixels respectively, $\mathbf{a}_l$ denotes the parameters of the selected $l$-th AR model, $\bm{\beta}_i$ collects the central pixels of similar patches searched across the whole image, and $\mathbf{b}_i$ lists the corresponding nonnegative weights. The elements of $\mathbf{b}_i$ are required to sum to $1$, i.e. $\sum_j\mathbf{b}_{i,j}=1$. This scheme considers both local smoothness and nonlocal structure similarity, which in cooperation with sparse representation, will provide an accurate estimation of the pixel by excavating as much of the information in the image as possible. In \citet{dong2011centralized,dong2013nonlocally}'s subsequent work, the same idea of nonlocal regularization is employed. Rather than directly estimating the central pixel according to the nonlocal similarity, the authors impose this regularization onto the sparse codes $\bm{\alpha}$ by replacing the last term in (\ref{ARnonlocal}) with
\begin{equation}\label{centralsparse}
\sum_i\|\bm{\alpha}_i-\mathbf{A}_i\bm{\omega}_i\|_p,
\end{equation}
where $p=1$ or $2$, the columns of $\mathbf{A}_i$ correspond to the sparse codes of the similar patches which are collected via block matching over all extracted patches, and $\bm{\omega}_i$ is the same weight vector as $\mathbf{b}_i$ in (\ref{ARnonlocal}). The resultant model is named \emph{centralized sparse representation}. The rationale behind this nonlocal regularization comes from an empirical investigation, showing that the sparse codes $\bm{\alpha}^e$ calculated according to (\ref{patchsparse}) usually have sparse coding noise with respect to the true codes $\bm{\alpha}^t$, i.e. $\bm{\alpha}^e=\bm{\alpha}^t+\mathbf{n}_{\bm{\alpha}}$ where $\mathbf{n}_{\bm{\alpha}}$ is the noise that can be approximately characterized by Laplace distribution. The authors proposed using $\mathbf{A}_i\bm{\omega}_i$ to approximate $\bm{\alpha}^t$, and thus minimizing the regularization in (\ref{centralsparse}) would restrain the noise emerging in the optimization procedure.

\subsection{Dictionary Learning} \label{dictionarylearning}

The selection of the dictionary $\mathbf{D}$ is crucial in the sparse representation-based methods, since a suitable $\mathbf{D}$ will result in low reconstruction error, otherwise high error will be caused. A classic selection is the predefined regular dictionary, such as the redundant discrete cosine transform (DCT) \citep{guleryuz2006nonlinear1,guleryuz2006nonlinear2} and the overcomplete Haar dictionary. These dictionaries can be pre-calculated before the optimization of problem (\ref{sparserepresentcom}) or (\ref{sparsereprensentl1}), saving a large amount of computational source, but these choices provide limited performance due to the ignoring of the real data.

Alternatively, a task-specific dictionary learning strategy has been proven to be advantageous. A typical example is the K-SVD \citep{elad2006image,aharon2006svd}, in which the redundant dictionary is leaned from a training set of high quality images or the currently processed degraded images. In this method, the optimization alternates between solving the sparse codes $\bm{\alpha}$ and the dictionary $\mathbf{D}$, thus increasing the computational cost compared with using a predefined regular dictionary. Fortunately, the performance of K-SVD is enhanced, and it is also shown that the scheme in which the dictionary is learned from the currently processed image is superior to other schemes that utilize a training set to learn the dictionary. This highlights the importance of task-specificity.

The K-SVD algorithm is utilized by \citet{ma2013dictionary} to update the dictionary in each iteration for debluring. Besides, \citet{perrone2012image} used a dictionary $\mathbf{D}$ composed of two parts, one of which consists of blurry patches extracted from the current processed image $y$, denoted as $\mathbf{D}_y$, while another collects the patches from a sharp dataset and corrupts them using the known blur kernel, denoted as $\mathbf{D}_d$. Thus we can write $\mathbf{D}=[\mathbf{D}_y,\mathbf{D}_d]$, corresponding to the self-similarity-based and dataset-based dictionaries, respectively. Instead of solving the $\ell_1$ optimization problem, the authors computed the representation of the blurry image on $\mathbf{D}$ by employing the nonlocal mean method and a subsequent refinement. This representation, denoted as $\bm{\alpha}^{nlm}$, is then regarded in a variational formulation as a constraint on the sharp image. The problem is written as
\begin{eqnarray}
&\min\limits_{\mathbf{x},\mathbf{n},\mathbf{e}}&\frac{1}{2}\|\mathbf{x}-\mathbf{D}\bm{\alpha}^{nlm}\|_2^2
+\lambda\|\nabla\mathbf{x}\|_2+\frac{\eta}{2}\|\mathbf{n}\|_2^2+\gamma\|\mathbf{e}\|_1,\nonumber\\
&s.t.& \mathbf{y}=\mathbf{Hx}+\mathbf{n}+\mathbf{e},
\end{eqnarray}
where $\lambda$, $\eta$, $\gamma$ balance the different terms, and $\mathbf{e}$ denotes the error introduced by the inaccurate blur kernel. The effectiveness of this formulation is experimentally demonstrated when a noisy blur kernel is assumed.

Generally, the local structures of the natural images exhibit a certain level of correlations and can be synthesized by similar atoms in a redundant dictionary. By intentionally arranging the dictionary atoms, the estimated sparse patterns will be constrained to a specific subset of atoms. This is the so-called structured sparse representation. By exploring structured sparse representation, \citet{dong2011centralized,dong2011image,dong2013nonlocally} introduced a PCA-based dictionary learning method. In the training phase, a large number of patches are extracted from an additional dataset of natural images, followed by a clustering procedure to create a set of clusters (say $K$ clusters), each of which encodes a certain type of local structure. Then PCA is applied to each cluster (denoted by its centroid $\bm{\mu}_k$, $1\leq k\leq K$), in which all the affiliated patches are vectorized. The calculated principal components, which encode the structural information, are utilized as the subdictionary (denoted as $\mathbf{D}_k$) for the corresponding cluster. Thus the final dictionary is a concatenation of all subdictionaries, i.e. $\mathbf{D}=[\mathbf{D}_1,...,\mathbf{D}_K]$. In the sparse coding phase, a patch $\mathbf{x}$ is adaptively assigned a subdictionary whose index is
\begin{equation}\label{pcasparse}
k^*=\argmin_k\|\tilde{\mathbf{D}}_k\mathbf{x}-\tilde{\mathbf{D}}_k\bm{\mu}_k\|_2,
\end{equation}
where $\tilde{\mathbf{D}}_k$ collects the first several most significant components in $\mathbf{D}_k$. Following the selection of the subdictionary, the sparse codes of $\mathbf{x}$ are solved by optimizing (\ref{sparsereprensentl1}). Since $\mathbf{x}$ is degraded by blur and noise, the selection in (\ref{pcasparse}) may be not optimal. The authors proposed to iteratively implement (\ref{pcasparse}) and (\ref{sparsereprensentl1}) until the estimation of $\mathbf{x}$ converged. This PCA-based structured sparsity is shown to be equivalent to \emph{piecewise linear estimation} (PLE) under the GMM assumption \citep{yu2012solving}. By contrast, a MAP-EM algorithm is proposed in \citep{yu2012solving} to solve the PLE problem instead of using PCA, showing that PLE can stabilize and improve traditional nonlinear sparse inverse problems.

\subsection{Combined Tasks under Sparse Representation}

Image deblurring is a low-level task aiming to produce high quality images for the subsequent high-level vision tasks. Several recent sparse representation-based methods combine the deblurring problem and other elements, such as compressed sensing \citep{amizic2013compressive,rostami2012image} and object recognition \citep{zhang2011close}, into a unified scheme. In \citet{amizic2013compressive}'s work, the blind image deconvolution technique is integrated into the compressed sensing (CS)-based imaging system. In this system, a blurry image is represented as
\begin{equation}
\mathbf{y}=\mathbf{M}\tilde{\mathbf{x}}+\mathbf{n}=\mathbf{MHx}+\mathbf{n},
\end{equation}
where $\mathbf{M}$ is the CS measurement matrix, and $\tilde{\mathbf{x}}=\mathbf{Hx}$. To sample the signal $\tilde{\mathbf{x}}$, CS involves the following optimization problem:
\begin{equation}
\min_{\bm{\alpha}}\|\mathbf{y}-\mathbf{MW}\bm{\alpha}\|_2^2+\tau\|\bm{\alpha}\|_1,
\end{equation}
which is similar to (\ref{sparsereprensentl1}), and $\mathbf{W}$ denotes the redundant transformation by which $\tilde{\mathbf{x}}=\mathbf{W}\bm{\alpha}$ can be sparsely represented. Combined with the variational framework for blind image deblurring in (\ref{blindvariational}), a unified objective function can be obtained
\begin{eqnarray}\label{sparsevariational}
&\min\limits_{\mathbf{x},\mathbf{H},\bm{\alpha}}&\|\mathbf{y}-\mathbf{MW}\bm{\alpha}\|_2^2+\tau\|\bm{\alpha}\|_1
+\lambda_x\mathrm{\Psi}_x(\mathbf{x})+\lambda_h\mathrm{\Psi}_h(\mathbf{H}),\nonumber\\
&s.t.& \mathbf{Hx}=\mathbf{W}\bm{\alpha}.
\end{eqnarray}
As noted, the advantage of (\ref{sparsevariational}) compared to the sequential approach (a two-step approach in which CS is followed by blind image deblurring) is the capacity of imposing an additional structural constraint on the sparse codes $\bm{\alpha}$ because in the sequential case, $\bm{\alpha}$ is determined when completing CS and cannot be changed in the subsequent deblurring procedure. \citet{rostami2012image} introduced a derivative compressed sensing (DCS) technique to derive the generalized pupil function (GPF) of the short-exposure optical system, where the GPF can uniquely determine the PSF introduced by air turbulence. Once the estimation of PSF via DCS is completed, a non-blind deconvolution procedure under the variational framework is conducted to produce the restored optical images.

The performance of face recognition heavily depends on the quality of the captured face images. When images are degraded by blur or noise, the extracted features may be significantly disturbed. To address this issue, \citet{zhang2011close} proposed a framework to simultaneously blind-restore the face images and recognize the face labels. The whole framework stems from the sparse representation-based classification (SRC) \citep{wright2009robust} which assumes that the data samples belonging to the same class lie in the same low-dimensional subspace. Different from SRC, the target image to be labeled in this method is a blurry face image, and the results include an estimated kernel, a restored image and its associated label. Formally, the proposed objective function is
\begin{eqnarray}
&\min\limits_{\mathbf{x},\mathbf{H},\bm{\alpha},c}&\|\mathbf{y}-\mathbf{Hx}\|_2^2+\eta\|\mathbf{x}-\mathbf{D}\bm{\alpha}\|_2^2
+\tau\|\bm{\alpha}\|_1\nonumber\\
&&+\lambda_x\mathrm{\Psi}_x(\mathbf{x})+\lambda_h\mathrm{\Psi}_h(\mathbf{H}),
\end{eqnarray}
where the dictionary $\mathbf{D}$ includes a set of $C$ subdictionaries, each of which contains the training face images of one specific class, i.e. $\mathbf{D}=[\mathbf{D}_1,...,\mathbf{D}_C]$, and $c$ is the class label embedded in $\mathbf{D}$ and $\bm{\alpha}$. In the above formulation, $\mathbf{x}$, $\mathbf{H}$, $\bm{\alpha}$, $c$ are optimized alternatively. It should be noted that $\mathbf{D}$ is realized as the collection of sharp training images when optimizing $\mathbf{x}$ and $\mathbf{H}$, whereas in the estimation of $\bm{\alpha}$ and $c$, $\mathbf{D}$ is corrupted by the intermediate estimation of $\mathbf{H}$ to generate the blurry dictionary $\mathbf{D}^b$, i.e. $\mathbf{D}^b=\mathbf{HD}$. Compared with the state-of-the-art blind deblurring algorithms and SRC, this framework obtains a significant performance improvement.

\section{Homography-based Methods} \label{homographymethod}

In this section and the next section, our discussion will mainly focus on the modeling and associated processing of the spatial variant blur effect.

\emph{Homography-based modeling} is generally proposed to simulate the blur effect induced by the camera's egomotion or camera shake. Recall that in the image formation process, a 3D scene point $(u,v,m)$ is mapped to a 2D image plane point $(i,j)$, which can be formulated in the homogeneous coordinates as
\begin{equation}
(i_t,j_t,1)^T=\mathcal{P}_t(u,v,m,1)^T,
\end{equation}
where $t$ denotes the time index. In the case of camera motion, $\mathcal{P}_t$ may vary with time as a function of camera translation and rotation, causing a fixed point in the scene to be projected onto different locations in the image plane at each time. As is well-known, when using a pinhole camera, all views seen by the camera are projectively equivalent except for the boundaries \citep{whyte2010non,joshi2010image}. This means that for a static scene with constant depth, the 2D images projected at different instances of time are related via a \emph{homography}. Denoting the image point at time $t=0$ as $(i_0,j_0)$, the homography and projected point at time $t$ are modeled as
\begin{eqnarray}
&&\bm{\mathcal{H}}_t(d)=\mathbf{K}\left(\mathbf{R}_t+\frac{1}{d}\mathbf{T}_t\mathbf{N}^T\right)\mathbf{K}^{-1},\\
&&(i_t,j_t,1)^T=\bm{\mathcal{H}}_t(d)(i_0,j_0,1)^T,
\end{eqnarray}
for a particular depth $d$, where $\mathbf{K}$ is the camera's internal calibration matrix, $\mathbf{R}_t$ and $\mathbf{T}_t$ are the rotation matrix and translation vector at time $t$, and $\mathbf{N}$ is the unit vector orthogonal to the image plane. Based on this formulation, the image captured at any time is expressed by the initial image $x_0$, i.e.
\begin{equation}\label{homographyfunc}
x_t(i,j)=x_0(\bm{\mathcal{H}}_t(d)(i,j,1)^T).
\end{equation}
For simplicity, we express the coordinates as a column vector $\bm{i}$ and the homography as $\bm{\mathcal{H}}_t$, and then (\ref{homographyfunc}) can be rewritten as
\begin{equation}\label{homographyfuncsimp}
x_t(\bm{i})=x_0(\bm{\mathcal{H}}_t\bm{i}).
\end{equation}
Thus we can define the blurry image $y$ as the accumulated result over the exposure duration $\tau$, which is given by
\begin{equation}\label{homographyint}
y(\bm{i})=\int_{t=0}^{\tau}x_0(\bm{\mathcal{H}}_t\bm{i})dt,
\end{equation}
where we omit the noise term. We rewrite (\ref{homographyfuncsimp}) in the matrix-vector form as
\begin{equation}\label{homographyfuncvec}
\mathbf{x}_t=\mathbf{H}_t\mathbf{x}_0,
\end{equation}
where $\mathbf{H}_t$ is a sparse resampling matrix that implements the image warping and resampling due to the homography, and then (\ref{homographyint}) becomes
\begin{equation}\label{homographyintsimp}
\mathbf{y}=\int_{t=0}^{\tau}\mathbf{H}_t\mathbf{x}_0dt.
\end{equation}
However, due to the successive duration $[0,\tau]$ in (\ref{homographyint}) and (\ref{homographyintsimp}), infinite instantiations of $\bm{\mathcal{H}}_t$ will be created, causing the ill-posedness of the problem. To handle this issue, two methods can be employed: one which discretizes the time duration and one which decomposes  $\mathbf{H}_t$ into the basis transformations.

\subsection{Time Discretization}

If we suppose that the duration $\tau$ is segmented to $N$ equivalent periods, in each of which the homography $\bm{\mathcal{H}}$ is approximately consistent, the equation (\ref{homographyint}) becomes
\begin{equation}\label{timediscrethomo}
y(\bm{i})\approx\frac{1}{N}\sum_{t=1}^Nx_0(\bm{\mathcal{H}}_t\bm{i}).
\end{equation}
The left hand side of (\ref{timediscrethomo}) is equal to the right hand side when $N\rightarrow\infty$. If $N$ is set to a limited value, the unknown variables in $\{\bm{\mathcal{H}}_t\}$ will be well-constrained, reducing the ill-posedness of the problem. Equation (6.9) is nothing but the \emph{projective motion blur} model proposed by \citet{tai2011richardson}. To estimate each homography, the motion is assumed to be uniform in the exposure time. Each $\bm{\mathcal{H}}_t$ can therefore be computed directly according to the whole homography generated from $t=0$ to $\tau$, i.e. $\bm{\mathcal{H}}_t=\sqrt[N]{\bm{\mathcal{H}}_{[0,\tau]}}$. Based on this formulation, \citet{tai2011richardson} also developed the projective motion Richardson-Lucy algorithm which has been mentioned in Section \ref{maximumlikelihood}. Following this model, \citet{tai2010coded} proposed a coded exposure technique to introduce discontinuities during the exposure period for the purpose of reducing the discretization error existing in equation (\ref{timediscrethomo}), while at the same time preserving the high-frequency spatial information. The captured image provides more details for the recovery of the blur kernel. In conjunction with the matting technique, the moving object in the image is isolated and deblurred appropriately.

A strong assumption in equation (\ref{timediscrethomo}) is the consistent homography in each equivalent period, which may not fit to the reality. A more general formulation is
\begin{equation}\label{adaptweighthomo}
y(\bm{i})\approx\sum_{t=1}^Nw_tx_0(\bm{\mathcal{H}}_t\bm{i}),
\end{equation}
where $w_t$ denotes the proportion of the period occupied by $\bm{\mathcal{H}}_t$, and $\sum_tw_t=1$. Under this model, \citet{cho2012registration} proposed a registration-based method to estimate the homographies $\{\bm{\mathcal{H}}_t\}$ and the weights $\{w_t\}$ by using the \emph{Lucas-Kanade} algorithm \citep{szeliski1997creating}. This estimation method is extended into a multiple image deblurring scheme.

\subsection{Homography Decomposition} \label{homographydecomposition}

This strategy decomposes the homography into a set of basic operations, i.e. representing $\mathbf{H}_t$ in (\ref{homographyfuncvec}) and (\ref{homographyintsimp}) as a weighted sum of \emph{predefined} transformations or homographies. This leads to the formulation as
\begin{equation}\label{homographydecompfunc}
\mathbf{y}=\sum_{l=1}^Lw_l\mathbf{H}_l\mathbf{x}_0,
\end{equation}
where $\{\mathbf{H}_l\}$ denotes the basis set with cardinality of $L$, $w_l$ is the weight assigned to the $l$-th basis, and $\sum_lw_l=1$.

A typical example is the \emph{transformation spread function} (TSF)-based modeling proposed by \citet{chandramouli2010inferring}. Their method focuses on the modeling of the camera's translations and in-plane rotations. $\{\mathbf{H}_l\}$ is assumed to be the set of possible geometric transformations the image points can undergo during the exposure. TSF $w(\mathbf{H}_l)$ is defined as
\begin{equation}
w(\mathbf{H}_l):\{\mathbf{H}_l\}\rightarrow\mathbb{R}_+,
\end{equation}
and (\ref{homographydecompfunc}) is rewritten as
\begin{equation}\label{adaptweighthomodecomp}
\mathbf{y}=\sum_{l=1}^Lw(\mathbf{H}_l)\mathbf{x}_{\mathbf{H}_l},
\end{equation}
where $\mathbf{x}_{\mathbf{H}_l}=\mathbf{H}_l\mathbf{x}_0$ denotes the reference image $\mathbf{x}_0$ warped by the transformation $\mathbf{H}_l$. A direct understanding of this formulation is similar to that in (\ref{adaptweighthomo}), i.e. the transformed images are weighted according to their exposure time. Depending on (\ref{adaptweighthomodecomp}), Chandramouli et al. derived the relationship between TSF and the spatial variant PSF. Let $h(\bm{i};\bm{u})$ denote the PSF where $\bm{i}$ is the location in the image plane and $\bm{u}$ is the position in the 2D PSF. Then
\begin{equation}
h(\bm{i};\bm{u})=\sum_{l=1}^Lw(\mathbf{H}_l)\delta(\bm{u}-(\bm{i}-\bm{i}_l)),
\end{equation}
in which $\delta(\cdot)$ is the Kronecker delta function, and $\bm{i}_l$ denotes the coordinate of the point projected from $\bm{i}$ according to $\mathbf{H}_l$. Applying this model to the deblurring problem, TSFs $w(\cdot)$ are estimated by solving a least-squares problem. \citet{vijay2013non} subsequently employed TSF-based modeling in the high dynamic range (HDR) image reconstruction problem which involves blurry/noisy image pairs or multiple blurry images. In their formulation, the TV regularization and the image pair collaborative constraint are utilized to ensure the smoothness and recoverability of fine details. Note that the equation (\ref{adaptweighthomodecomp}) is limited to the situation of constant depth \citep{chandramouli2010inferring}. If there are depth variations in the scene, the blurry effect cannot be modeled using (\ref{adaptweighthomodecomp}) where only one TSF is involved. \citet{paramanand2013non} proposed a method involving two TSFs to handle the blurry image by capturing a bilayer scene which consists of two dominant layers of different depths. The two TSFs, each of which corresponds to a certain depth, are estimated as follows. Local blur kernels are first calculated at different image locations, followed by a grouping operation on the kernels according to the depth of each location. Each TSF is then estimated from the blur kernels in the corresponding depth layer under a regularization framework.

Another scheme for setting the basis set $\{\mathbf{H}_l\}$ was proposed by \citet{whyte2010non,whyte2012non}. Considering all-directional rotation of the camera, the homographies are correlated with the camera's orientation. The resultant formulation is
\begin{equation}
y(\bm{i})=\int_{\mathrm{\Theta}}w(\theta)x_0(\bm{\mathcal{H}}_{\theta}\bm{i})d\theta,
\end{equation}
and its discretized version is
\begin{equation}
y(\bm{i})=\sum_{\theta}w(\theta)x_0(\bm{\mathcal{H}}_{\theta}\bm{i}),
\end{equation}
where $\bm{\mathcal{H}}_{\theta}$ is the homography specifying the orientation $\theta$ of the camera rotation, and $w(\theta)$ denotes the weighting function of $\theta$. By setting $\{\bm{\mathcal{H}}_{\theta}\}$ according to the rotations along the three Cartesian axes, the spatially variant blur kernel can be well-approximated. This model is applied to a single image deblurring problem using the variational Bayesian method, as well as a blurry/noisy image pair deblurring problem under the regularized least-squares formulation. Note that an integral over $\mathrm{\Theta}$ implies a search over the full orientation space, incurring a high computational cost when the camera rotates significantly. Instead, \citet{hu2012fast} proposed to constrain the camera poses by imposing an initial guess of the pose subspace and searching the optimal solution within this subspace. This initialization of the pose subspace is implemented using back-projection. Compared with the method in \citep{whyte2010non,whyte2012non}, \citet{hu2012fast}'s method produces more favorable results.

\emph{Motion density function} (MDF) proposed by \citet{gupta2010single} is similar to the above two schemes, and has the formulation of (\ref{homographydecompfunc}). Here the basis set $\{\mathbf{H}_l\}$ is called the \emph{motion response basis} (MRB) while $\{w_l\}$ is called MDF. In the method, MRB is again pre-computed according to both camera translation and rotation, while the sharp image $\mathbf{x}_0$ and the MDF $\{w_l\}$ are iteratively optimized using the proposed RANSAC (short for "Random Sample consensus")-based scheme. \citet{zheng2013forward} applied this type of method to a more practical scenario that is forward/backward motion blur removal.

\section{Region-based Methods} \label{regionmethod}

The spatial variant blur kernel implies that the local kernels vary from region to region, or more strictly, from point to point. Region-based methods express the blur model in that each region, or ideally, each point is assigned to a local consistent kernel \citep{levin2006blind}. The local kernel at the location $i$ of the image is denoted as $h_i$, and the windowing operation extracting the fixed size patch whose center is located at $i$ is denoted as $\omega_i$. The blurry image is then modeled as
\begin{equation}\label{regionformulation}
y_i=c(h_i*(\omega_i\odot x)),
\end{equation}
where $c(\cdot)$ is a function to extract the center point of the patch. Thus the set $\{h_i\}$ forms the spatial variant blur model. Note that in equation (\ref{regionformulation}), only the correspondence between $y_i$ and the center of the patch $h_i*(\omega_i\odot x)$ is accessible due to the operation of $c(\cdot)$, and thus it is impractical to estimate $x$ and all $\{h_i\}$. By considering the redundancy information (in particular, the overlaps between sampled patches) in the modeling, the operator $c(\cdot)$ is generally replaced by the averaging of overlaps, which results in the \emph{efficient filter flow} (EFF) model proposed by \citet{hirsch2010efficient} and \citet{harmeling2010space},
\begin{equation}\label{regionformulationwhole}
y=\sum_{\{i\}}h_i*(\omega_i\odot x).
\end{equation}
In (\ref{regionformulationwhole}), for computational efficiency, the overlapping patches are only extracted at the locations indicated by the set $\{i\}$ (not all locations in the image). $\omega_i$ encodes not only the extracting operation, but also the averaging operation. Thus its values are not necessarily $0$ or $1$ but in the range of $[0,1]$, and furthermore the normalization constraint $\sum_{\{i\}}\omega_i = \mathbf{1}$ should be satisfied. Thanks to the matrix-vector-multiplication expression of convolution, equation (\ref{regionformulationwhole}) can be rewritten as $\mathbf{y}=\mathbf{Hx}$, where
\begin{equation}\label{EFFH}
\mathbf{H}=\mathbf{Z}_y^T\sum_{\{i\}}\mathbf{C}_i^T\mathbf{F}^H\mathrm{Diag}(\mathbf{F}\mathbf{Z}_hh_i)\mathbf{F}\mathbf{C}_i\mathrm{Diag}(\omega_i).
\end{equation}
Here $\mathrm{Diag}(v)$ is a diagonal matrix that has vector $v$ along its diagonal, $\mathbf{C}_i$ is the chopping matrix for the $i$-th patch in the image, $\mathbf{F}$ and $\mathbf{F}^H$ are respectively the Discrete Fourier Transform matrix and its Hermitian, $\mathbf{Z}_h$ is the zero-padding matrix used to expand the $h_i$ by zero to the patch size, and $\mathbf{Z}_y^T$ chops out the valid part of the space-variant convolution. Regarding equation (\ref{regionformulationwhole}) as $\mathbf{y}=\mathbf{Xh}$, in which the vector $\mathbf{h}$ is obtained by stacking $h_1,...,h_R$, $\mathbf{X}$ is given by
\begin{equation}\label{EFFX}
\mathbf{X}=\mathbf{Z}_y^T\sum_{\{i\}}\mathbf{C}_i^T\mathbf{F}^H\mathrm{Diag}(\mathbf{F}\mathbf{C}_i\mathrm{Diag}(\omega_i)x)\mathbf{F}\mathbf{Z}_h\mathbf{B}_i,
\end{equation}
where $\mathbf{B}_i$ denotes a matrix that chops the blur kernel $h_i$ from $\mathbf{h}$. Equations (\ref{EFFH}) and (\ref{EFFX}) provide an efficient way to obtain $\mathbf{H}$, $\mathbf{H}^T$, $\mathbf{X}$ and $\mathbf{X}^T$, which are essential to iteratively estimate $\{h_i\}$ and $x$. The experiments on processing of both atmospheric turbulence blur and camera shake blur show promising recovery accuracy and computational efficiency.

In their subsequent work, \citet{schuler2011non} apply this framework to the problem of correcting optical aberrations, which exploits three channels (i.e. RGB) of the captured image. The problem is formulated as a joint objective function that combines respective constraints on three channels. Additionally, by incorporating the idea discussed in Section \ref{homographydecomposition}, \citet{hirsch2011fast} and \citet{schuler2012blind} reformulated (\ref{regionformulationwhole}) to model the camera rotation blur by representing the local blur kernel as the weighted sum of a set of bases, i.e.
\begin{equation}\label{regionhomodecomp}
y=\sum_{\{i\}}(\sum_{k=1}^K\mu_kb_{k,i})*(\omega_i\odot x),
\end{equation}
where $\{\mu_k\}$ are the weights assigned to the bases $\{b_{k,i}\}$. It should be noted that $\mu_k$ is independent of the patch with index $i$, whereas $b_{k,i}$ depends that patch. This is because when modeling the camera rotation blur, different locations in the sensor may undergo very different routes, i.e. the local kernel in a specific point is expressible only by the basis kernels at the same point, each of which is induced by different rotation. Specifically, for modeling the camera rotation blur, we need to first generate the kernel maps, each of which consists of the local kernel for all points in an image, for different rotations, and then extract the local kernels at the same point, say $i$, of all maps to constitute the bases $\{b_{k,i}\}$. In such a situation, $\{\mu_k\}$ and $x$ are unknown variables in equation (\ref{regionhomodecomp}), whose number is significantly decreased. By employing the edge emphasizing operation in the pre-steps of optimization, the deblurring results on shaken blurred images and optical aberrant images are impressive. Apart from modeling camera rotation blur, this scheme is applicable to modeling general camera shaken blur, air turbulence blur, as well as defocus blur.

In contrast to the above EFF methods, \citet{ji2012two} proposed an intuitive approach to estimate point-wise blur kernels. In camera egomotion, the sensor is forced to undergo rigid translation or rotation. This rigidness naturally implies that without the consideration of depth change, the variation of the local kernels between two adjacent points is limited, meaning that if we have two kernels located in nearby positions, the other kernels located within the line between these two points can be recovered by a kernel interpolation scheme. Based on this strategy, \citet{ji2012two}'s method initially estimates the local kernels for overlapping regions using existing spatially invariant blind deblurring methods. However, because the scene depth varies in reality, some of the estimated kernels are erroneous and should be corrected according to the affine relationship between the adjacent regions. The authors then presented a kernel interpolation algorithm for transforming the region-wise kernels into point-wise kernels. Note that either kernel initialization or interpolation inevitably introduces errors to the kernel estimation. Based on \citep{ji2012robust}, the induced error term is considered in a non-blind deblurring model which is an analysis formulation (cf. equation (\ref{analysisform}) in Section \ref{analysissynthesis}).

Camera translational blur is spatially variant when capturing a scene with depth variation. \citet{xu2012depth} developed a model by discovering the depth information of the scene. The depth layers, each of which corresponds to a local region, are produced by a stereo-matching-based disparity estimation algorithm that involves two blurry images of different views. In the estimation of the blur kernel for each layer, a tree structure is constructed over all depth layers, where the nodes for two layers having neighboring depths are merged to form their parent node. The blur kernels are then estimated from the top-level of the tree to the bottom-level, in which the kernel refinement is operated to make the estimation robust. Once the kernels are estimated, the two images are simultaneously deblurred in a unified formulation. To produce a pleasant result, the whole process from disparity estimation to blur removal is required to repeat at least once.

In terms of object motion blur, the blurry image can typically be segmented into two regions, where one is for the moving object inducing the blur and the other is generally the static background that has probably been corrupted by defocus blur. In this case, \citet{chakrabarti2010analyzing} determined the spatially variant blur model by mining the statistics of the localized frequency representation, in which each local region of the blurry image is independently transformed into the frequency domain. To reduce the number of unknown parameters, pixels in one region correspond to a same local kernel. To handle the boundary effects in convolution, the statistics, in particular the first- and second-order moments, are involved in a $\text{MAP}_k$ formulation. Solving the $\text{MAP}_k$ provides the estimation of the blur kernel. What follows is a binary segmentation of the motion-blurred foreground from the background via a Markov Random Field modeling strategy. The non-blind deblurring procedure is conducted independently in each region. Similarly, \citet{kim2013dynamic} addressed to deblur an image which contains multiple moving objects. Each object occupies a region and corresponds to a single kernel. In this method, the blur segmentation, the kernel estimation and the image restoration are alternatively processed in a unified variational formulation (cf. equation (\ref{blindvariational}) in Section \ref{variationalmethod}).

\section{Others} \label{others}

In this section, we will discuss several methods which cannot be grouped into the above categories.

\textbf{\emph{Projection-based method}}: The projection-based methods involve two terms, one of which concerns knowledge about the true solution that can be incorporated into the prior constraint set, while the other encodes the noisy blurry image that specifies the observation constraint set. Similar to the variational framework, this type of method also has the objective of (\ref{unifyvariational}), containing the above two terms. Specifically, \citet{li2011fine} proposed the following weighted deblurring function:
\begin{equation}\label{projectionobjective}
\min_{\mathbf{x}}\frac{1}{2}\sum_{i=0}^r\|\mathcal{B}_i(\mathbf{y}-\mathbf{Hx})\|_2^2+\lambda\mathrm{\Psi}(\mathbf{x}),
\end{equation}
where $r$ is a positive integral parameter, and the operator $\mathcal{B}_i$ is defined as
\begin{equation}
\mathcal{B}_i=(\mathbf{I}-\beta\mathbf{H}\mathbf{H}^*)^i,\text{ }i=0,...,r,
\end{equation}
and $0<\beta<1$. The first term in (\ref{projectionobjective}) forms a variant of the problem in the traditional Landweber-iteration algorithm \citep{landweber1951iteration} which is used to deal with the ill-posed linear inverse problem, and can be solved by the $r$-times Landweber iteration. As noted by Li, the introduction of the $r$ operators $\{\mathcal{B}_i\}$ has three advantages. The first is that in the variational framework, the estimated solution is sensitive to the setting of the regularization parameter $\lambda$. However in (\ref{projectionobjective}), the degree of regularization is scaled by a factor of $1/r$, because of the addition of $\{\mathcal{B}_i\}$. Thus we can adjust $r$ instead of $\lambda$ to control the impact of regularization. Second, operators $\{\mathcal{B}_i\}$ generalize the definition of deblurring error which is determined by $r$. When $r=0$, the functional (\ref{projectionobjective}) reduces to the standard variational formulation. When $r\rightarrow\infty$, (\ref{projectionobjective}) becomes the standard Landweber iteration without regularization. Therefore by setting $r\in(0,\infty)$, the global minimum can fall everywhere except the two extremes. The third advantage is that based on the eigenvalue analysis, $\{\mathcal{B}_i\}$ can be deemed as a weighting scheme for improving the numerical stability of the Landweber iteration. To solve (\ref{projectionobjective}), Li proved that the minimization can be realized by a concatenation of $r$-times Landweber iteration (projection on the observation constraint set) and regularized filtering (projection on the prior constraint set). Formally, denoting the projection operator of $r$-times Landweber iteration as $\mathcal{P}_l$ and the projection operator of regularized filtering as $\mathcal{P}_r$, the optimization is then iterated between $\mathbf{x}^{k+1/2}=\mathcal{P}_l\mathbf{x}^k$ and $\mathbf{x}^{k+1}=\mathcal{P}_r\mathbf{x}^{k+1/2}$ until convergence.

\textbf{\emph{Kernel regression}}: Kernel regression techniques have been widely used in image denoising. However, for deblurring, this type of method has received relatively less attention. To understand how a kernel regression framework handles the ill-posedness, local similarity is conventionally investigated to ensure the smoothness of local structures, whereas more recently, the focus has moved to non-local kernels that can efficiently exploit the repetitive patterns in the blurry image. Mathematically, kernel regression computes the central pixel of a window as follows:
\begin{equation}\label{kernelregressionfunc}
\mathbf{x}_{\bm{i}}^*=\mathop{\argmin}\limits_{\mathbf{x}_{\bm{i}}}\sum_{\bm{j}\in\mathcal{N}(\bm{i})}(\mathbf{y}_{\bm{j}}-\mathbf{x}_{\bm{i}})^2\bm{K}_{\bm{i}}(\bm{j}-\bm{i}),
\end{equation}
where $\bm{i}$ and $\bm{j}$ are the locations, $\mathbf{x}_{\bm{i}}$ and $\mathbf{y}_{\bm{j}}$ are the corresponding pixels, $\mathcal{N}(\bm{i})$ denotes the neighbors of location $\bm{i}$, and $\bm{K}_{\bm{i}}(\cdot)$ is a generic spatial kernel at $\bm{i}$ which typically assigns large weights to the nearby similar pixels while assigning small weights to the farther dissimilar pixels. To apply (\ref{kernelregressionfunc}) to deblurring, \citet{takeda2008deblurring} derived a formulation for kernel-based deblurring through exploiting the Taylor expansion of local structures. In their work, two types of locally adaptive kernel, i.e. bilateral kernel function and steering kernel function, are utilized in the proposed framework. The bilateral kernel function considers both the similarity of pixel values and the distance of spatial locations, while the steering kernel function is adaptive to the local gradients. \citet{zhang2013image} extended the local kernel regression to the nonlocal case, in which the kernel $\bm{K}(\cdot)$ encodes the nonlocal similarity of high orders by searching similar patches across the whole image and calculating the zero-, first- and second-order similarities. In their formulation, both local kernel and nonlocal kernel are incorporated since the local structures regularize the noisy candidates found by the nonlocal similarity search, and nonlocal similarity provides the redundancy that prevents possible overfitting of the local kernel regression. In the application of deblurring, the authors pre-deconvolved the image using Wiener filtering, which amplifies the effect of the noise. The proposed nonlocal kernel regression is then employed to suppress the noise, resulting in a deblurred image.

\textbf{\emph{Stochastic deconvolution}}: It is not easy to develop specific optimization algorithms to solve the deblurring problems defined under the Bayesian framework (or the variational framework) due to the complex priors (or the regularizers). To derive a general-purpose deconvolution algorithm, \citet{hullin2013stochastic} proposed stochastic deconvolution, which is not only capable of effectively handling arbitrary priors, but also of tackling the boundary conditions, saturated pixels and spatial variant kernels. Stochastic deconvolution is based on the random walk optimization strategy from Stochastic Tomography \citep{gregson2012stochastic}, in which a stochastic coordinate-descent method employs a Metropolis-Hasting style heuristic for picking the next coordinate axis to descend along. Specifically, this strategy picks a single pixel in each iteration and checks whether the objective can be improved by adding (or removing) an energy quantum to (or from) this pixel. Any change will be recorded if the objective is improved, otherwise the process will jump to the next pixel.

\textbf{\emph{Spectral analysis}}: The power spectra of sharp natural images exhibit canonical behaviors expressing strong regularities \citep{field1987relations,burton1987color}:
\begin{equation}
|\hat{x}(\bm{\omega})|^2\propto\|\bm{\omega}\|^{-\beta},
\end{equation}
for $\bm{\omega}\neq(0,0)$, where $\hat{x}$ is the Fourier transform of the sharp image $x$, $\bm{\omega}$ is the frequency coordinates. However, such behaviors will be disturbed by the degradation of the blur effect, which might display statistical irregularities. \citet{goldstein2012blur} derived the power spectrum of the blurry image $y$ as
\begin{equation}
|\widehat{y*d}(\bm{\omega})|^2\approx c|\hat{h}(\bm{\omega})|^2,
\end{equation}
where $d$ denotes the spectral whitening operation and $c$ is a constant. Based on this relationship, the power spectrum of the blur kernel $h$ can be recovered from the statistics of the whitened spectrum of $y$ \citep{goldstein2012blur,hu2012psf}. Further, recovering the kernel $h$ given its power spectrum $|\hat{h}|^2$, requires the estimation of the phase component of $\hat{h}(\bm{\omega})$. This mission is generally accomplished by a phase retrieval technique. For example, \citet{goldstein2012blur} proposed a robust version of the relaxed averaged alternating reflections (RAAR), which was originally developed by \citet{luke2005relaxed} for phase retrieval. \citet{hu2012psf} recovered the phase using the error-reduction phase retrieval algorithm \citep{fienup1982phase}.

\section{Practical Issues} \label{practicalissue}

\subsection{Boundary}

In image deblurring tasks, pixels located around the boundary of the blurry image are dependent upon the unknown pixels outside the observed region. Inappropriate processing on these pixels can bring severe artifacts. Typically, several kinds of \emph{boundary condition} (BC) are utilized to formulize the boundary issue. For example, the periodic BC, which assumes a periodic convolution, is frequently used. This condition facilitates the implementation of fast Fourier transform (FFT) and thus speeds up the optimization. Alternatively, two other ways to define the condition are zero BC and reflexive BC. The zero BC pads the external region with zero values, while the reflexive BC indicates that the pixels outside the image are a reflection of those near the boundary but in the image. Even though these BCs make the deblurring tasks addressable and computationally convenient, they are intrinsically an approximate procedure and do not correspond to the real imaging systems. Deconvolution using these BCs could produce staircase artifacts in the deblurred image. To properly handle the boundary issue, \citet{almeida2013deconvolving} and \citet{matakos2013accelerated} proposed the integration of a masking scheme into the image blurring model, which can be formulated as
\begin{equation}\label{smallboundary}
\mathbf{y}_{\mathbf{M}}=\mathbf{MHx}+\mathbf{n},
\end{equation}
where $\mathbf{M}$ is a masking matrix designed to select only the subset of pixels which do not depend on the boundary pixels, and $\mathbf{y}_{\mathbf{M}}=\mathbf{My}$. In this case, the assumed BC for the convolution is then irrelevant to the deconvolution.

Note that the above model implies that the pixels around the boundary but inside the image cannot be estimated. These pixels, however, are calculated in \citet{ji2012robust}'s method and \citet{sorel2012removing}'s method by extending the size of the mask to involve all the pixels of the observed image. This model can be summarized as
\begin{equation}\label{largeboundary}
\mathbf{y}=\mathbf{MHx}+\mathbf{n},
\end{equation}
which biasedly estimates the boundary pixels. In practice, the above models (defined by (\ref{smallboundary}) and (\ref{largeboundary})) can be accordingly employed by considering their specific properties.

\subsection{Noise and Outliers}

Noises in images are usually caused by insufficient exposure. The longer the exposure time is, the lower the noise level will be exhibited. In general conditions, we can therefore expect that the noises in blurry images have not reached a sensitive level, and these noises can be effectively removed by appropriately choosing the parameters of the noise model in the Bayesian inference framework (Section \ref{bayesinferframe}) or the regularization parameters in variational methods (Sections \ref{variationalmethod} and \ref{sparserepmethod}). Nevertheless, noises should be carefully handled in extreme cases, such as in low-light conditions, or when capturing a very fast object. This is because deblurring an image with noticeable noise will produce ringing artifacts in the results. To handle this issue, \citet{tai2012motion} applied an existing denoising algorithm as a preprocessing step, and successively conducted blind deconvolution on the denoised image to estimate the blur kernel and the sharp image. However, a drawback of this method has been noticed by \citet{zhong2013handling}, i.e. the denoising operation can bias the accurate estimation of the kernel. Thus, Zhong et al. designed a set of denoising filters based on the directional filters so that the denoising operation has no effect on the estimated kernel.

Another source of deblurring that should be adequately addressed is the outliers. According to \citet{cho2011handling}'s definition, outliers include all factors which cannot be explained by the linear model (\ref{simpleform}), e.g. saturated/clipped pixels, non-Gaussian noise and nonlinear CRF. If these outliers are processed in the same way as the inliers, the resultant image will show severe ringing artifacts. In \citet{cho2011handling}'s method, the blurry image is classified into two parts, i.e. inliers and outliers. Different statistical assumptions are respectively imposed on these two parts, and the estimation of the sharp image and the classification of the inlier/outlier are alternated until a reasonable result is obtained.

\section{Promising Future Directions} \label{futuredirect}

Most of conventional methods for image deblurring process a single blurry image, in either non-blind or blind ways. The non-blind methods try to suppress the artifacts in the resultant image, while the blind methods are developed to recover an accuracy blur kernel. However, a single image has very limited information for the deblurring task. Although the blur kernel is known in the non-blind case, the high frequency information which is lost in the blurring process, is difficult to be restored from the degraded image. On the other hand, the deblurring problem is underdetermined in the blind case since the parameters to be estimated (i.e., $x$ and $h$) outnumber the observations. In both cases, the performance of deblurring using single image is limited.

Effective options to resolve the above problem include 1) involving multiple images, each of which can provide different information in recovering the sharp target image, and 2) improving the camera systems to capture more detailed information that can be exploited in the recovering process. Regarding these options, two promising directions, \emph{learning-based deblurring} and \emph{hardware modifications}, have emerged in recent years, which are discussed in this section.

Another possible but difficult direction is to develop new models for single image deblurring, especially for blind single image deblurring. The recent progresses in this area include the $\text{MAP}_h$ scheme, homography-based modeling, efficient filter flow modeling (please refer to Sections \ref{bayesinferframe}, \ref{homographymethod}, \ref{regionmethod} respectively). However, the performance of these methods are far from optimal, and potential techniques should be exploited.

\subsection{Learning-based Deblurring} \label{learningdeblurring}

Learning-based deblurring methods explore machine learning techniques to learn how to restore a target image from additional images which can be easily obtained from other resources, e.g. the Internet. Since different images share similar local patterns, the images used for learning can provide both sharp information and high frequency details that can be used to deduce the target image, even though they have different contents. This strategy has also been incorporated in various image processing tasks, such as image denoising and image super-resolution.

The first type of learning-based method is to learn a subspace where the sharp image may be located. By extracting the local patterns from multiple sharp images, a subspace can be constructed, in which the target image and the sharp images share similar details and thus details in the target image can be accurately represented. For example, \citet{joshi2010personal} proposed to construct one person's eigenfaces (which form the subspace) by using this person's multiple sharp facial images. To restore the same person's blurry facial image, these eigenfaces are employed to constrain the corresponding sharp image. Similarly, \citet{dong2011image} and \citet{ni2011example} introduced to build the patch space or the patch manifold by first clustering the patches extracted from a sharp image database, and then applying PCA on each cluster to form its corresponding subspace. The whole patch space is composed of the resultant subspaces, each of which provides specific local structural information, and can therefore constrain the local patterns of the target image. An alternative scheme to specify the subspace is to estimate the patch distribution from sharp image databases, by assuming that the local patterns of different natural images should follow the common patch distribution. A typical example was developed by \citet{zoran2011learning}, in which the GMM was learned to model the prior distribution of sharp patches. Straightforwardly using this prior under the MAP framework, however, could produce a deblurred image in which the restored patches might not follow the distribution of the patches sampled from the original sharp image. Thus, to reduce this problem, the authors proposed a regularization technique that minimizes the difference between the patch distribution of the target image and that of the learned GMM. \citet{sun2013edge} developed a nonparametric method to model the distribution of the edge patches extracted from the BSDS500 dataset \citep{arbelaez2011contour}. In this method, the nonparametric distribution is learned by clustering the extracted patches and then counting the percentage of the patches falling into each cluster.

The second type of learning-based method is to learn a restoring function which recovers the sharp target image from its blurry version. In this case, multiple sharp images together with their blurry correspondences are generally utilized to train the parameters of the restoring function. For example, \citet{schmidt2013discriminative} employed the \emph{regression tree field} (RTF) to model a non-linear regressor that specifies the local deblurring parameters. Before deblurring an image, RTFs are learned by maximizing a peak signal-to-noise ratio-based loss function. Rather than directly learning a deblurring function, \citet{schuler2013machine} estimated the deblurred image by using \emph{direct deconvolution} \citep{hirsch2011fast} in the Fourier domain, and then incorporated the \emph{multilayer perceptron} (MLP) to remove the artifacts of that deblurred image. Note that in both methods, RTF and MLP are trained based on a collection of sharp images and their corresponding synthesized blurry counterparts.

Finally, the learning scheme has been employed to facilitate the blur-related tasks by exploiting additional sharp information. For example, \citet{hu2012good} developed a novel strategy to determine which region of the blurry image is good to deblur, or in other words, is good to estimate the blur kernel. To detect the good regions, \emph{conditional random field} is employed as the learning scheme from which the local structural information can be well exploited. The regions are considered to be good if they could provide as many informative structures as possible. In addition, \citet{couzinie2013learning} presented a novel scheme to handle the non-uniform blur such as defocus blur and linear motion blur. In this scheme, the kernel size is regarded as the class label indicating how large the blur kernel is for each blurry pixel. A multi-label segmentation method is introduced to estimate each pixel's kernel size. Once the kernel sizes for all pixels are obtained, the blurry image is restored within a variational framework.

The sharp information of additional images are beneficial not only for recovering the details of a barget blurry image, such as in the cases of learning a subspace and learning a restoring function, but also for providing assistant information for deblurring tasks, such as in the cases of specifying the good regions and determining the kernel sizes. As demonstrated by the attractive experimental results in these methods, learning-based delburring is a promising direction in this domain.

\subsection{Hardware Modifications}

Image deblurring has recently benefited from the development of imaging systems. A typical instance is that most camera systems have been integrated with the image stabilization modular, reducing the blur caused by slight shakes of the systems during exposure. However, this image stabilization technique has limited performance in some extreme conditions, such as when the captured object is moving fast, or in the low light environments. Fortunately, more advanced techniques have been proposed to facilitate the image deblurring task.

Computational photography, which is an emerging field, inspires many researchers to develop new techniques for deblurring. Coded exposure photography \citep{ding2010analysis,mccloskey2012design} is one kind of technique in computational photography. In conventional exposure technique, the camera records the photons in a successive duration. In this technique, however, the camera's shutter is fluttered open and closed during the chosen exposure time, while the open-close period is determined by a binary pseudo-random shutter sequence. Its superiority is that the coded exposure photography can produce a blurry image preserving more high-frequency spatial details than the conventional technique, especially when the image contains large motions, textured backgrounds, or partially occluded regions. Considering this, the preserved details can make the deconvolution to be a well-posed problem. Additionally, the shutter sequence can be designed as velocity-dependent when capturing very fast moving object, such that the details of the moving object can be retained \citep{mccloskey2010velocity}. Another kind of technique is the coded aperture photography. The aperture allows the light to go through it, and at the same time, can block the light with different wavelengths by reshaping itself. The rationale behind this technique is that the light through different shapes of apertures provides different information, because the light with different wavelengths has different optical characteristics. Using this property, \citet{zhou2011coded} derived a criterion for selecting a pair of coded apertures, by which the light through the two apertures can preserve the complementary information covering a broad band of scene frequencies. The third kind of computational photography is the coded flash technique, which is specifically designed for low light conditions \citep{mccloskey2011temporally}. In traditional lighting technique, the flash is generally activated in a successive duration. In contrast, by coding the flash timing sequence, this technique can control the illuminance and the exposure time in photographing, which has a similar effect to the coded exposure. When capturing a moving object in low light environment, coded flash system can produce a well-exposed image, and meanwhile can make the PSF estimation easier.

Besides the computational photography, constructing a multi-camera system can provide multiple images of the same scene, each of which carries different information that benefits deblurring tasks. For example, \citet{cho2010motion} developed an orthogonal parabolic camera to capture two images of the scene by moving the sensor with parabolic velocities and in two orthogonal directions. \citet{tai2010correction} designed a hybrid camera system combined of a high-resolution, low-frame-rate camera and a low-resolution, high-frame-rate camera. This system captures a high-resolution but blurry image, as well as multiple low-resolution but sharp images. A beam splitter is used to ensure that the alignment of different images is well solved. By exploiting the information in the sharp images, the blur in the high-resolution image can be removed. \citet{li2011exploring} proposed a hybrid system to capture two well-aligned images with a specific relationship (for example, one image is a $90^{\circ}$-rotated version of the other). This relationship facilitates the mathematical derivation of the sharp image from two blurry images, making the deblurring task more effective.

For deblurring tasks, researchers have also added additional components into the camera systems. For instance, instead of using the conventional color image sensors which only have red (R), green (G), blue (B) patterns, \citet{wang2012high} employed a new color image sensor which adds the panchromatic (pan) pixels to the original RGB pixels. Due to the high light sensitivity, these pan pixels can help to restore the images captured in low light conditions. \citet{bando2013near} employed the focus sweep technique during exposure. Since the scene has varying depths or in-plane motion, the captured image may exhibit a spatially variant blur effect. By using Bando et al.'s method, the blur effect of the image can be turned to a near-spatially invariant scenario. \citet{joshi2010image} integrated a gyroscope and an accelerometer to estimate the camera's acceleration and angular velocity, from which the translation and rotation of the camera can be calculated and used for deblurring.

Hardware modification is always an effective manner to remove the blurry effects in captured images. Benefitting from the advanced hardware techniques, more information of the scenes can be easily recorded, and thus facilitate the deblurring task. Although some achievements have been obtained, various techniques are waiting to be invented in the future.

\section{Performance Evaluation} \label{evaluation}

To complement the discussions of the previous sections, we provide experimental evaluations for representative techniques. The evaluation for image deblurring can be subjective quality assessment or objective quality assessment. Subjective quality assessment predicts the observers' perception without a well-defined numerical quantification. Although the subjective image quality assessment is the most direct and most accurate metric to reflect a person's perception, it is too subject to cater for different persons. In contrast, objective quality assessment metrics can operate in an automatic and numerical manner. The metrics used in this section include \emph{peak signal-to-noise ratio} (PSNR) and \emph{structural similarity} (SSIM) , which are widely utilized to evaluate the performance of image deblurring algorithms. Assuming the restored image $x$ and the un-degraded image $\tilde{x}$ are in the range of $[0,255]$, PSNR and SSIM are defined as
\begin{eqnarray}
\text{PSNR}&=&10\cdot\log_{10}\left(\frac{N\times 255\times255}{\sum_i(x_i-\tilde{x}_i)^2}\right),\\
\text{SSIM}&=&\frac{(2\mu_x\mu_{\tilde{x}}+c_1)(2\sigma_{x\tilde{x}}+c_2)}{(\mu_x^2+\mu_{\tilde{x}}^2+c_1)(\sigma_x^2+\sigma_{\tilde{x}}^2+c_2)},
\end{eqnarray}
where $N$ is the total number of pixels, $\mu_x$ and $\mu_{\tilde{x}}$ are the means of $x$ and $\tilde{x}$, $\sigma_x^2$ and $\sigma_{\tilde{x}}^2$ are the variances of $x$ and $\tilde{x}$, $c_1$ and $c_2$ are two variables to stabilize the division with weak denominator.

According to different task settings, existing methods can be categorized as non-blind uniform deblurring, blind uniform deblurring and non-uniform deblurring. Non-uniform deblurring is always blind since the kernel is hard to obtain in real applications. For each kind of setting, we evaluate the representative methods where the codes are publicly accessible.

\subsection{Non-blind Uniform Deblurring}

The test images used for non-blind uniform deblurring come from \citep{levin2009understanding,levin2011understanding}, including 4 images and 8 blur kernels (see Fig. 9). To generate blurry images, the 4 sharp images are corrupted by each blur kernels and additive white Gaussian noise with standard deviation $\sigma=0.01$, resulting in 32 degraded images. The methods are evaluated on these synthetic blurry images, including \citet{schuler2013machine}, \citet{zoran2011learning}, \citet{schmidt2011bayesian}, \citet{danielyan2012bm3d}, \citet{dong2013nonlocally}, \citet{cho2011handling}, \citet{afonso2010fast}, \citet{almeida2013deconvolving}, and \citet{hullin2013stochastic}.

\begin{figure}[t]
\centering
\subfloat{\includegraphics[width=1.45in]{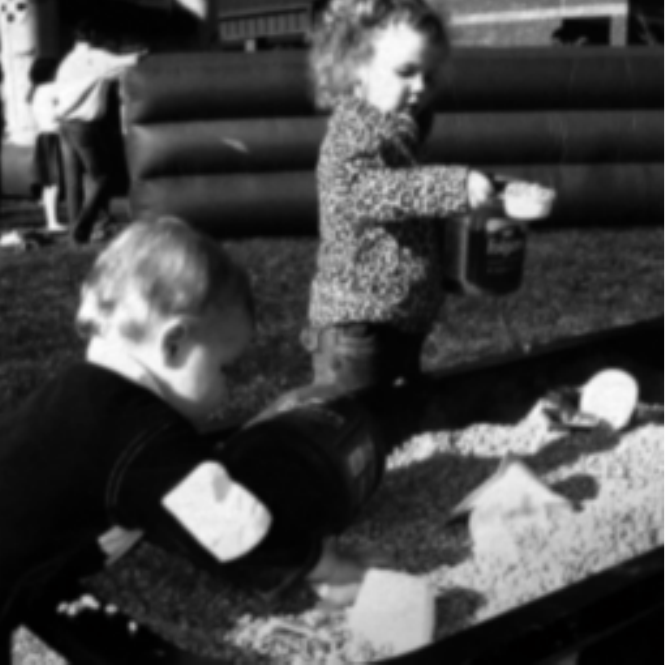}}\hspace{0.01cm}
\subfloat{\includegraphics[width=1.45in]{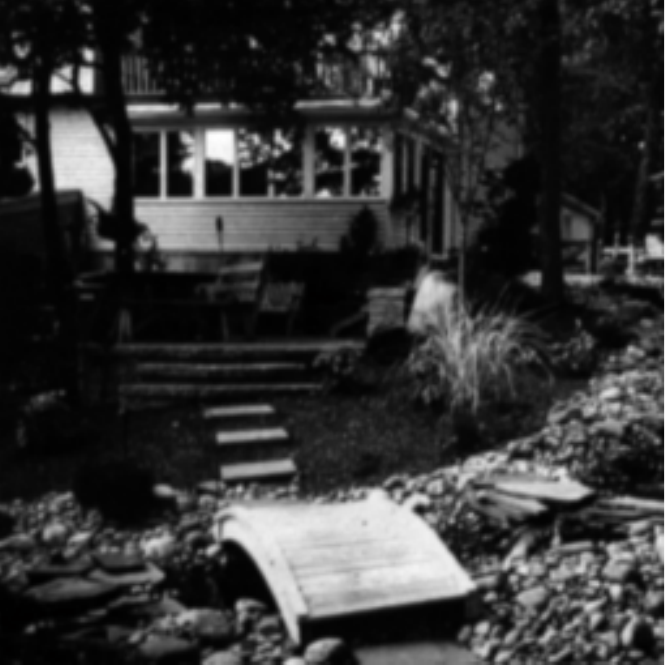}}\hspace{0.01cm}
\subfloat{\includegraphics[width=1.45in]{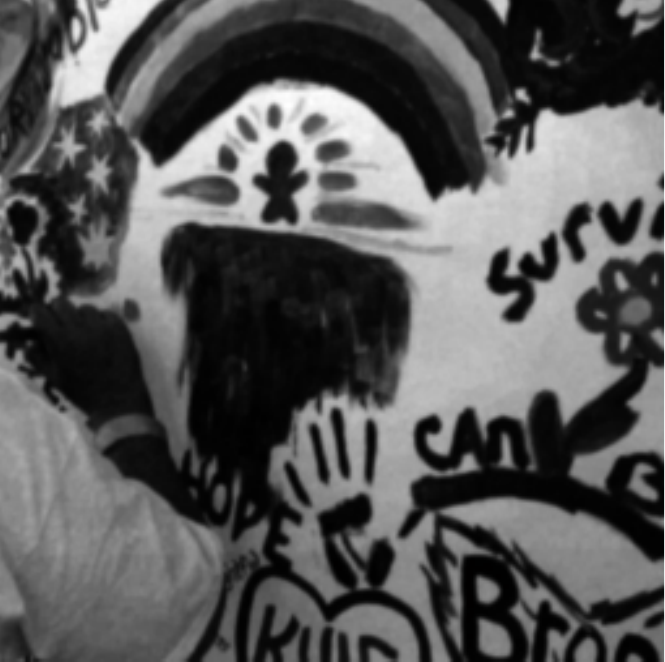}}\hspace{0.01cm}
\subfloat{\includegraphics[width=1.45in]{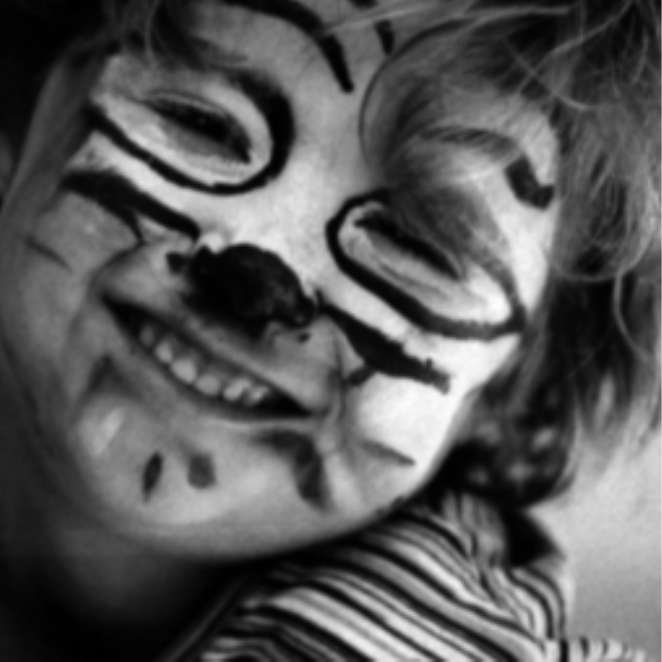}}\hspace{0.01cm}\\
\subfloat{\includegraphics[width=0.7in]{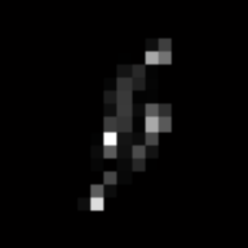}}\hspace{0.005cm}
\subfloat{\includegraphics[width=0.7in]{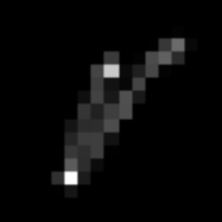}}\hspace{0.005cm}
\subfloat{\includegraphics[width=0.7in]{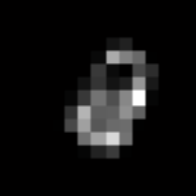}}\hspace{0.005cm}
\subfloat{\includegraphics[width=0.7in]{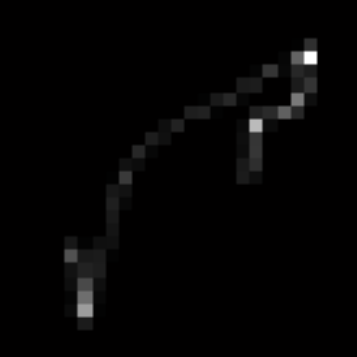}}\hspace{0.005cm}
\subfloat{\includegraphics[width=0.7in]{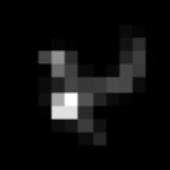}}\hspace{0.005cm}
\subfloat{\includegraphics[width=0.7in]{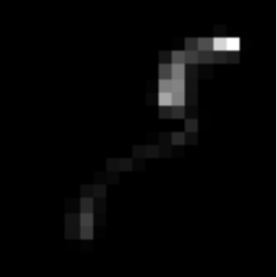}}\hspace{0.005cm}
\subfloat{\includegraphics[width=0.7in]{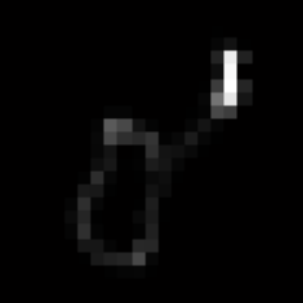}}\hspace{0.005cm}
\subfloat{\includegraphics[width=0.7in]{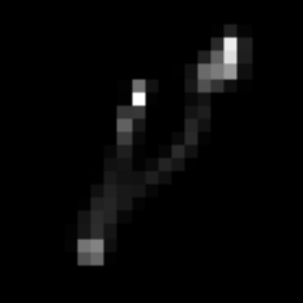}}\hspace{0.005cm}
\caption{First row: 4 test images. Second row: 8 blur kernels.}
\label{testdata}
\end{figure}

\begin{figure}[t]
\begin{minipage}[t]{0.49\linewidth}
\centering
\includegraphics[width=3.2in]{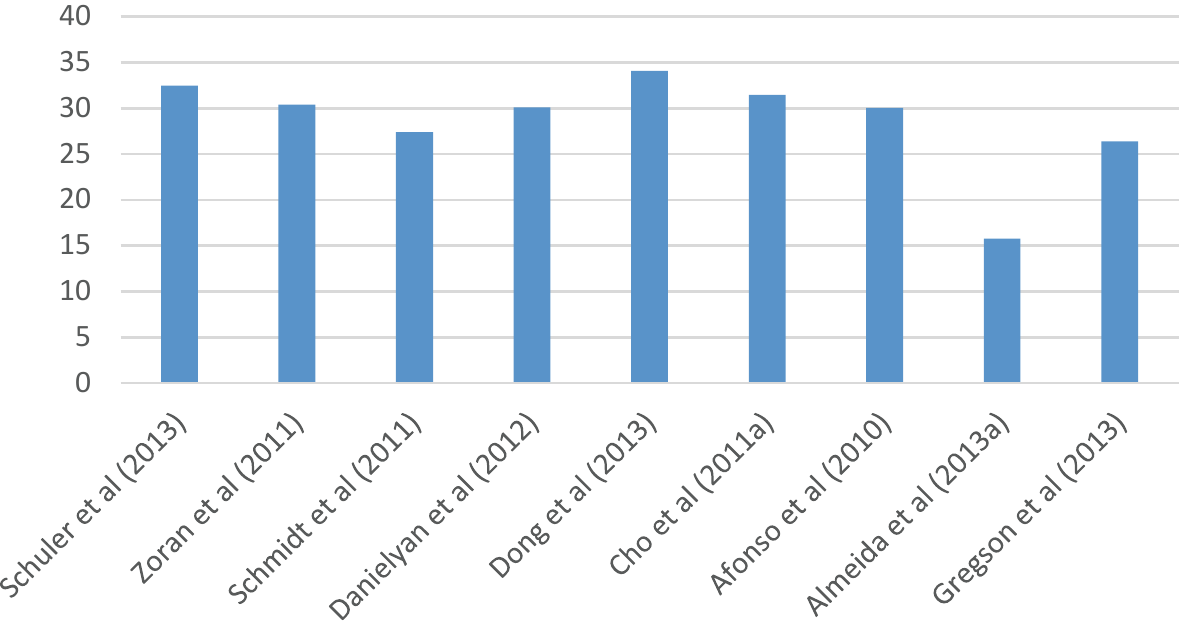}
\caption{Mean PSNR (dB) results of non-blind uniform deblurring.}
\label{nonblindpsnr}
\end{minipage}
\hspace{0.02\linewidth}
\begin{minipage}[t]{0.49\linewidth}
\centering
\includegraphics[width=3.2in]{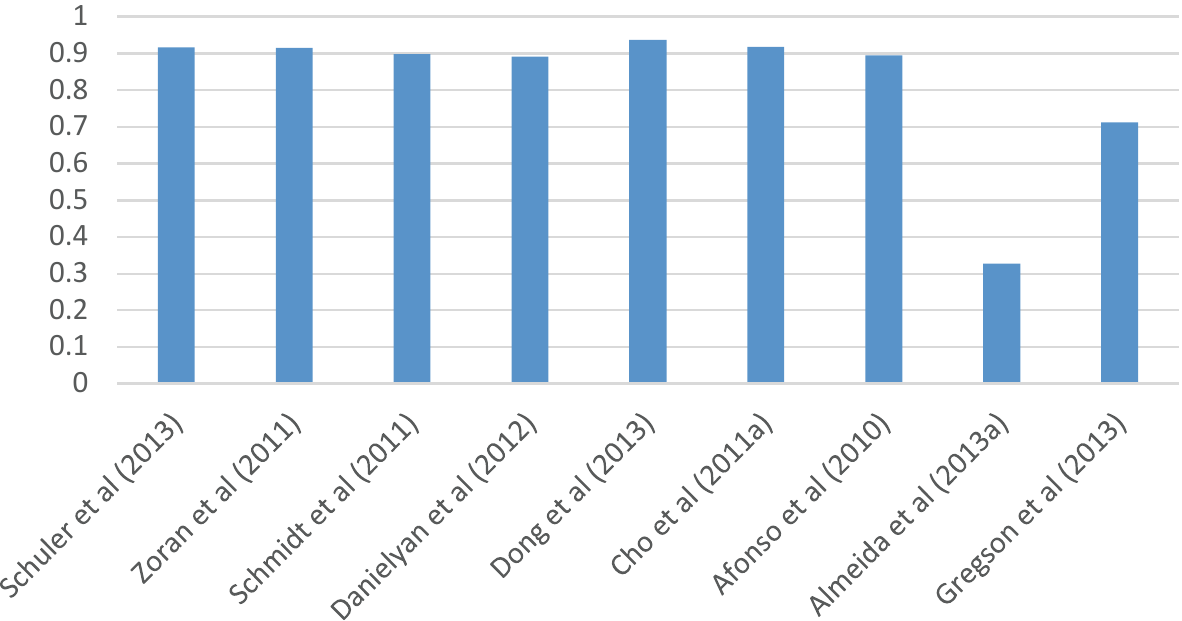}
\caption{Mean SSIM results of non-blind uniform deblurring.}
\label{nonblindssim}
\end{minipage}
\end{figure}

\begin{figure}[t]
\centering
\subfloat[Blurry image]{\includegraphics[width=1.45in]{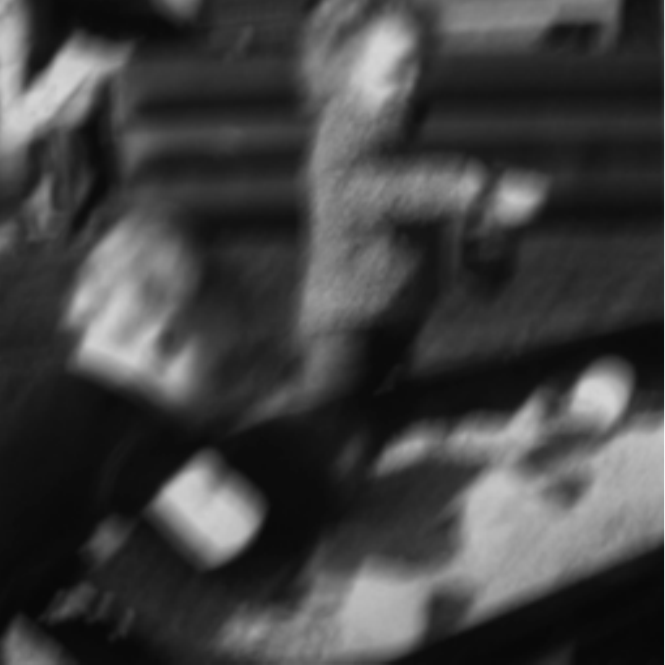}}\hspace{0.01cm}
\subfloat[\citet{schuler2013machine}\protect\\(31.85dB)]{\includegraphics[width=1.45in]{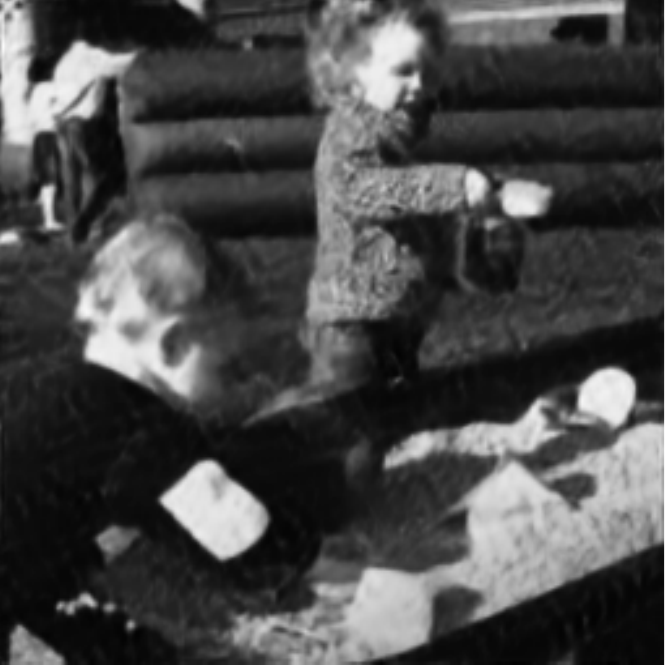}}\hspace{0.01cm}
\subfloat[\citet{zoran2011learning}\protect\\(31.90dB)]{\includegraphics[width=1.45in]{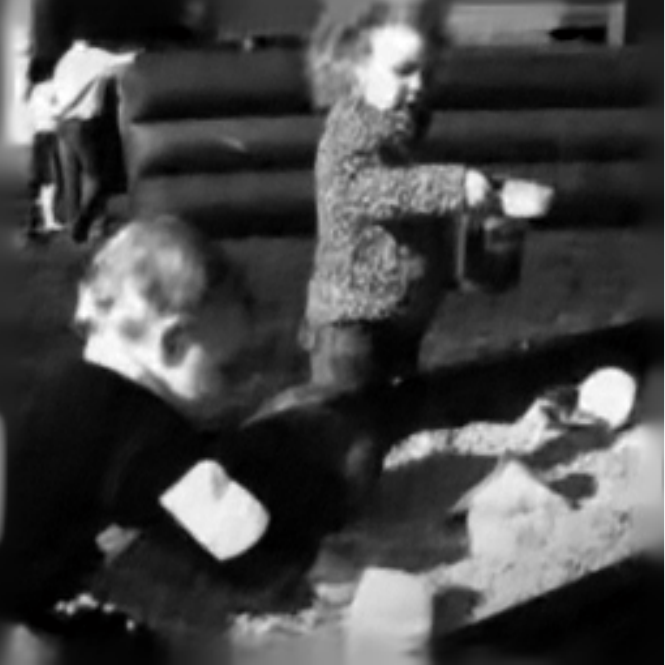}}\hspace{0.01cm}
\subfloat[\citet{schmidt2011bayesian}\protect\\(30.29dB)]{\includegraphics[width=1.45in]{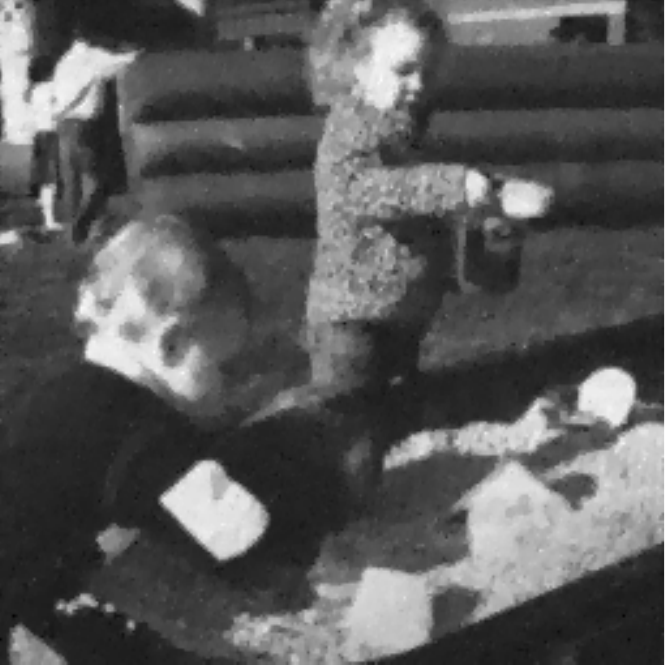}}\\
\subfloat[\citet{danielyan2012bm3d}\protect\\(31.68dB)]{\includegraphics[width=1.45in]{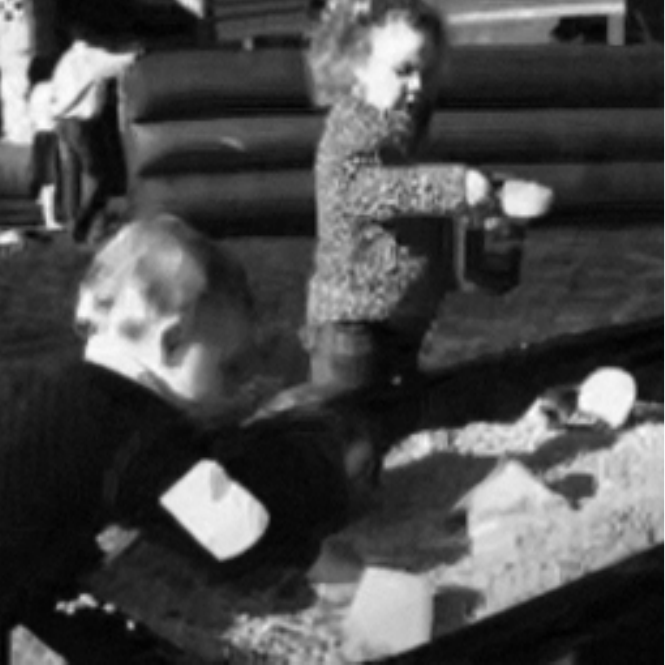}}\hspace{0.01cm}
\subfloat[\citet{dong2013nonlocally}\protect\\(33.50dB)]{\includegraphics[width=1.45in]{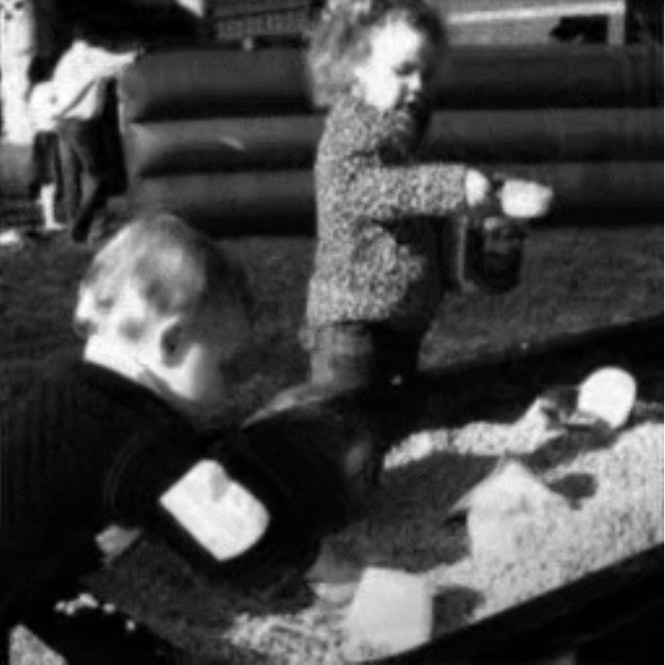}}\hspace{0.01cm}
\subfloat[\citet{cho2011handling}\protect\\(31.49dB)]{\includegraphics[width=1.45in]{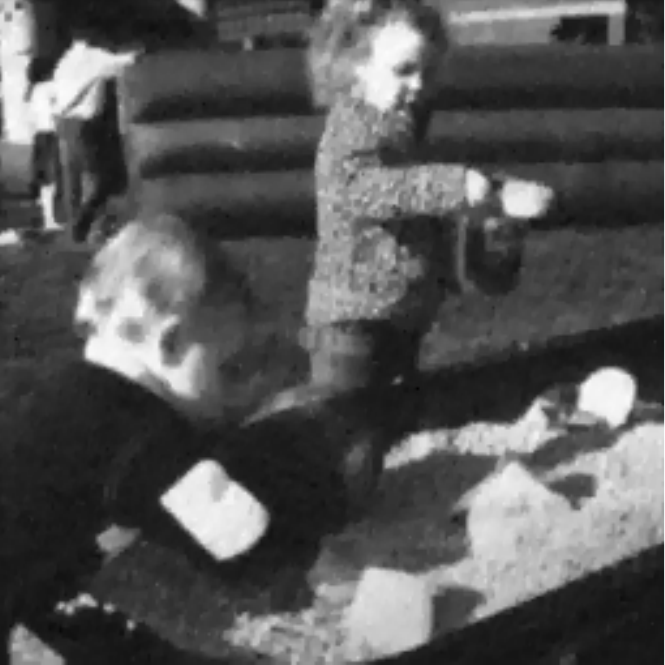}}\hspace{0.01cm}
\subfloat[\citet{afonso2010fast}\protect\\(29.40dB)]{\includegraphics[width=1.45in]{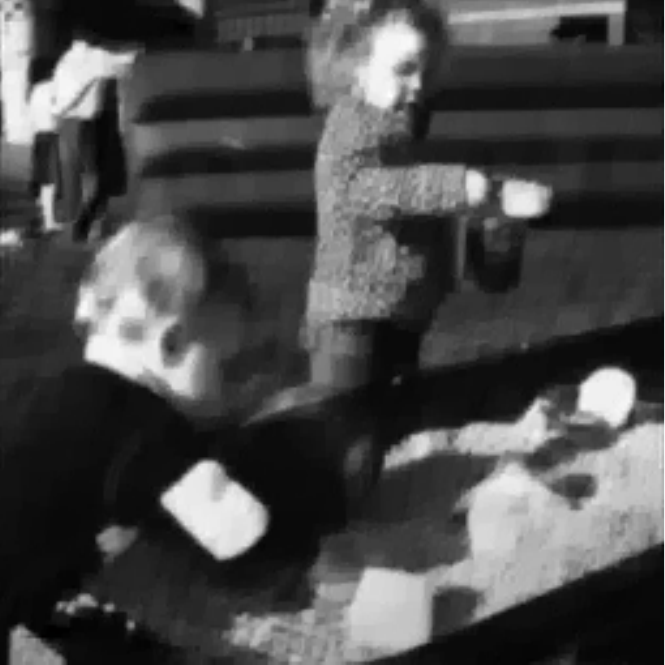}}\\
\subfloat[\citet{almeida2013deconvolving} (17.15dB)]{\includegraphics[width=1.45in]{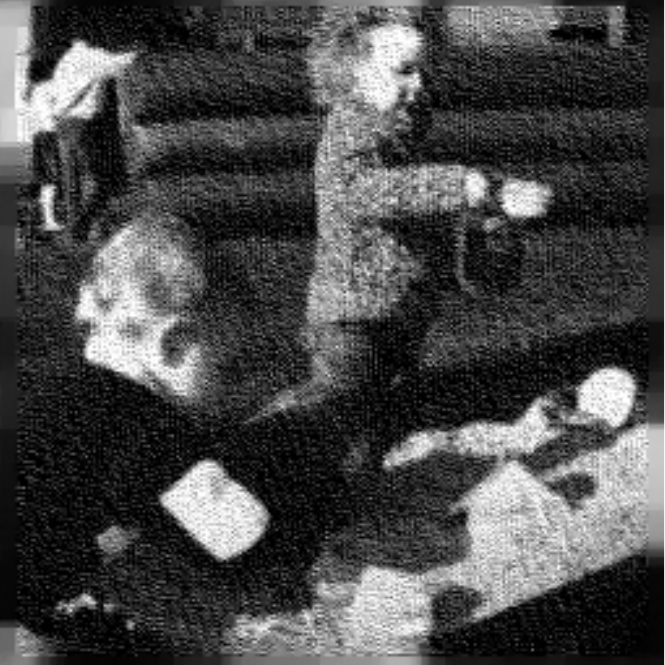}\label{almeidares}}\hspace{0.01cm}
\subfloat[\citet{hullin2013stochastic}\protect\\(27.04dB)]{\includegraphics[width=1.45in]{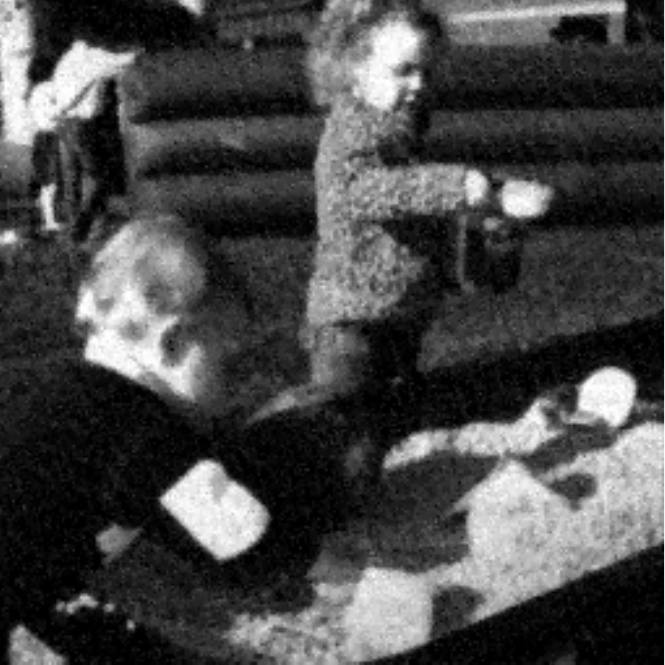}}
\caption{An illustrative example of non-blind uniform deblurring results by different methods. The blurry image is synthesized with the first image and the first kernel in Fig. \ref{testdata}.}
\label{nonblindexample}
\end{figure}

The objective assessment on the deblurred images obtained by these methods are shown in Fig. \ref{nonblindpsnr} and Fig. \ref{nonblindssim}, while the visual comparisons of the deblurring results on an exemplary image are shown in Fig. \ref{nonblindexample}. In summary, \citet{schuler2013machine}, \citet{zoran2011learning} and \citet{schmidt2011bayesian} employ learning schemes to constrain the restored images by exploring additional sharp information, and the resultant images exhibit promising recoveries. Compared with these three methods, the best performance acquired by \citet{dong2013nonlocally} suggests that the nonlocal similar patterns discovered in the blurry image are more helpful for restoration than the information from other irrelevant sharp images. This is mainly because the nonlocal patterns in a blurry image are more stable and consistent than those in other images, and therefore have good expressivity. The results of both \citet{cho2011handling} and \citet{schmidt2011bayesian} recommend that an appropriate processing of outliers and noises is important, otherwise like \citet{almeida2013deconvolving}, the noise in the restored image (Fig. \ref{nonblindexample}(i)) might be severe.

\subsection{Blind Uniform Deblurring}

We consider 10 evaluated methods for blind uniform deblurring, including \citet{zhang2013multi}, \citet{sun2013edge}, \citet{xu2010two}, \citet{levin2011efficient}, \citet{babacan2012bayesian}, \citet{krishnan2011blind}, \citet{cai2012framelet}, \citet{cho2011blur}, \citet{goldstein2012blur}, and \citet{zhong2013handling}. The algorithms are applied on the same set of test images as shown in Fig. \ref{testdata}.

\begin{figure}[t]
\begin{minipage}[t]{0.49\linewidth}
\centering
\includegraphics[width=3.2in]{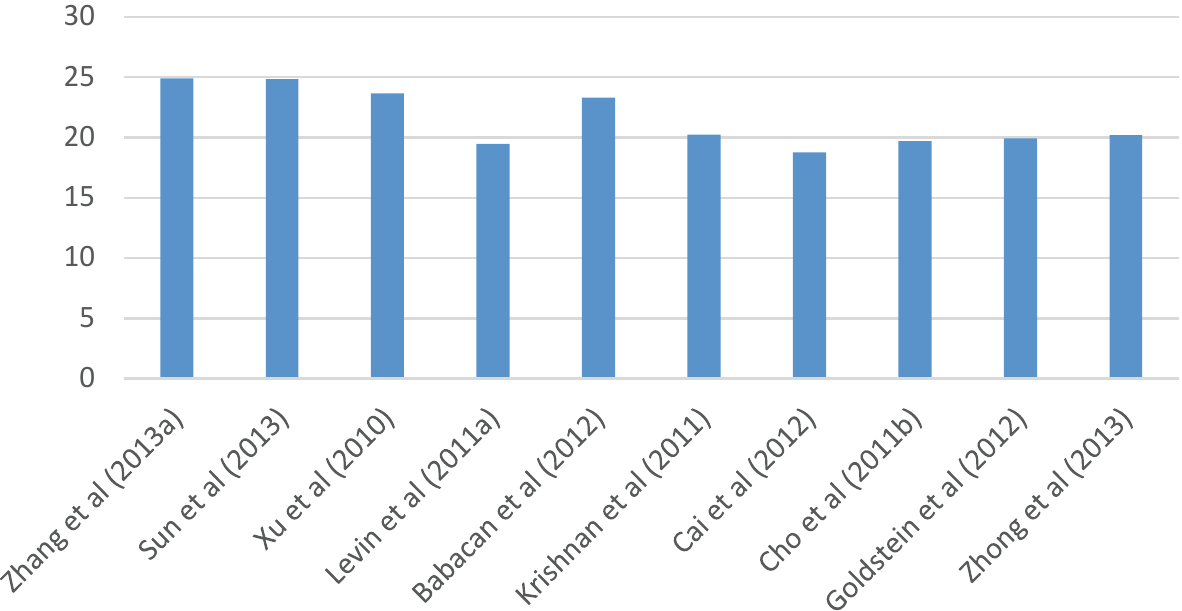}
\caption{Mean PSNR (dB) results of blind uniform deblurring.}
\label{blindpsnr}
\end{minipage}
\hspace{0.02\linewidth}
\begin{minipage}[t]{0.49\linewidth}
\centering
\includegraphics[width=3.2in]{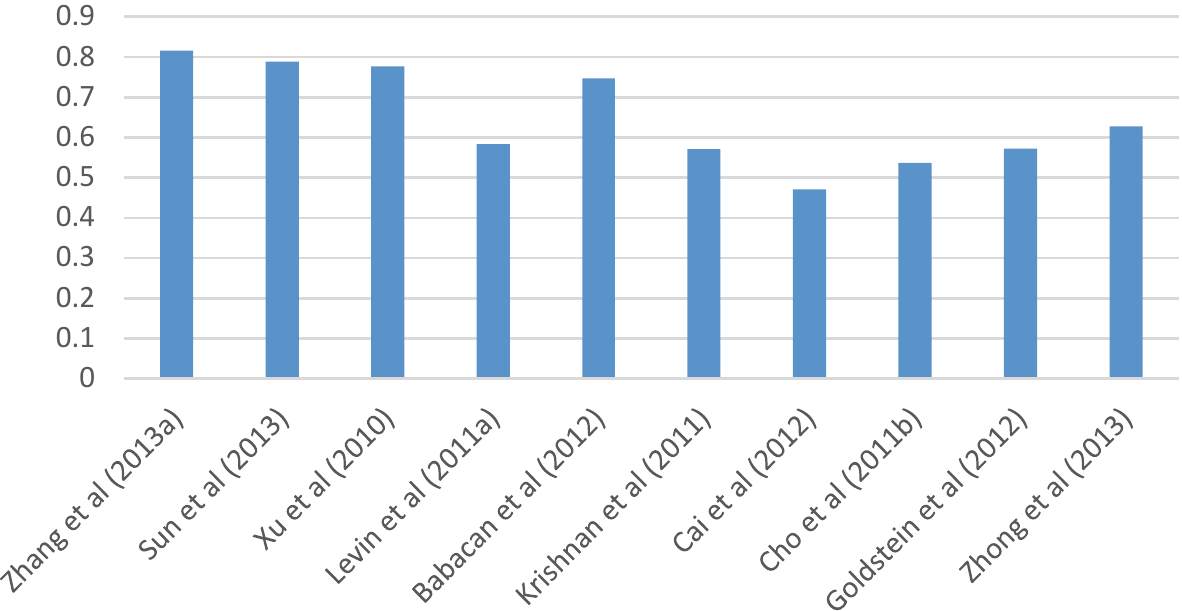}
\caption{Mean SSIM results of blind uniform deblurring.}
\label{blindssim}
\end{minipage}
\end{figure}

\begin{figure}[t]
\centering
\subfloat[Blurry image]{\includegraphics[width=1.45in]{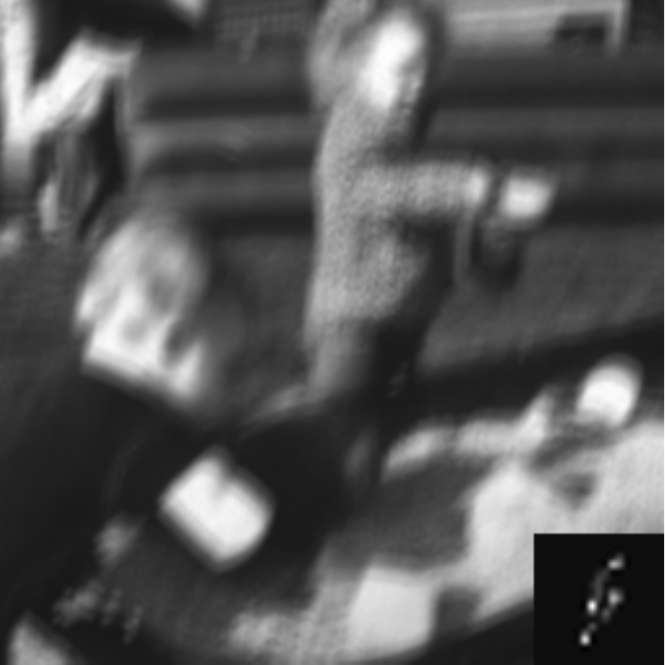}}\hspace{0.01cm}
\subfloat[\citet{zhang2013multi}\protect\\(26.35dB)]{\includegraphics[width=1.45in]{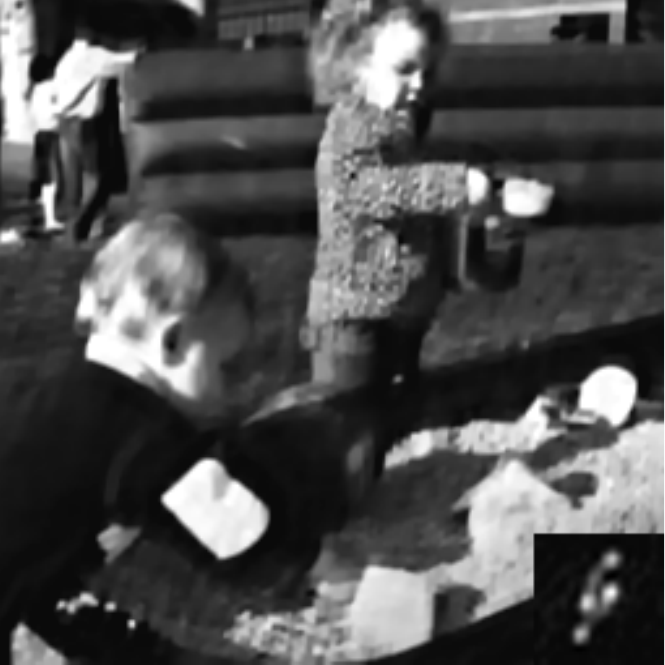}}\hspace{0.01cm}
\subfloat[\citet{sun2013edge}\protect\\(28.40dB)]{\includegraphics[width=1.45in]{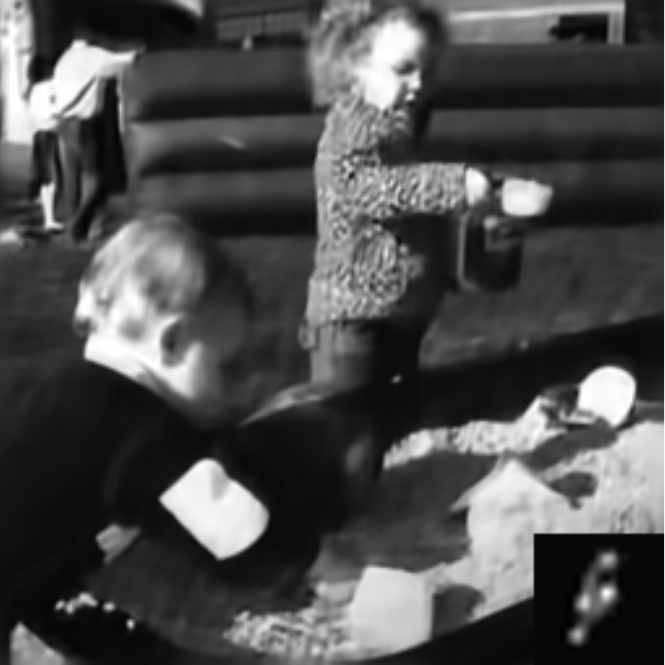}}\hspace{0.01cm}
\subfloat[\citet{xu2010two}\protect\\(26.09dB)]{\includegraphics[width=1.45in]{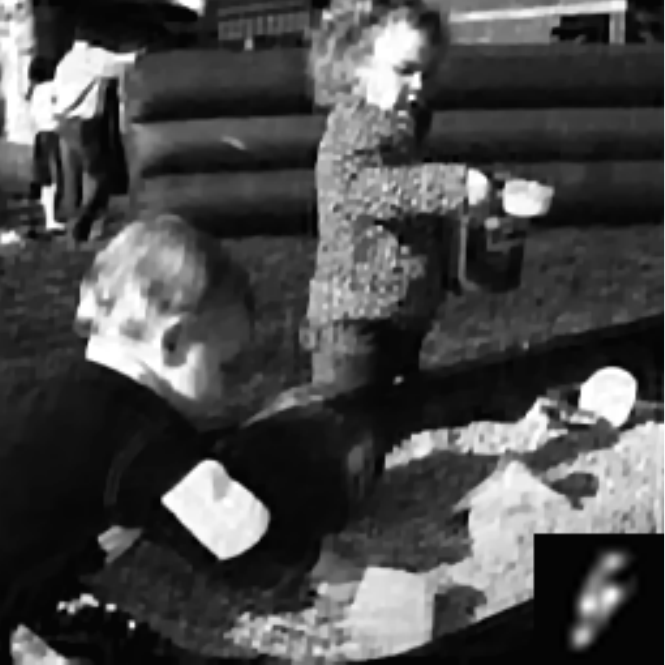}}\\
\subfloat[\citet{levin2011efficient}\protect\\(29.16dB)]{\includegraphics[width=1.45in]{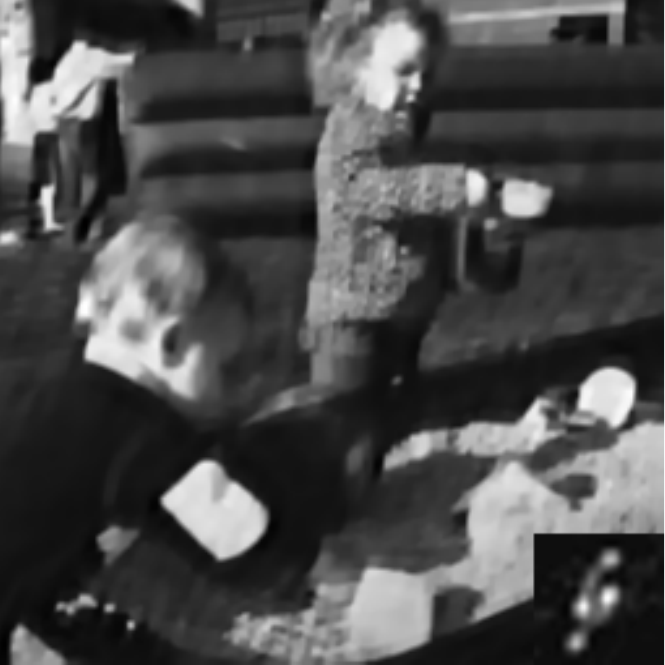}}\hspace{0.01cm}
\subfloat[\citet{babacan2012bayesian}\protect\\(28.39dB)]{\includegraphics[width=1.45in]{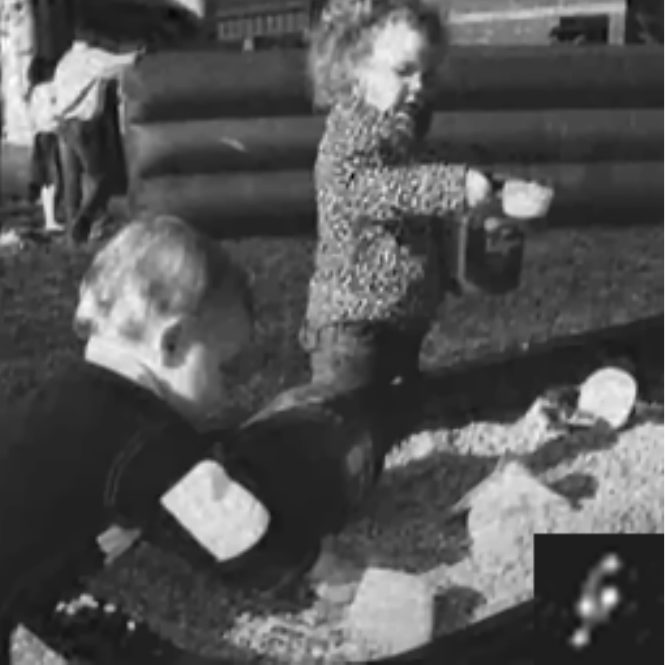}}\hspace{0.01cm}
\subfloat[\citet{krishnan2011blind}\protect\\(22.71dB)]{\includegraphics[width=1.45in]{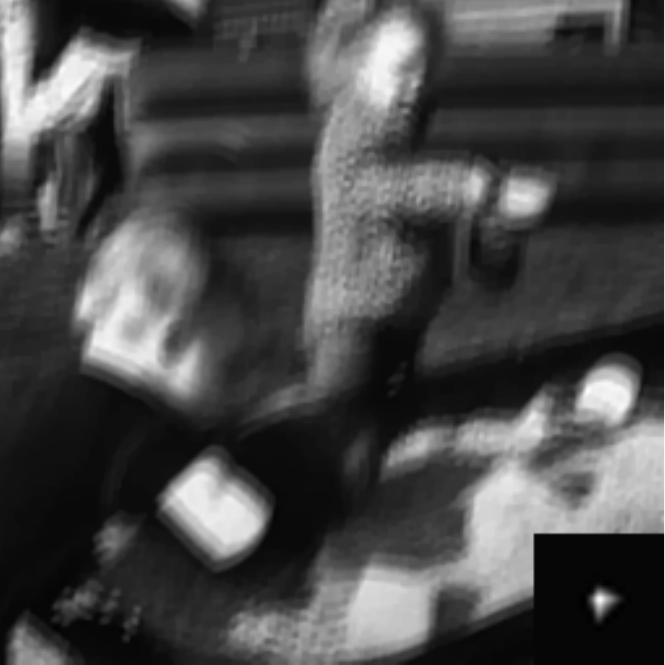}}\hspace{0.01cm}
\subfloat[\citet{cai2012framelet}\protect\\(22.12dB)]{\includegraphics[width=1.45in]{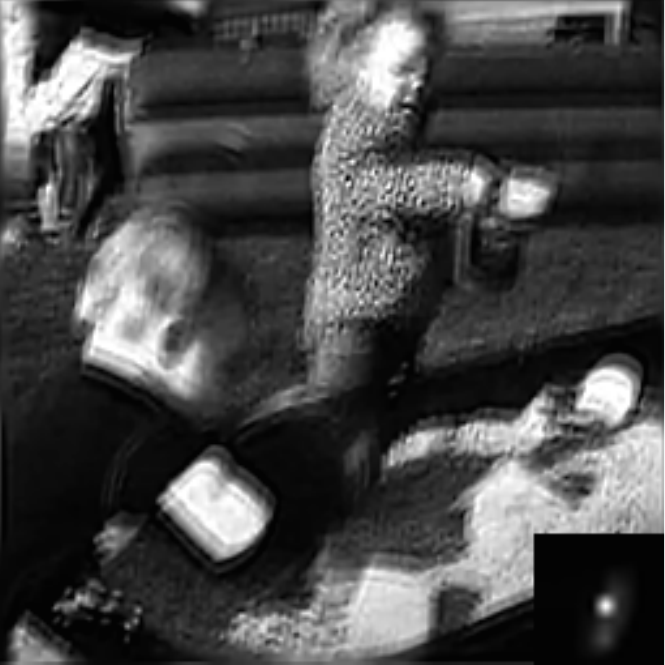}}\\
\subfloat[\citet{cho2011blur}\protect\\(22.84dB)]{\includegraphics[width=1.45in]{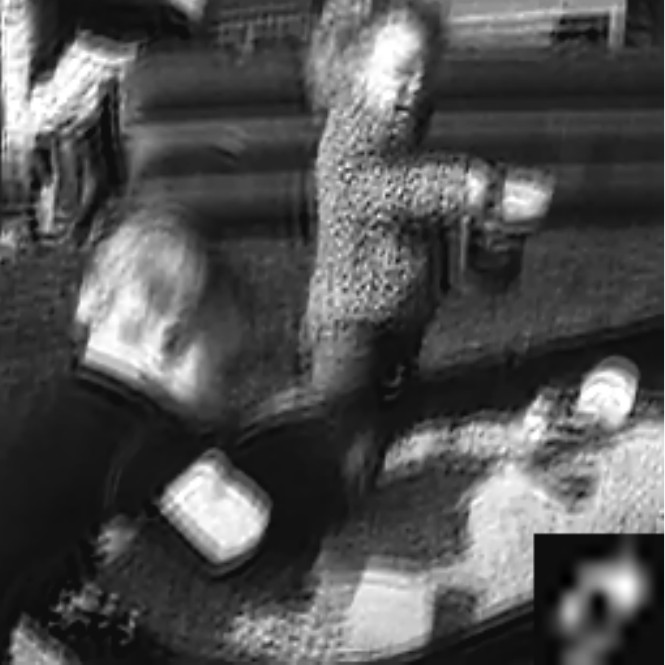}}\hspace{0.01cm}
\subfloat[\citet{goldstein2012blur}(21.47dB)]{\includegraphics[width=1.45in]{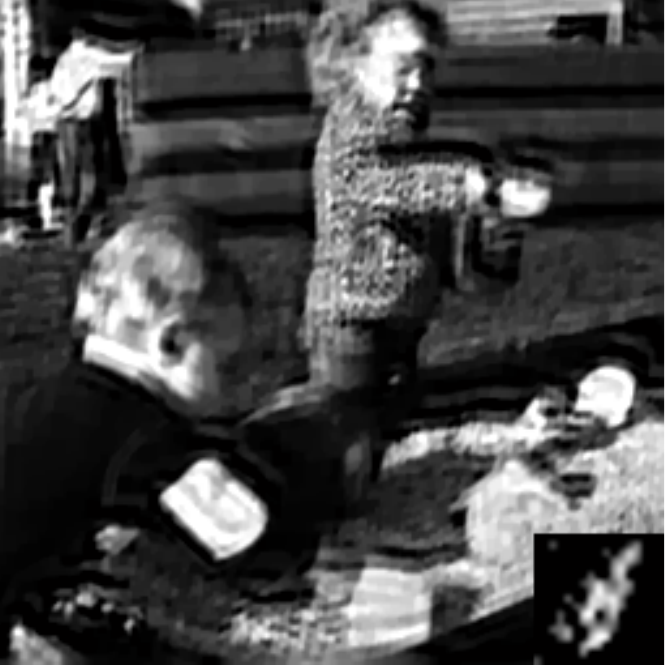}}\hspace{0.01cm}
\subfloat[\citet{zhong2013handling}\protect\\(24.90dB)]{\includegraphics[width=1.45in]{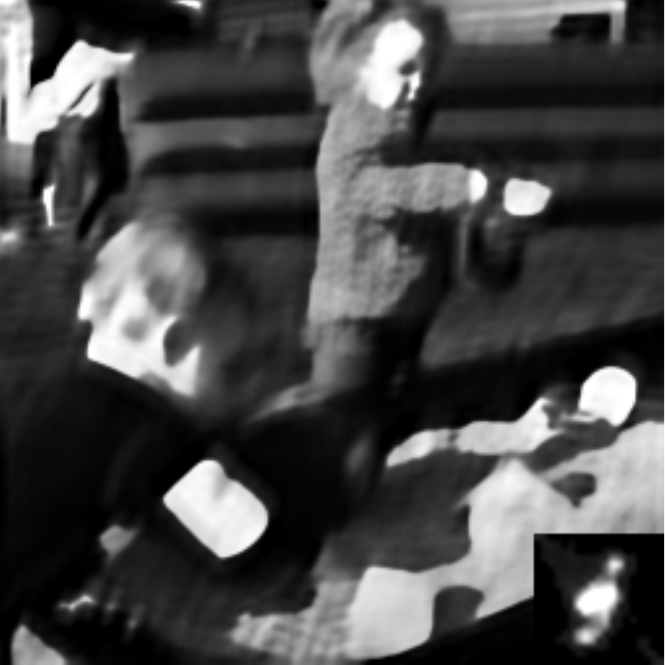}}
\caption{An illustrative example of blind uniform deblurring results by different methods. The blurry image is synthesized with the first image and the first kernel in Fig. \ref{testdata}.}
\label{blindexample}
\end{figure}

Fig. \ref{blindpsnr} and Fig. \ref{blindssim} illustrate the performance of the evaluated methods, while the difference in visual quality between the methods can be inspected in the examples shown in Fig. \ref{blindexample}. Through the joint restoration of multiple blurry images, \citet{zhang2013multi}'s method acquires the best averaged results among the methods. \citet{sun2013edge} obtained comparable performance by using a learning strategy to estimate the patch prior. These observations are consistent with the discussions in Section \ref{futuredirect}. The superiority of \citep{xu2010two} reveals the importance of detecting useful edges in kernel estimation phase. Both \citet{levin2011efficient} and \citet{babacan2012bayesian} incorporated the $\text{MAP}_h$ strategy to estimate the blur kernel and then conducted a non-blind deconvolution, whose effectiveness can be justified as in the figures, even though the averaged performance of \citep{levin2011efficient} is limited. The difference between these two methods indicates that a general image prior (i.e. Gaussian scale model) is crucial in the MAP framework. \citet{krishnan2011blind}'s method, which uses a normalized sparsity measure, is expected to behave promisingly; but in the experiments the performance is sensitive to parameter settings. As in the non-blind deblurring tasks, the performance obtained by \citet{zhong2013handling}'s method also suggests that the noise handling is important in blind deblurring task. In kernel estimation, neither the spectral analysis \citep{goldstein2012blur} nor relatively simple methods such as the framelet \citep{cai2012framelet} and Radon transform \citep{cho2011blur} appear to be competitive.

\subsection{Non-uniform Deblurring}

Non-uniform blur simulates the effect in real captured images, but is hard to synthesize. As a result, \citet{kohler2012recording} presented a benchmark dataset for simulating the camera shakes, which cause non-uniform blur. However, existing methods cannot perform well on this dataset, and are generally sensitive to parameter settings. To provide illustrative examples, we employ 3 real blurry images for evaluation, where the blur is mainly caused by camera shake, as shown in Fig. 16. Non-uniform deblurring methods considered for evaluation here include \citet{xu2013unnatural}, \citet{whyte2010non,whyte2012non}, and \citet{hu2012fast}. Since the corresponding sharp images are inaccessible, we perform the visual comparison in Fig. 17. From the resultant images, we can see that non-uniform deblurring is a difficult task because of the limited available information. The models used in these methods can only simulate specific motion blur, and therefore have strong restrictions. This property could help them perform well in some cases (e.g. the left-column images and the middle-column images in Fig. 17 which become sharper than the original), whereas reduce their effectiveness in other cases (e.g. the right-column images in Fig. 17 which are still blurry).

\begin{figure*}
\centering
\subfloat{\includegraphics[width=2.0in]{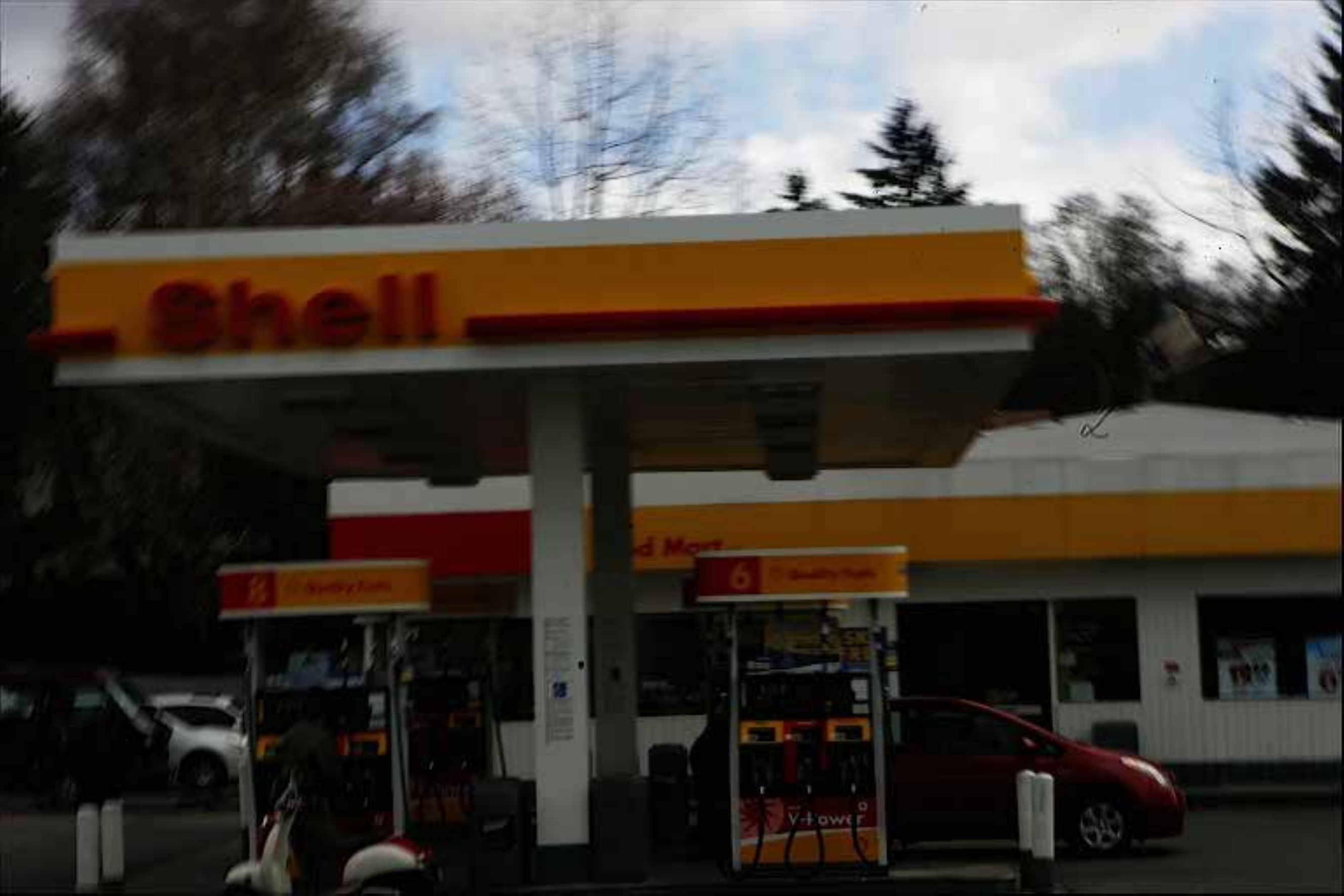}}\hspace{0.01cm}
\subfloat{\includegraphics[width=2.0in]{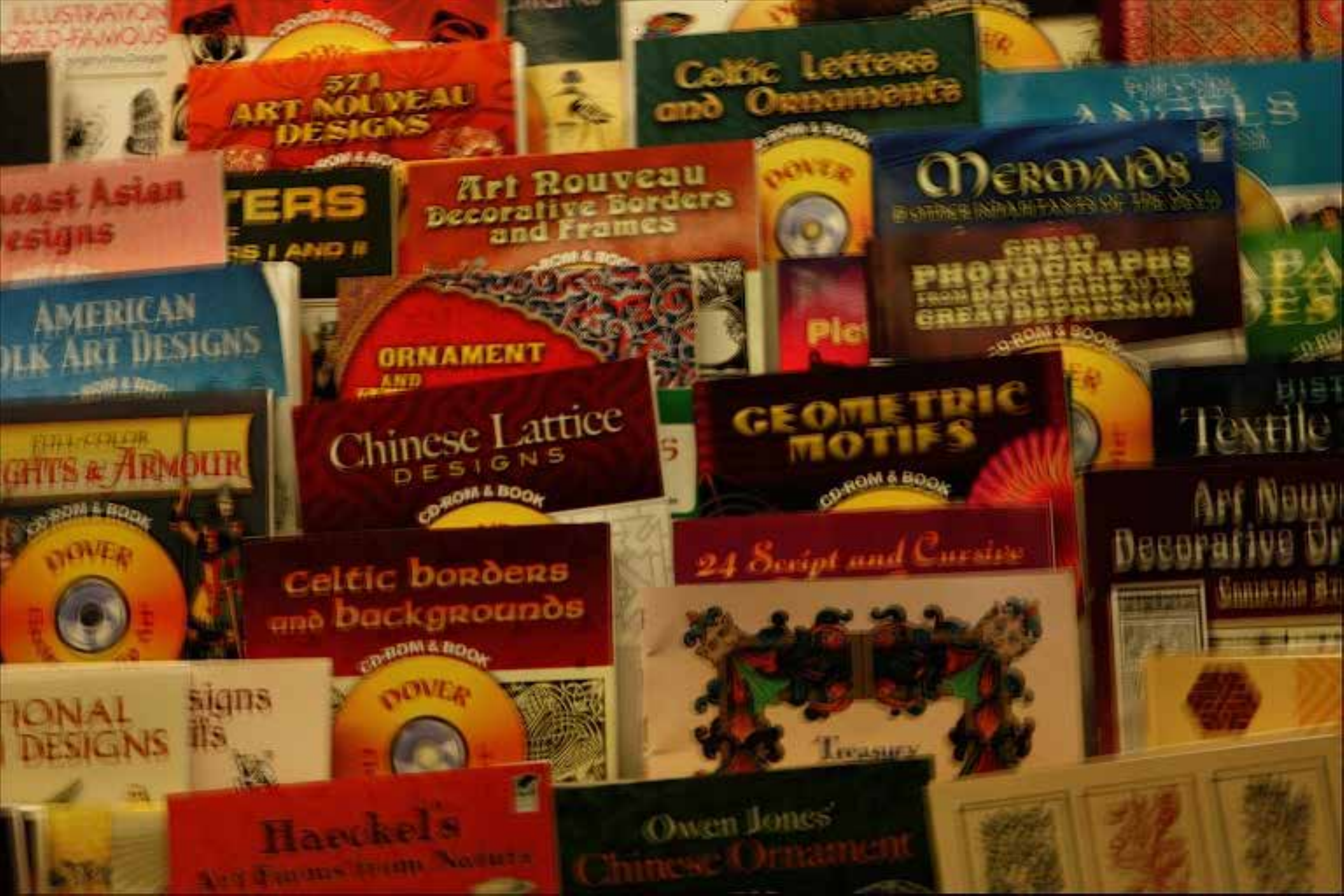}}\hspace{0.01cm}
\subfloat{\includegraphics[width=2.0in]{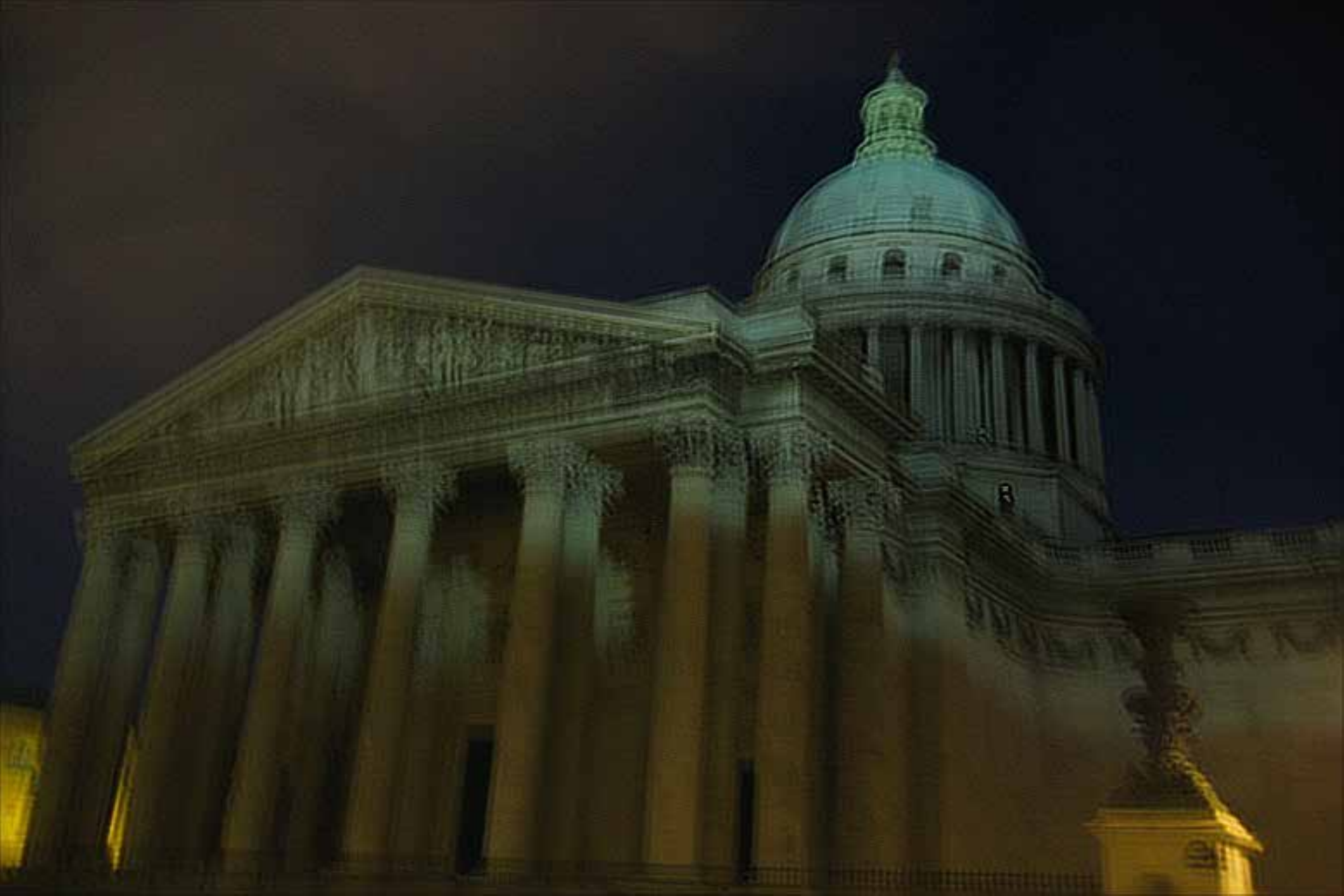}}\hspace{0.01cm}
\caption{Real captured blurry images.}
\end{figure*}

\begin{figure*}
\centering
\subfloat{\includegraphics[width=2.0in]{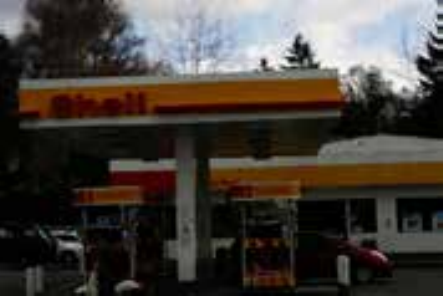}}\hspace{0.02cm}
\subfloat{\includegraphics[width=2.0in]{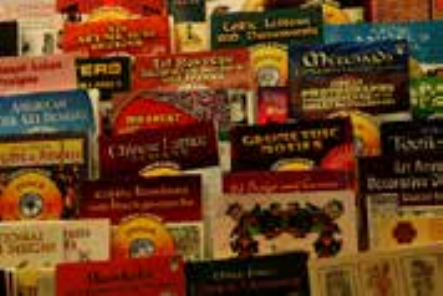}}\hspace{0.02cm}
\subfloat{\includegraphics[width=2.0in]{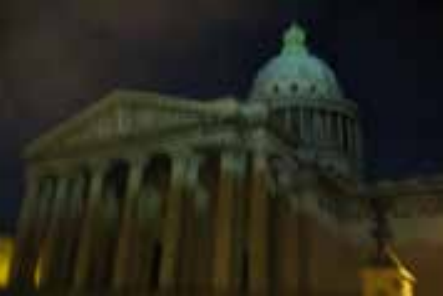}}\\
\subfloat{\includegraphics[width=2.0in]{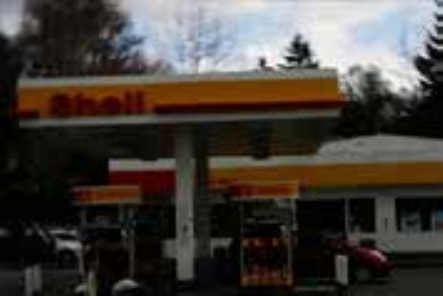}}\hspace{0.02cm}
\subfloat{\includegraphics[width=2.0in]{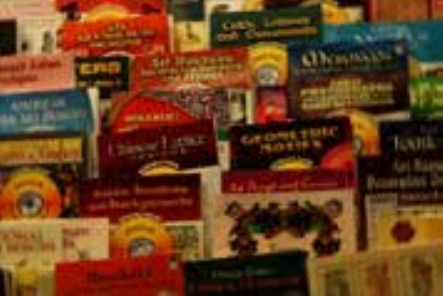}}\hspace{0.02cm}
\subfloat{\includegraphics[width=2.0in]{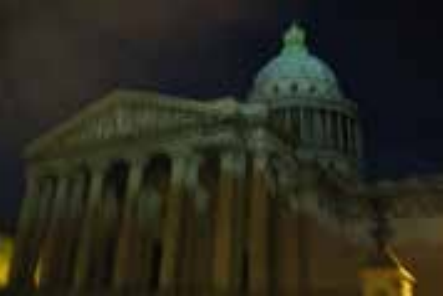}}\\
\subfloat{\includegraphics[width=2.0in]{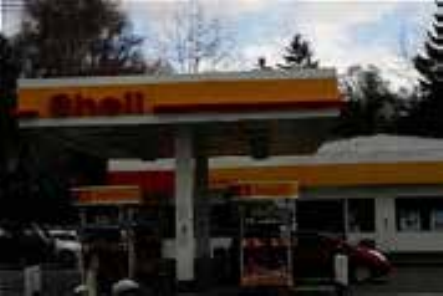}}\hspace{0.02cm}
\subfloat{\includegraphics[width=2.0in]{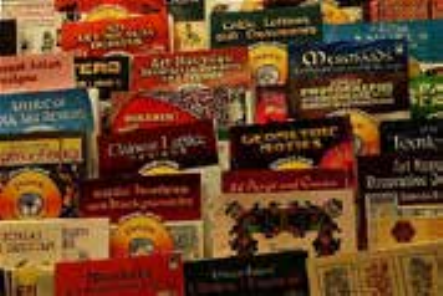}}\hspace{0.02cm}
\subfloat{\includegraphics[width=2.0in]{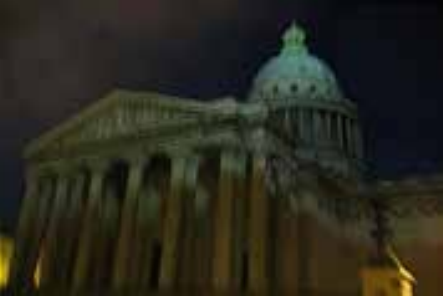}}
\caption{Restored images by \citet{xu2013unnatural} (the first row), \citet{whyte2010non,whyte2012non} [103, 104] (the second row), and \citet{hu2012fast} (the third row).}
\end{figure*}

\section{Conclusion} \label{conclusion}

In this paper, we reviewed recent developments for image deblurring, including non-blind/blind, and spatially invariant/variant techniques. We mainly classify these methods into five categories according to the manners of handling ill-posedness: Bayesian inference framework, variational methods, sparse representation-based methods, homography-based modeling, and region-based methods. Bayesian inference framework-based approaches can be grouped into three sub-categories: MAP methods, MMSE methods, and variational Bayesian methods. Researches on variational framework are focused on the development of various regularization techniques and optimization methods, as well as the interpretation of analysis operators, synthesis operators and their combinations. Sparse representation-based methods assume that the sharp images share similar geometric structures which form a redundant dictionary possessing strong expressivity. How to induce the dictionary as well as how to enhance the smoothness in the resultant images under the dictionary are two critical issues of the sparse representation framework. Homography-based modeling and region-based modeling are two techniques designed for spatially variant deblurring, where the former accounts for the temporary accumulation of photons, while the later concerns the spatial variation of the blur kernel. Other types of technique include projection-based methods, kernel regression, stochastic deconvolution, spectral analysis. Besides these mathematical considerations, hardware modification techniques are also attractive to improve the performance of image deblurring.

In these conclusive comments, we would like to discuss some theoretical and practical aspects of these developments which, we expect, will help readers regarding their future research:

\noindent 1) In blind uniform deblurring, the $\text{MAP}_{x,h}$ schemes under a sparse gradient prior will result in a delta kernel and a blurry image. However, the $\text{MAP}_h$ settings are able to approximate the true blur kernel, which can be solved by variational Bayesian methods.

\noindent 2) Even though the estimator (such as $\text{MAP}_h$ estimator) is important in deblurring task, the selection of prior is always critical. In blind uniform deblurring, $\text{MAP}_h$ is guaranteed to provide an accurate kernel when the size of the blurry image is infinitely large. In real applications, however, the size of the obtained image is generally limited, under which condition an accurate kernel would be estimated by incorporating a proper prior (such as GSM). The same conclusion can also apply to the variational framework where the prior is replaced by the regularizer.

\noindent 3) Either the analysis operator or the synthesis operator has specific properties benefitting the deblurring task. Combining these operators (such as BM3D) is appreciated for improving the deblurring performance.

\noindent 4) Compared with human heuristics, data-driven techniques are promising for deriving priors in Bayesian inference framework, regularizers in variational framework, and dictionaries in sparse representation-based framework. The "data-driven" includes two types of aspect: the first is to explore the information from the target blurry image, while the second is to obtain the additional information from other sharp images (cf. the learning scheme in Section \ref{dictionarylearning} and \ref{learningdeblurring}). The difference between these two types of aspect concerns that the first type can provide a better performance but in low efficiency, whereas the second type results in a slight degradation in the results but in high efficiency, since the learning procedure can be deployed in an offline manner.

\noindent 5) For deblurring, since an images can be viewed as a composition of large amount of repetitive patterns, exploiting the repetitive nature of images, which is often termed as a non-local strategy, is an effective way to boost the deblurring performance. This has been already proven in various image processing tasks including image deblurring, image denoising, image super-resolution, etc. Besides, the conventional local strategies can overcome a common issue of the non-local strategies that is the loss of local smoothness. Therefore, the combination of local and non-local strategies is recommended.

\noindent 6) Spatially variant deblurring is a difficult task and the progress in this aspect is very limited. Incorporating a certain physical model (such as the modeling of object motion and the modeling of camera motion) is an option to improve the performance. However, this cannot satisfy the complex conditions in real applications.

\noindent 7) Benefitting from the massive high-quality images available on the Internet and the advanced technologies of hardware, extra information can be recorded and employed for the subsequent deblurring as well as other restoration tasks. Learning-based deblurring and hardware modifications are therefore two promising directions to improve the performance of image deblurring.

\noindent 8) Image deblurring is a crucial step to produce high-quality images for high-level vision tasks. However, when limited information is accessible, the deblurring performance could be restricted. In this case, a recommended option is to combine image deblurring and high-level vision tasks into a unified problem, such as classification of blurry images and detection of blurry objects. Then the deblurring task can benefit from the high-level vision tasks because they are generally human supervised. Therefore, the cross information among the images and among the tasks are worth to be exploited.


\bibliographystyle{spbasic}      
\bibliography{myref}   

\end{document}